\documentclass[preprint,12pt]{elsarticle}

\usepackage[british]{babel}

\usepackage{natbib} 
    
\usepackage{mathtools} 
\usepackage{booktabs} 
\usepackage{tikz} 
\usetikzlibrary{calc}
\usetikzlibrary{patterns}
\usepackage{pgfplots}
\usepackage{pgfplotstable}
\pgfplotsset{width=10cm,compat=newest}
\usetikzlibrary{external,matrix}
\usepackage[draft]{todonotes}
\usepackage[dvips]{rotating}
\usepackage[noend]{RSalgorithmic}
\usepackage{algorithm}
\usepackage{amscd}
\usepackage{amsfonts}
\usepackage{amssymb}
\usepackage{color}
\usepackage{comment}
\usepackage{datetime}
\usepackage{epsfig}
\usepackage{graphicx}
\usepackage{xifthen}
\usepackage{latexsym}
\usepackage{marginnote}

\usepackage{multirow}
\usepackage{paralist}
\usepackage{placeins}
\usepackage{rotate} 
\usepackage{stmaryrd}
\usepackage{tikz}
\usepackage[draft]{todonotes}
\usepackage{varwidth}
\usepackage{verbatim} 
\usepackage{wrapfig}
\usepackage{xcolor,colortbl}
\usepackage{xspace}

\usepackage[dvips]{rotating}
\usepackage[noend]{RSalgorithmic}
\usepackage{algorithm}
\usepackage{amscd}
\usepackage{amsfonts}
\usepackage{amssymb}
\usepackage{color}
\usepackage{comment}
\usepackage{datetime}
\usepackage{epsfig}
\usepackage{ifthen}
\usepackage{latexsym}
\usepackage{marginnote}
\usepackage{multirow}
\usepackage{placeins}
\usepackage{rotate} 
\usepackage{stmaryrd}
\usepackage{tikz}

\usepackage{varwidth}
\usepackage{verbatim} 
\usepackage{wrapfig}
\usepackage{xcolor,colortbl}
\usepackage{xspace}
\usepackage{caption}
\usepackage{subcaption}
\usepackage{colour-blind}
\usepackage{lipsum}  
\usepackage[T1]{fontenc}
\usepackage{lmodern}
\usepackage{microtype}

\newtheorem{example}{Example}

\usepackage[normalem]{ulem}

\usepackage{float}

\usepackage{lineno,hyperref}
\modulolinenumbers[5]

\journal{Journal of \LaTeX\ Templates}

\definecolor {snow}                {rgb}{1.00,0.98,0.98}
\definecolor {ghostwhite}          {rgb}{0.97,0.97,1.00}
\definecolor {whitesmoke}          {rgb}{0.96,0.96,0.96}
\definecolor {gainsboro}           {rgb}{0.86,0.86,0.86}
\definecolor {floralwhite}         {rgb}{1.00,0.98,0.94}
\definecolor {oldlace}             {rgb}{0.99,0.96,0.90}
\definecolor {linen}               {rgb}{0.98,0.94,0.90}
\definecolor {antiquewhite}        {rgb}{0.98,0.92,0.84}
\definecolor {papayawhip}          {rgb}{1.00,0.94,0.84}
\definecolor {blanchedalmond}      {rgb}{1.00,0.92,0.80}
\definecolor {bisque}              {rgb}{1.00,0.89,0.77}
\definecolor {peachpuff}           {rgb}{1.00,0.85,0.73}
\definecolor {navajowhite}         {rgb}{1.00,0.87,0.68}
\definecolor {moccasin}            {rgb}{1.00,0.89,0.71}
\definecolor {cornsilk}            {rgb}{1.00,0.97,0.86}
\definecolor {ivory}               {rgb}{1.00,1.00,0.94}
\definecolor {lemonchiffon}        {rgb}{1.00,0.98,0.80}
\definecolor {seashell}            {rgb}{1.00,0.96,0.93}
\definecolor {honeydew}            {rgb}{0.94,1.00,0.94}
\definecolor {mintcream}           {rgb}{0.96,1.00,0.98}
\definecolor {azure}               {rgb}{0.94,1.00,1.00}
\definecolor {aliceblue}           {rgb}{0.94,0.97,1.00}
\definecolor {lavender}            {rgb}{0.90,0.90,0.98}
\definecolor {lavenderblush}       {rgb}{1.00,0.94,0.96}
\definecolor {mistyrose}           {rgb}{1.00,0.89,0.88}
\definecolor {white}               {rgb}{1.00,1.00,1.00}
\definecolor {black}               {rgb}{0.00,0.00,0.00}
\definecolor {darkslategray}       {rgb}{0.18,0.31,0.31}
\definecolor {dimgray}             {rgb}{0.41,0.41,0.41}
\definecolor {slategray}           {rgb}{0.44,0.50,0.56}
\definecolor {lightslategray}      {rgb}{0.47,0.53,0.60}
\definecolor {gray}                {rgb}{0.75,0.75,0.75}
\definecolor {lightgrey}           {rgb}{0.83,0.83,0.83}
\definecolor {midnightblue}        {rgb}{0.10,0.10,0.44}
\definecolor {navy}                {rgb}{0.00,0.00,0.50}
\definecolor {cornflowerblue}      {rgb}{0.39,0.58,0.93}
\definecolor {darkslateblue}       {rgb}{0.28,0.24,0.55}
\definecolor {slateblue}           {rgb}{0.42,0.35,0.80}
\definecolor {mediumslateblue}     {rgb}{0.48,0.41,0.93}
\definecolor {lightslateblue}      {rgb}{0.52,0.44,1.00}
\definecolor {mediumblue}          {rgb}{0.00,0.00,0.80}
\definecolor {royalblue}           {rgb}{0.25,0.41,0.88}
\definecolor {blue}                {rgb}{0.00,0.00,1.00}
\definecolor {dodgerblue}          {rgb}{0.12,0.56,1.00}
\definecolor {deepskyblue}         {rgb}{0.00,0.75,1.00}
\definecolor {skyblue}             {rgb}{0.53,0.81,0.92}
\definecolor {lightskyblue}        {rgb}{0.53,0.81,0.98}
\definecolor {steelblue}           {rgb}{0.27,0.51,0.71}
\definecolor {lightsteelblue}      {rgb}{0.69,0.77,0.87}
\definecolor {lightblue}           {rgb}{0.68,0.85,0.90}
\definecolor {powderblue}          {rgb}{0.69,0.88,0.90}
\definecolor {paleturquoise}       {rgb}{0.69,0.93,0.93}
\definecolor {darkturquoise}       {rgb}{0.00,0.81,0.82}
\definecolor {mediumturquoise}     {rgb}{0.28,0.82,0.80}
\definecolor {turquoise}           {rgb}{0.25,0.88,0.82}
\definecolor {cyan}                {rgb}{0.00,1.00,1.00}
\definecolor {lightcyan}           {rgb}{0.88,1.00,1.00}
\definecolor {cadetblue}           {rgb}{0.37,0.62,0.63}
\definecolor {mediumaquamarine}    {rgb}{0.40,0.80,0.67}
\definecolor {aquamarine}          {rgb}{0.50,1.00,0.83}
\definecolor {darkgreen}           {rgb}{0.00,0.39,0.00}
\definecolor {darkolivegreen}      {rgb}{0.33,0.42,0.18}
\definecolor {darkseagreen}        {rgb}{0.56,0.74,0.56}
\definecolor {seagreen}            {rgb}{0.18,0.55,0.34}
\definecolor {mediumseagreen}      {rgb}{0.24,0.70,0.44}
\definecolor {lightseagreen}       {rgb}{0.13,0.70,0.67}
\definecolor {palegreen}           {rgb}{0.60,0.98,0.60}
\definecolor {springgreen}         {rgb}{0.00,1.00,0.50}
\definecolor {lawngreen}           {rgb}{0.49,0.99,0.00}
\definecolor {green}               {rgb}{0.00,1.00,0.00}
\definecolor {chartreuse}          {rgb}{0.50,1.00,0.00}
\definecolor {mediumspringgreen}   {rgb}{0.00,0.98,0.60}
\definecolor {greenyellow}         {rgb}{0.68,1.00,0.18}
\definecolor {limegreen}           {rgb}{0.20,0.80,0.20}
\definecolor {yellowgreen}         {rgb}{0.60,0.80,0.20}
\definecolor {forestgreen}         {rgb}{0.13,0.55,0.13}
\definecolor {olivedrab}           {rgb}{0.42,0.56,0.14}
\definecolor {darkkhaki}           {rgb}{0.74,0.72,0.42}
\definecolor {khaki}               {rgb}{0.94,0.90,0.55}
\definecolor {palegoldenrod}       {rgb}{0.93,0.91,0.67}
\definecolor {lightgoldenrodyellow} {rgb}{0.98,0.98,0.82}
\definecolor {lightyellow}         {rgb}{1.00,1.00,0.88}
\definecolor {yellow}              {rgb}{1.00,1.00,0.00}
\definecolor {gold}                {rgb}{1.00,0.84,0.00}
\definecolor {lightgoldenrod}      {rgb}{0.93,0.87,0.51}
\definecolor {goldenrod}           {rgb}{0.85,0.65,0.13}
\definecolor {darkgoldenrod}       {rgb}{0.72,0.53,0.04}
\definecolor {rosybrown}           {rgb}{0.74,0.56,0.56}
\definecolor {indianred}           {rgb}{0.80,0.36,0.36}
\definecolor {saddlebrown}         {rgb}{0.55,0.27,0.07}
\definecolor {sienna}              {rgb}{0.63,0.32,0.18}
\definecolor {peru}                {rgb}{0.80,0.52,0.25}
\definecolor {burlywood}           {rgb}{0.87,0.72,0.53}
\definecolor {beige}               {rgb}{0.96,0.96,0.86}
\definecolor {wheat}               {rgb}{0.96,0.87,0.70}
\definecolor {sandybrown}          {rgb}{0.96,0.64,0.38}
\definecolor {tan}                 {rgb}{0.82,0.71,0.55}
\definecolor {chocolate}           {rgb}{0.82,0.41,0.12}
\definecolor {firebrick}           {rgb}{0.70,0.13,0.13}
\definecolor {brown}               {rgb}{0.65,0.16,0.16}
\definecolor {darksalmon}          {rgb}{0.91,0.59,0.48}
\definecolor {salmon}              {rgb}{0.98,0.50,0.45}
\definecolor {lightsalmon}         {rgb}{1.00,0.63,0.48}
\definecolor {orange}              {rgb}{1.00,0.65,0.00}
\definecolor {darkorange}          {rgb}{1.00,0.55,0.00}
\definecolor {coral}               {rgb}{1.00,0.50,0.31}
\definecolor {lightcoral}          {rgb}{0.94,0.50,0.50}
\definecolor {tomato}              {rgb}{1.00,0.39,0.28}
\definecolor {orangered}           {rgb}{1.00,0.27,0.00}
\definecolor {red}                 {rgb}{1.00,0.00,0.00}
\definecolor {hotpink}             {rgb}{1.00,0.41,0.71}
\definecolor {deeppink}            {rgb}{1.00,0.08,0.58}
\definecolor {pink}                {rgb}{1.00,0.75,0.80}
\definecolor {lightpink}           {rgb}{1.00,0.71,0.76}
\definecolor {palevioletred}       {rgb}{0.86,0.44,0.58}
\definecolor {maroon}              {rgb}{0.69,0.19,0.38}
\definecolor {mediumvioletred}     {rgb}{0.78,0.08,0.52}
\definecolor {violetred}           {rgb}{0.82,0.13,0.56}
\definecolor {magenta}             {rgb}{1.00,0.00,1.00}
\definecolor {violet}              {rgb}{0.93,0.51,0.93}
\definecolor {plum}                {rgb}{0.87,0.63,0.87}
\definecolor {orchid}              {rgb}{0.85,0.44,0.84}
\definecolor {mediumorchid}        {rgb}{0.73,0.33,0.83}
\definecolor {darkorchid}          {rgb}{0.60,0.20,0.80}
\definecolor {darkviolet}          {rgb}{0.58,0.00,0.83}
\definecolor {blueviolet}          {rgb}{0.54,0.17,0.89}
\definecolor {purple}              {rgb}{0.63,0.13,0.94}
\definecolor {mediumpurple}        {rgb}{0.58,0.44,0.86}
\definecolor {thistle}             {rgb}{0.85,0.75,0.85}
\definecolor {snow2}               {rgb}{0.93,0.91,0.91}
\definecolor {snow3}               {rgb}{0.80,0.79,0.79}
\definecolor {snow4}               {rgb}{0.55,0.54,0.54}
\definecolor {seashell2}           {rgb}{0.93,0.90,0.87}
\definecolor {seashell3}           {rgb}{0.80,0.77,0.75}
\definecolor {seashell4}           {rgb}{0.55,0.53,0.51}
\definecolor {antiquewhite1}       {rgb}{1.00,0.94,0.86}
\definecolor {antiquewhite2}       {rgb}{0.93,0.87,0.80}
\definecolor {antiquewhite3}       {rgb}{0.80,0.75,0.69}
\definecolor {antiquewhite4}       {rgb}{0.55,0.51,0.47}
\definecolor {bisque2}             {rgb}{0.93,0.84,0.72}
\definecolor {bisque3}             {rgb}{0.80,0.72,0.62}
\definecolor {bisque4}             {rgb}{0.55,0.49,0.42}
\definecolor {peachpuff2}          {rgb}{0.93,0.80,0.68}
\definecolor {peachpuff3}          {rgb}{0.80,0.69,0.58}
\definecolor {peachpuff4}          {rgb}{0.55,0.47,0.40}
\definecolor {navajowhite2}        {rgb}{0.93,0.81,0.63}
\definecolor {navajowhite3}        {rgb}{0.80,0.70,0.55}
\definecolor {navajowhite4}        {rgb}{0.55,0.47,0.37}
\definecolor {lemonchiffon2}       {rgb}{0.93,0.91,0.75}
\definecolor {lemonchiffon3}       {rgb}{0.80,0.79,0.65}
\definecolor {lemonchiffon4}       {rgb}{0.55,0.54,0.44}
\definecolor {cornsilk2}           {rgb}{0.93,0.91,0.80}
\definecolor {cornsilk3}           {rgb}{0.80,0.78,0.69}
\definecolor {cornsilk4}           {rgb}{0.55,0.53,0.47}
\definecolor {ivory2}              {rgb}{0.93,0.93,0.88}
\definecolor {ivory3}              {rgb}{0.80,0.80,0.76}
\definecolor {ivory4}              {rgb}{0.55,0.55,0.51}
\definecolor {honeydew2}           {rgb}{0.88,0.93,0.88}
\definecolor {honeydew3}           {rgb}{0.76,0.80,0.76}
\definecolor {honeydew4}           {rgb}{0.51,0.55,0.51}
\definecolor {lavenderblush2}      {rgb}{0.93,0.88,0.90}
\definecolor {lavenderblush3}      {rgb}{0.80,0.76,0.77}
\definecolor {lavenderblush4}      {rgb}{0.55,0.51,0.53}
\definecolor {mistyrose2}          {rgb}{0.93,0.84,0.82}
\definecolor {mistyrose3}          {rgb}{0.80,0.72,0.71}
\definecolor {mistyrose4}          {rgb}{0.55,0.49,0.48}
\definecolor {azure2}              {rgb}{0.88,0.93,0.93}
\definecolor {azure3}              {rgb}{0.76,0.80,0.80}
\definecolor {azure4}              {rgb}{0.51,0.55,0.55}
\definecolor {slateblue1}          {rgb}{0.51,0.44,1.00}
\definecolor {slateblue2}          {rgb}{0.48,0.40,0.93}
\definecolor {slateblue3}          {rgb}{0.41,0.35,0.80}
\definecolor {slateblue4}          {rgb}{0.28,0.24,0.55}
\definecolor {royalblue1}          {rgb}{0.28,0.46,1.00}
\definecolor {royalblue2}          {rgb}{0.26,0.43,0.93}
\definecolor {royalblue3}          {rgb}{0.23,0.37,0.80}
\definecolor {royalblue4}          {rgb}{0.15,0.25,0.55}
\definecolor {blue2}               {rgb}{0.00,0.00,0.93}
\definecolor {blue4}               {rgb}{0.00,0.00,0.55}
\definecolor {dodgerblue2}         {rgb}{0.11,0.53,0.93}
\definecolor {dodgerblue3}         {rgb}{0.09,0.45,0.80}
\definecolor {dodgerblue4}         {rgb}{0.06,0.31,0.55}
\definecolor {steelblue1}          {rgb}{0.39,0.72,1.00}
\definecolor {steelblue2}          {rgb}{0.36,0.67,0.93}
\definecolor {steelblue3}          {rgb}{0.31,0.58,0.80}
\definecolor {steelblue4}          {rgb}{0.21,0.39,0.55}
\definecolor {deepskyblue2}        {rgb}{0.00,0.70,0.93}
\definecolor {deepskyblue3}        {rgb}{0.00,0.60,0.80}
\definecolor {deepskyblue4}        {rgb}{0.00,0.41,0.55}
\definecolor {skyblue1}            {rgb}{0.53,0.81,1.00}
\definecolor {skyblue2}            {rgb}{0.49,0.75,0.93}
\definecolor {skyblue3}            {rgb}{0.42,0.65,0.80}
\definecolor {skyblue4}            {rgb}{0.29,0.44,0.55}
\definecolor {lightskyblue1}       {rgb}{0.69,0.89,1.00}
\definecolor {lightskyblue2}       {rgb}{0.64,0.83,0.93}
\definecolor {lightskyblue3}       {rgb}{0.55,0.71,0.80}
\definecolor {lightskyblue4}       {rgb}{0.38,0.48,0.55}
\definecolor {slategray1}          {rgb}{0.78,0.89,1.00}
\definecolor {slategray2}          {rgb}{0.73,0.83,0.93}
\definecolor {slategray3}          {rgb}{0.62,0.71,0.80}
\definecolor {slategray4}          {rgb}{0.42,0.48,0.55}
\definecolor {lightsteelblue1}     {rgb}{0.79,0.88,1.00}
\definecolor {lightsteelblue2}     {rgb}{0.74,0.82,0.93}
\definecolor {lightsteelblue3}     {rgb}{0.64,0.71,0.80}
\definecolor {lightsteelblue4}     {rgb}{0.43,0.48,0.55}
\definecolor {lightblue1}          {rgb}{0.75,0.94,1.00}
\definecolor {lightblue2}          {rgb}{0.70,0.87,0.93}
\definecolor {lightblue3}          {rgb}{0.60,0.75,0.80}
\definecolor {lightblue4}          {rgb}{0.41,0.51,0.55}
\definecolor {lightcyan2}          {rgb}{0.82,0.93,0.93}
\definecolor {lightcyan3}          {rgb}{0.71,0.80,0.80}
\definecolor {lightcyan4}          {rgb}{0.48,0.55,0.55}
\definecolor {paleturquoise1}      {rgb}{0.73,1.00,1.00}
\definecolor {paleturquoise2}      {rgb}{0.68,0.93,0.93}
\definecolor {paleturquoise3}      {rgb}{0.59,0.80,0.80}
\definecolor {paleturquoise4}      {rgb}{0.40,0.55,0.55}
\definecolor {cadetblue1}          {rgb}{0.60,0.96,1.00}
\definecolor {cadetblue2}          {rgb}{0.56,0.90,0.93}
\definecolor {cadetblue3}          {rgb}{0.48,0.77,0.80}
\definecolor {cadetblue4}          {rgb}{0.33,0.53,0.55}
\definecolor {turquoise1}          {rgb}{0.00,0.96,1.00}
\definecolor {turquoise2}          {rgb}{0.00,0.90,0.93}
\definecolor {turquoise3}          {rgb}{0.00,0.77,0.80}
\definecolor {turquoise4}          {rgb}{0.00,0.53,0.55}
\definecolor {cyan2}               {rgb}{0.00,0.93,0.93}
\definecolor {cyan3}               {rgb}{0.00,0.80,0.80}
\definecolor {cyan4}               {rgb}{0.00,0.55,0.55}
\definecolor {darkslategray1}      {rgb}{0.59,1.00,1.00}
\definecolor {darkslategray2}      {rgb}{0.55,0.93,0.93}
\definecolor {darkslategray3}      {rgb}{0.47,0.80,0.80}
\definecolor {darkslategray4}      {rgb}{0.32,0.55,0.55}
\definecolor {aquamarine2}         {rgb}{0.46,0.93,0.78}
\definecolor {aquamarine4}         {rgb}{0.27,0.55,0.45}
\definecolor {darkseagreen1}       {rgb}{0.76,1.00,0.76}
\definecolor {darkseagreen2}       {rgb}{0.71,0.93,0.71}
\definecolor {darkseagreen3}       {rgb}{0.61,0.80,0.61}
\definecolor {darkseagreen4}       {rgb}{0.41,0.55,0.41}
\definecolor {seagreen1}           {rgb}{0.33,1.00,0.62}
\definecolor {seagreen2}           {rgb}{0.31,0.93,0.58}
\definecolor {seagreen3}           {rgb}{0.26,0.80,0.50}
\definecolor {palegreen1}          {rgb}{0.60,1.00,0.60}
\definecolor {palegreen2}          {rgb}{0.56,0.93,0.56}
\definecolor {palegreen3}          {rgb}{0.49,0.80,0.49}
\definecolor {palegreen4}          {rgb}{0.33,0.55,0.33}
\definecolor {springgreen2}        {rgb}{0.00,0.93,0.46}
\definecolor {springgreen3}        {rgb}{0.00,0.80,0.40}
\definecolor {springgreen4}        {rgb}{0.00,0.55,0.27}
\definecolor {green2}              {rgb}{0.00,0.93,0.00}
\definecolor {green3}              {rgb}{0.00,0.80,0.00}
\definecolor {green4}              {rgb}{0.00,0.55,0.00}
\definecolor {chartreuse2}         {rgb}{0.46,0.93,0.00}
\definecolor {chartreuse3}         {rgb}{0.40,0.80,0.00}
\definecolor {chartreuse4}         {rgb}{0.27,0.55,0.00}
\definecolor {olivedrab1}          {rgb}{0.75,1.00,0.24}
\definecolor {olivedrab2}          {rgb}{0.70,0.93,0.23}
\definecolor {olivedrab4}          {rgb}{0.41,0.55,0.13}
\definecolor {darkolivegreen1}     {rgb}{0.79,1.00,0.44}
\definecolor {darkolivegreen2}     {rgb}{0.74,0.93,0.41}
\definecolor {darkolivegreen3}     {rgb}{0.64,0.80,0.35}
\definecolor {darkolivegreen4}     {rgb}{0.43,0.55,0.24}
\definecolor {khaki1}              {rgb}{1.00,0.96,0.56}
\definecolor {khaki2}              {rgb}{0.93,0.90,0.52}
\definecolor {khaki3}              {rgb}{0.80,0.78,0.45}
\definecolor {khaki4}              {rgb}{0.55,0.53,0.31}
\definecolor {lightgoldenrod1}     {rgb}{1.00,0.93,0.55}
\definecolor {lightgoldenrod2}     {rgb}{0.93,0.86,0.51}
\definecolor {lightgoldenrod3}     {rgb}{0.80,0.75,0.44}
\definecolor {lightgoldenrod4}     {rgb}{0.55,0.51,0.30}
\definecolor {lightyellow2}        {rgb}{0.93,0.93,0.82}
\definecolor {lightyellow3}        {rgb}{0.80,0.80,0.71}
\definecolor {lightyellow4}        {rgb}{0.55,0.55,0.48}
\definecolor {yellow2}             {rgb}{0.93,0.93,0.00}
\definecolor {yellow3}             {rgb}{0.80,0.80,0.00}
\definecolor {yellow4}             {rgb}{0.55,0.55,0.00}
\definecolor {gold2}               {rgb}{0.93,0.79,0.00}
\definecolor {gold3}               {rgb}{0.80,0.68,0.00}
\definecolor {gold4}               {rgb}{0.55,0.46,0.00}
\definecolor {goldenrod1}          {rgb}{1.00,0.76,0.15}
\definecolor {goldenrod2}          {rgb}{0.93,0.71,0.13}
\definecolor {goldenrod3}          {rgb}{0.80,0.61,0.11}
\definecolor {goldenrod4}          {rgb}{0.55,0.41,0.08}
\definecolor {darkgoldenrod1}      {rgb}{1.00,0.73,0.06}
\definecolor {darkgoldenrod2}      {rgb}{0.93,0.68,0.05}
\definecolor {darkgoldenrod3}      {rgb}{0.80,0.58,0.05}
\definecolor {darkgoldenrod4}      {rgb}{0.55,0.40,0.03}
\definecolor {rosybrown1}          {rgb}{1.00,0.76,0.76}
\definecolor {rosybrown2}          {rgb}{0.93,0.71,0.71}
\definecolor {rosybrown3}          {rgb}{0.80,0.61,0.61}
\definecolor {rosybrown4}          {rgb}{0.55,0.41,0.41}
\definecolor {indianred1}          {rgb}{1.00,0.42,0.42}
\definecolor {indianred2}          {rgb}{0.93,0.39,0.39}
\definecolor {indianred3}          {rgb}{0.80,0.33,0.33}
\definecolor {indianred4}          {rgb}{0.55,0.23,0.23}
\definecolor {sienna1}             {rgb}{1.00,0.51,0.28}
\definecolor {sienna2}             {rgb}{0.93,0.47,0.26}
\definecolor {sienna3}             {rgb}{0.80,0.41,0.22}
\definecolor {sienna4}             {rgb}{0.55,0.28,0.15}
\definecolor {burlywood1}          {rgb}{1.00,0.83,0.61}
\definecolor {burlywood2}          {rgb}{0.93,0.77,0.57}
\definecolor {burlywood3}          {rgb}{0.80,0.67,0.49}
\definecolor {burlywood4}          {rgb}{0.55,0.45,0.33}
\definecolor {wheat1}              {rgb}{1.00,0.91,0.73}
\definecolor {wheat2}              {rgb}{0.93,0.85,0.68}
\definecolor {wheat3}              {rgb}{0.80,0.73,0.59}
\definecolor {wheat4}              {rgb}{0.55,0.49,0.40}
\definecolor {tan1}                {rgb}{1.00,0.65,0.31}
\definecolor {tan2}                {rgb}{0.93,0.60,0.29}
\definecolor {tan4}                {rgb}{0.55,0.35,0.17}
\definecolor {chocolate1}          {rgb}{1.00,0.50,0.14}
\definecolor {chocolate2}          {rgb}{0.93,0.46,0.13}
\definecolor {chocolate3}          {rgb}{0.80,0.40,0.11}
\definecolor {firebrick1}          {rgb}{1.00,0.19,0.19}
\definecolor {firebrick2}          {rgb}{0.93,0.17,0.17}
\definecolor {firebrick3}          {rgb}{0.80,0.15,0.15}
\definecolor {firebrick4}          {rgb}{0.55,0.10,0.10}
\definecolor {brown1}              {rgb}{1.00,0.25,0.25}
\definecolor {brown2}              {rgb}{0.93,0.23,0.23}
\definecolor {brown3}              {rgb}{0.80,0.20,0.20}
\definecolor {brown4}              {rgb}{0.55,0.14,0.14}
\definecolor {salmon1}             {rgb}{1.00,0.55,0.41}
\definecolor {salmon2}             {rgb}{0.93,0.51,0.38}
\definecolor {salmon3}             {rgb}{0.80,0.44,0.33}
\definecolor {salmon4}             {rgb}{0.55,0.30,0.22}
\definecolor {lightsalmon2}        {rgb}{0.93,0.58,0.45}
\definecolor {lightsalmon3}        {rgb}{0.80,0.51,0.38}
\definecolor {lightsalmon4}        {rgb}{0.55,0.34,0.26}
\definecolor {orange2}             {rgb}{0.93,0.60,0.00}
\definecolor {orange3}             {rgb}{0.80,0.52,0.00}
\definecolor {orange4}             {rgb}{0.55,0.35,0.00}
\definecolor {darkorange1}         {rgb}{1.00,0.50,0.00}
\definecolor {darkorange2}         {rgb}{0.93,0.46,0.00}
\definecolor {darkorange3}         {rgb}{0.80,0.40,0.00}
\definecolor {darkorange4}         {rgb}{0.55,0.27,0.00}
\definecolor {coral1}              {rgb}{1.00,0.45,0.34}
\definecolor {coral2}              {rgb}{0.93,0.42,0.31}
\definecolor {coral3}              {rgb}{0.80,0.36,0.27}
\definecolor {coral4}              {rgb}{0.55,0.24,0.18}
\definecolor {tomato2}             {rgb}{0.93,0.36,0.26}
\definecolor {tomato3}             {rgb}{0.80,0.31,0.22}
\definecolor {tomato4}             {rgb}{0.55,0.21,0.15}
\definecolor {orangered2}          {rgb}{0.93,0.25,0.00}
\definecolor {orangered3}          {rgb}{0.80,0.22,0.00}
\definecolor {orangered4}          {rgb}{0.55,0.15,0.00}
\definecolor {red2}                {rgb}{0.93,0.00,0.00}
\definecolor {red3}                {rgb}{0.80,0.00,0.00}
\definecolor {red4}                {rgb}{0.55,0.00,0.00}
\definecolor {deeppink2}           {rgb}{0.93,0.07,0.54}
\definecolor {deeppink3}           {rgb}{0.80,0.06,0.46}
\definecolor {deeppink4}           {rgb}{0.55,0.04,0.31}
\definecolor {hotpink1}            {rgb}{1.00,0.43,0.71}
\definecolor {hotpink2}            {rgb}{0.93,0.42,0.65}
\definecolor {hotpink3}            {rgb}{0.80,0.38,0.56}
\definecolor {hotpink4}            {rgb}{0.55,0.23,0.38}
\definecolor {pink1}               {rgb}{1.00,0.71,0.77}
\definecolor {pink2}               {rgb}{0.93,0.66,0.72}
\definecolor {pink3}               {rgb}{0.80,0.57,0.62}
\definecolor {pink4}               {rgb}{0.55,0.39,0.42}
\definecolor {lightpink1}          {rgb}{1.00,0.68,0.73}
\definecolor {lightpink2}          {rgb}{0.93,0.64,0.68}
\definecolor {lightpink3}          {rgb}{0.80,0.55,0.58}
\definecolor {lightpink4}          {rgb}{0.55,0.37,0.40}
\definecolor {palevioletred1}      {rgb}{1.00,0.51,0.67}
\definecolor {palevioletred2}      {rgb}{0.93,0.47,0.62}
\definecolor {palevioletred3}      {rgb}{0.80,0.41,0.54}
\definecolor {palevioletred4}      {rgb}{0.55,0.28,0.36}
\definecolor {maroon1}             {rgb}{1.00,0.20,0.70}
\definecolor {maroon2}             {rgb}{0.93,0.19,0.65}
\definecolor {maroon3}             {rgb}{0.80,0.16,0.56}
\definecolor {maroon4}             {rgb}{0.55,0.11,0.38}
\definecolor {violetred1}          {rgb}{1.00,0.24,0.59}
\definecolor {violetred2}          {rgb}{0.93,0.23,0.55}
\definecolor {violetred3}          {rgb}{0.80,0.20,0.47}
\definecolor {violetred4}          {rgb}{0.55,0.13,0.32}
\definecolor {magenta2}            {rgb}{0.93,0.00,0.93}
\definecolor {magenta3}            {rgb}{0.80,0.00,0.80}
\definecolor {magenta4}            {rgb}{0.55,0.00,0.55}
\definecolor {orchid1}             {rgb}{1.00,0.51,0.98}
\definecolor {orchid2}             {rgb}{0.93,0.48,0.91}
\definecolor {orchid3}             {rgb}{0.80,0.41,0.79}
\definecolor {orchid4}             {rgb}{0.55,0.28,0.54}
\definecolor {plum1}               {rgb}{1.00,0.73,1.00}
\definecolor {plum2}               {rgb}{0.93,0.68,0.93}
\definecolor {plum3}               {rgb}{0.80,0.59,0.80}
\definecolor {plum4}               {rgb}{0.55,0.40,0.55}
\definecolor {mediumorchid1}       {rgb}{0.88,0.40,1.00}
\definecolor {mediumorchid2}       {rgb}{0.82,0.37,0.93}
\definecolor {mediumorchid3}       {rgb}{0.71,0.32,0.80}
\definecolor {mediumorchid4}       {rgb}{0.48,0.22,0.55}
\definecolor {darkorchid1}         {rgb}{0.75,0.24,1.00}
\definecolor {darkorchid2}         {rgb}{0.70,0.23,0.93}
\definecolor {darkorchid3}         {rgb}{0.60,0.20,0.80}
\definecolor {darkorchid4}         {rgb}{0.41,0.13,0.55}
\definecolor {purple1}             {rgb}{0.61,0.19,1.00}
\definecolor {purple2}             {rgb}{0.57,0.17,0.93}
\definecolor {purple3}             {rgb}{0.49,0.15,0.80}
\definecolor {purple4}             {rgb}{0.33,0.10,0.55}
\definecolor {mediumpurple1}       {rgb}{0.67,0.51,1.00}
\definecolor {mediumpurple2}       {rgb}{0.62,0.47,0.93}
\definecolor {mediumpurple3}       {rgb}{0.54,0.41,0.80}
\definecolor {mediumpurple4}       {rgb}{0.36,0.28,0.55}
\definecolor {thistle1}            {rgb}{1.00,0.88,1.00}
\definecolor {thistle2}            {rgb}{0.93,0.82,0.93}
\definecolor {thistle3}            {rgb}{0.80,0.71,0.80}
\definecolor {thistle4}            {rgb}{0.55,0.48,0.55}
\definecolor {gray1}               {rgb}{0.01,0.01,0.01}
\definecolor {gray2}               {rgb}{0.02,0.02,0.02}
\definecolor {gray3}               {rgb}{0.03,0.03,0.03}
\definecolor {gray4}               {rgb}{0.04,0.04,0.04}
\definecolor {gray5}               {rgb}{0.05,0.05,0.05}
\definecolor {gray6}               {rgb}{0.06,0.06,0.06}
\definecolor {gray7}               {rgb}{0.07,0.07,0.07}
\definecolor {gray8}               {rgb}{0.08,0.08,0.08}
\definecolor {gray9}               {rgb}{0.09,0.09,0.09}
\definecolor {gray10}              {rgb}{0.10,0.10,0.10}
\definecolor {gray11}              {rgb}{0.11,0.11,0.11}
\definecolor {gray12}              {rgb}{0.12,0.12,0.12}
\definecolor {gray13}              {rgb}{0.13,0.13,0.13}
\definecolor {gray14}              {rgb}{0.14,0.14,0.14}
\definecolor {gray15}              {rgb}{0.15,0.15,0.15}
\definecolor {gray16}              {rgb}{0.16,0.16,0.16}
\definecolor {gray17}              {rgb}{0.17,0.17,0.17}
\definecolor {gray18}              {rgb}{0.18,0.18,0.18}
\definecolor {gray19}              {rgb}{0.19,0.19,0.19}
\definecolor {gray20}              {rgb}{0.20,0.20,0.20}
\definecolor {gray21}              {rgb}{0.21,0.21,0.21}
\definecolor {gray22}              {rgb}{0.22,0.22,0.22}
\definecolor {gray23}              {rgb}{0.23,0.23,0.23}
\definecolor {gray24}              {rgb}{0.24,0.24,0.24}
\definecolor {gray25}              {rgb}{0.25,0.25,0.25}
\definecolor {gray26}              {rgb}{0.26,0.26,0.26}
\definecolor {gray27}              {rgb}{0.27,0.27,0.27}
\definecolor {gray28}              {rgb}{0.28,0.28,0.28}
\definecolor {gray29}              {rgb}{0.29,0.29,0.29}
\definecolor {gray30}              {rgb}{0.30,0.30,0.30}
\definecolor {gray31}              {rgb}{0.31,0.31,0.31}
\definecolor {gray32}              {rgb}{0.32,0.32,0.32}
\definecolor {gray33}              {rgb}{0.33,0.33,0.33}
\definecolor {gray34}              {rgb}{0.34,0.34,0.34}
\definecolor {gray35}              {rgb}{0.35,0.35,0.35}
\definecolor {gray36}              {rgb}{0.36,0.36,0.36}
\definecolor {gray37}              {rgb}{0.37,0.37,0.37}
\definecolor {gray38}              {rgb}{0.38,0.38,0.38}
\definecolor {gray39}              {rgb}{0.39,0.39,0.39}
\definecolor {gray40}              {rgb}{0.40,0.40,0.40}
\definecolor {gray42}              {rgb}{0.42,0.42,0.42}
\definecolor {gray43}              {rgb}{0.43,0.43,0.43}
\definecolor {gray44}              {rgb}{0.44,0.44,0.44}
\definecolor {gray45}              {rgb}{0.45,0.45,0.45}
\definecolor {gray46}              {rgb}{0.46,0.46,0.46}
\definecolor {gray47}              {rgb}{0.47,0.47,0.47}
\definecolor {gray48}              {rgb}{0.48,0.48,0.48}
\definecolor {gray49}              {rgb}{0.49,0.49,0.49}
\definecolor {gray50}              {rgb}{0.50,0.50,0.50}
\definecolor {gray51}              {rgb}{0.51,0.51,0.51}
\definecolor {gray52}              {rgb}{0.52,0.52,0.52}
\definecolor {gray53}              {rgb}{0.53,0.53,0.53}
\definecolor {gray54}              {rgb}{0.54,0.54,0.54}
\definecolor {gray55}              {rgb}{0.55,0.55,0.55}
\definecolor {gray56}              {rgb}{0.56,0.56,0.56}
\definecolor {gray57}              {rgb}{0.57,0.57,0.57}
\definecolor {gray58}              {rgb}{0.58,0.58,0.58}
\definecolor {gray59}              {rgb}{0.59,0.59,0.59}
\definecolor {gray60}              {rgb}{0.60,0.60,0.60}
\definecolor {gray61}              {rgb}{0.61,0.61,0.61}
\definecolor {gray62}              {rgb}{0.62,0.62,0.62}
\definecolor {gray63}              {rgb}{0.63,0.63,0.63}
\definecolor {gray64}              {rgb}{0.64,0.64,0.64}
\definecolor {gray65}              {rgb}{0.65,0.65,0.65}
\definecolor {gray66}              {rgb}{0.66,0.66,0.66}
\definecolor {gray67}              {rgb}{0.67,0.67,0.67}
\definecolor {gray68}              {rgb}{0.68,0.68,0.68}
\definecolor {gray69}              {rgb}{0.69,0.69,0.69}
\definecolor {gray70}              {rgb}{0.70,0.70,0.70}
\definecolor {gray71}              {rgb}{0.71,0.71,0.71}
\definecolor {gray72}              {rgb}{0.72,0.72,0.72}
\definecolor {gray73}              {rgb}{0.73,0.73,0.73}
\definecolor {gray74}              {rgb}{0.74,0.74,0.74}
\definecolor {gray75}              {rgb}{0.75,0.75,0.75}
\definecolor {gray76}              {rgb}{0.76,0.76,0.76}
\definecolor {gray77}              {rgb}{0.77,0.77,0.77}
\definecolor {gray78}              {rgb}{0.78,0.78,0.78}
\definecolor {gray79}              {rgb}{0.79,0.79,0.79}
\definecolor {gray80}              {rgb}{0.80,0.80,0.80}
\definecolor {gray81}              {rgb}{0.81,0.81,0.81}
\definecolor {gray82}              {rgb}{0.82,0.82,0.82}
\definecolor {gray83}              {rgb}{0.83,0.83,0.83}
\definecolor {gray84}              {rgb}{0.84,0.84,0.84}
\definecolor {gray85}              {rgb}{0.85,0.85,0.85}
\definecolor {gray86}              {rgb}{0.86,0.86,0.86}
\definecolor {gray87}              {rgb}{0.87,0.87,0.87}
\definecolor {gray88}              {rgb}{0.88,0.88,0.88}
\definecolor {gray89}              {rgb}{0.89,0.89,0.89}
\definecolor {gray90}              {rgb}{0.90,0.90,0.90}
\definecolor {gray91}              {rgb}{0.91,0.91,0.91}
\definecolor {gray92}              {rgb}{0.92,0.92,0.92}
\definecolor {gray93}              {rgb}{0.93,0.93,0.93}
\definecolor {gray94}              {rgb}{0.94,0.94,0.94}
\definecolor {gray95}              {rgb}{0.95,0.95,0.95}
\definecolor {gray97}              {rgb}{0.97,0.97,0.97}
\definecolor {gray98}              {rgb}{0.98,0.98,0.98}
\definecolor {gray99}              {rgb}{0.99,0.99,0.99}
\definecolor {darkgrey}            {rgb}{0.66,0.66,0.66}

\newcommand{\resp}[1]{[resp.\ #1]}

\newcommand{\TODO}[1]{{}}

\newcommand{\ignore}[1]{}

\newenvironment{rschange}{\color{darkgreen}}{\normalcolor}

\newcommand{\ignoreinshort}[1]{}
 \newcommand{\ignoreinlong}[1]{{#1}}
 \renewcommand{\ignoreinshort}[1]{\textcolor{blue}{#1}}
 \renewcommand{\ignoreinlong}[1]{}

\def\makenewenumerate#1#2{%
\newcounter{cnt#1}
\newenvironment{#1}%
{\begin{list}{\makebox[0pt][r]{#2}}%
{\setlength{\itemsep}{0pt}%
 \setlength{\parsep}{.2em}%
 \setlength{\leftmargin}{1.5em}%
 \setlength{\labelwidth}{.4em}%
 \usecounter{cnt#1}}}
{\end{list}}}

\makenewenumerate{myenumerate}{\arabic{cntmyenumerate}.}

\makenewenumerate{renumerate}{\rm(\roman{cntrenumerate})}
\makenewenumerate{renumerateprime}{\rm(\roman{cntrenumerateprime}$'$)}
\makenewenumerate{renumeratesecond}{\rm(\roman{cntrenumeratesecond}$''$)}

\makenewenumerate{aenumerate}{({\it\alph{cntaenumerate}})}
\makenewenumerate{aenumerateprime}{({\it\alph{cntaenumerateprime}$'$})}
\makenewenumerate{aenumeratesecond}{({\it\alph{cntaenumeratesecond}$''$})}

\def\newplaintheorem#1#2{%
\newtheorem{#1plain}{#2}
\newenvironment{#1}{\begin{#1plain}\rm }{\end{#1plain}}}

\newtheorem{definition}{Definition}
\newplaintheorem{notation}{Notation}
\newplaintheorem{remark}{Remark}
\newplaintheorem{cexample}{Counter-Example}

\newcommand{\sref}[1]{\S{}\ref{#1}}
\newcommand{\noi}{\noindent}

\newcommand{\tuple}[1]{\ensuremath{\langle{#1}\rangle}\xspace}
\newcommand{\set}[1]{\ensuremath{\{{#1}\}}\xspace}
\newcommand{\imp}{\ensuremath{\rightarrow}\xspace}
\newcommand{\limp}{\ensuremath{\leftarrow}\xspace}
\renewcommand{\iff}{\ensuremath{\leftrightarrow}\xspace}
\newcommand{\defas}{\ensuremath{\stackrel{\text{\tiny def}}{=}}\xspace}
\newcommand{\thus}{\ensuremath{\Longrightarrow}\xspace}

\newcommand{\pos}{\phantom{\neg}}

\newcommand\cala{\ensuremath{\mathcal{A}}\xspace}

\newcommand\calm{\ensuremath{\mathcal{M}}\xspace}

\newcommand\calt{\ensuremath{\mathcal{T}}\xspace}

\newcommand{\mub}{\ensuremath{\mu_{\calb}}\xspace}

\newcommand{\mularat}{\ensuremath{\mu_{\larat}}\xspace}
\newcommand{\muT}{\ensuremath{\mu_{\T}}\xspace}

\newcommand\mysout{\bgroup \markoverwith{{-}}\ULon}
\newcommand\nosout{\bgroup \markoverwith{{ }}\ULon}
\definecolor{mygray}{rgb}{0.90,0.90,0.90}
\definecolor{mywhite}{rgb}{1.00,1.00,1.00}

\newcommand{\muy}{\ensuremath{\mu_{\ally}}\xspace}
\newcommand{\munoy}{\ensuremath{\mu_{\overline{\ally}}}\xspace}

\newcommand{\atoms}[1]{\ensuremath{Atoms(#1)}\xspace}

\newcommand{\T}{\ensuremath{\mathcal{T}}\xspace}

\newcommand{\smt}{SMT\xspace}

\newcommand{\smttt}[1]{\ensuremath{\text{SMT}(#1)}\xspace}

\newcommand{\euf}{\ensuremath{\mathcal{EUF}}\xspace}

\newcommand{\larat}{\ensuremath{\mathcal{LA}(\mathbb{Q})}\xspace}

\renewcommand{\larat}{\ensuremath{\mathcal{LRA}}\xspace}

\newcommand{\smtT}{\smttt{\T}}
\newcommand{\smtlarat}{\smttt{\larat}}

\newcommand{\Tmodels}{\models_{\T}}
\newcommand{\laratmodels}{\models_{\larat}}

\newcommand{\pmodels}{\models_p}

\newcommand{\mathsat}{\textsc{MathSAT}\xspace}

\newcommand{\Bool}{\ensuremath{\mathsf{Bool}}\xspace}

\renewcommand{\TODO}[1]{\noindent{\textcolor{darkviolet}{{\fbox{TODO:} #1}}}}

\renewcommand{\Bool}{\ensuremath{\mathbb{B}}\xspace}

\newcommand{\conditionset}{\ensuremath{\mathbf{\Psi}}\xspace}
\newcommand{\supportwff}{\ensuremath{\chi}\xspace}

\newcommand{\myint}[3]{\ensuremath{\int_{#1}^{}\! \mathrm{d}#2 \,#3}\xspace}
\renewcommand{\myint}[3]{\ensuremath{\int_{#1}^{}#3 \ \mathrm{d}#2 }\xspace}
\newcommand{\allx}{\ensuremath{\mathbf{x}}\xspace}
\newcommand{\ally}{\ensuremath{\mathbf{y}}\xspace}

\newcommand{\allA}{\ensuremath{\mathbf{A}}\xspace}
\newcommand{\allB}{\ensuremath{\mathbf{B}}\xspace}

\newcommand{\allAstar}{\ensuremath{\mathbf{A}^*}\xspace}

\newcommand{\allpsi}{\ensuremath{\mathbf{\Psi}}\xspace}

\newcommand{\vi}{\ensuremath{\varphi}}

\newcommand{\vixa}{\ensuremath{\vi(\allx,\allA)}\xspace}
\newcommand{\vixab}{\ensuremath{\vi(\allx,\allA\cup\allB)}\xspace}
\newcommand{\vixya}{\ensuremath{\vi(\allx\cup\ally,\allA)}\xspace}

\newcommand{\mua}{\ensuremath{\mu^{\Bool}}\xspace}
\renewcommand{\mua}{\ensuremath{\mu^{\allA}}\xspace}
\renewcommand{\mularat}{\ensuremath{\mu^{\larat}}\xspace}
\renewcommand{\muT}{\ensuremath{\mu^{\T}}\xspace}

\newcommand{\muastar}{\ensuremath{\mu^{\allAstar}}\xspace}
\newcommand{\MUASTAR}{{\mathcal M}^{\allAstar}}

\renewcommand{\mub}{\ensuremath{\mu^{\allB}}\xspace}
\newcommand{\mupsi}{\ensuremath{\mu^{\conditionset}}\xspace}

\newcommand{\vistar}{\ensuremath{\vi^{*}}\xspace}
\newcommand{\vistarstar}{\ensuremath{\vi^{**}}\xspace}
\newcommand{\vistarof}[2]{\ensuremath{\vi^{*}(#1,#2)}\xspace}

\newcommand{\vimuagen}[2]{\ensuremath{#1_{|#2}}\xspace}
\renewcommand{\vimuagen}[2]{\ensuremath{#1_{[#2]}}\xspace}
\newcommand{\vimua}{\vimuagen{\vi}{\mua}}

\newcommand{\vistarmuastar}{\vimuagen{\vistar}{\muastar}}
\newcommand{\vistarstarmuastar}{\vimuagen{\vistarstar}{\muastar}}
\newcommand{\vistarstarmua}{\vimuagen{\vistarstar}{\mua}}
\newcommand{\viencsk}{\ensuremath{\vi^{***}}\xspace}
\renewcommand{\viencsk}{\ensuremath{\vi^{{\sf sk}}}\xspace}
\newcommand{\viencskmu}{\vimuagen{\viencsk}{\mu}}
\newcommand{\viencskmua}{\vimuagen{\viencsk}{\mua}}

\newcommand{\w}{\ensuremath{w}\xspace}
\newcommand{\wof}[2]{\ensuremath{w(#1,#2)}\xspace}
\newcommand{\wxa}{\wof{\allx}{\allA}}
\newcommand{\wmuagen}[1]{\ensuremath{w_{|#1}}\xspace}
\renewcommand{\wmuagen}[1]{\ensuremath{w_{[#1]}}\xspace}
\newcommand{\wmua}{\wmuagen{\mua}}

\newcommand{\wstar}{\ensuremath{w^{*}}\xspace}
\newcommand{\wstarof}[2]{\ensuremath{w^{*}(#1,#2)}\xspace}

\newcommand{\wstarmuagen}[1]{\ensuremath{w^{*}_{|#1}}\xspace}
\renewcommand{\wstarmuagen}[1]{\ensuremath{w^{*}_{[#1]}}\xspace}
\newcommand{\wstarmuastar}{\wstarmuagen{\muastar}}

\newcommand{\WMIgen}[4]{\ensuremath{{\sf WMI}(#1,#2|#3,#4)}\xspace}
\newcommand{\WMIviwxa}{\WMIgen{\vi}{\w}{\allx}{\allA}}
\newcommand{\WMINBgen}[3]{\ensuremath{{\sf WMI_{nb}}(#1,#2|#3)}\xspace}

\newcommand{\WMINBviwx}{\WMINBgen{\vi}{\w}{\allx}}

\newcommand{\WMI}{\ensuremath{{\sf WMI}}\xspace}

\renewcommand{\pmodels}{\models_{\Bool}}

\newcommand{\Tequiv}{\ensuremath{\Leftrightarrow_{\T}}}
\newcommand{\boolequiv}{\ensuremath{\Leftrightarrow_{\Bool}}}

\newcommand{\TA}[1]{\ensuremath{\calt\hspace{-.1cm}\cala(#1)}\xspace}
\newcommand{\TTA}[1]{\ensuremath{\calt\hspace{-.1cm}\calt\hspace{-.1cm}\cala(#1)}\xspace}

\newcommand{\ti}[1]{\ensuremath{\sf{t}^{(#1)}}\xspace}
\newcommand{\tn}[1]{\ti{n}}

\newcommand{\FI}{\ensuremath{{\sf FI}^{\larat}}\xspace}
\newcommand{\FIUC}{\ensuremath{{\sf FIUC}^{\larat}}\xspace}

\newcommand{\openshortcut}{\llbracket}
\newcommand{\closeshortcut}{\rrbracket}
\newcommand{\shortcut}[1]{\ensuremath{\openshortcut #1\closeshortcut\xspace}}

\newcommand{\inside}[2]{\shortcut{#1\!\in#2{}}}

\newcommand{\wencxya}{\ensuremath{\shortcut{y=\w}(\allx\cup\ally,\allA)}}
\newcommand{\wenc}{\ensuremath{\shortcut{y=\w}}}

\newcommand{\eufwenc}{\ensuremath{\shortcut{y=\w}_\euf}}
\newcommand{\skwenc}{\skeleton{\w}}
\newcommand{\skw}{\skwenc}
\newcommand{\skwencwone}{\skeleton{\w_1}}
\newcommand{\skwencwtwo}{\skeleton{\w_2}}
\newcommand{\skwencwi}{\skeleton{\w_i}}

\newcommand{\wmuastar}{\wmuagen{\muastar}}

\renewcommand{\TODO}[1]{\todo[inline,color=green!40]{{\small{#1}}}}

\newenvironment{gmchange}{\color{orange!80!red}}{\normalcolor}

\newcommand{\und}[1]{\underline{#1}}

\newcommand{\skeleton}[1]{\ensuremath{{\sf sk}(#1)}}

\newcommand{\newvistar}{\ensuremath{\vi^{**}}\xspace}

\newcommand{\myf}[1]{f_{#1}}
\newcommand{\myeuf}[1]{f_{#1}}
\renewcommand{\myeuf}[1]{f_{#1}(\allx)}

\renewcommand{\myf}[1]
{\ifthenelse%
 {\equal{#1}{11}}{x_1^2x_2}%
 {\ifthenelse%
  {\equal{#1}{12}}{x_1^3x_2}
  {\ifthenelse%
   {\equal{#1}{21}}{x_1x_2^2}
   {\ifthenelse%
    {\equal{#1}{22}}{x_1x_2^3}
    {\ifthenelse%
     {\equal{#1}{3}}{2x_1x_2}
     {3x_1x_2}
}}}}}

\newcommand{\convert}{{\sf Convert}}
\newcommand{\convertsk}{\ensuremath{{\sf Convert}_{\mathcal{S}\mathcal{K}}}}
\newcommand{\term}{{\sf term}}
\renewcommand{\term}{{w}}
\newcommand{\defs}{{\sf defs}}
\newcommand{\branch}{{\sf branch}}
\newcommand{\newvars}{\ally}
\newcommand{\conds}{{\sf conds}}
\newcommand{\vol}{\ensuremath{\mathit{vol}}}

\newcommand{\OptFN}[2]{{\ifx&#2&\ensuremath{#1}\else\ensuremath{#1(#2)}\fi}}

\newcommand{\wmis}{\ensuremath{\chi}}

\newcommand{\fgen}[2]{\ensuremath{#1_{[#2]}}}
\newcommand{\Ite}[3]{\ensuremath{{\sf If}\ #1\ {\sf Then}\ #2\ {\sf Else}\ #3{}}}
\newcommand{\CNFplaisted}[1]{\OptFN{{\sf CNF_{\textsf{pg}}}}{#1}}
\newcommand{\CNFtseitin}[1]{\OptFN{{\sf CNF_{\textsf{ts}}}}{#1}}

\newcommand{\muA}{\ensuremath{{\mu}^{\allA}}}
\newcommand{\muAres}{\ensuremath{\muA_{\mathit{residual}}}}

\newcommand{\SK}{\ensuremath{\mathcal{SK}}}
\newcommand{\LRA}{\ensuremath{\mathcal{LRA}}}
\newcommand{\SMT}{{SMT}}
\newcommand{\CNF}[1]{\OptFN{{\sf CNF}}{#1}}

\newcommand{\encSK}{\skwenc}
\newcommand{\defeq}{\defas}
\newcommand{\muLRA}{\ensuremath{{\mu}^{\LRA}}}
    
\newcommand{\ConvertSK}[1]{\OptFN{{\sf Convert}_\SK}{#1}}

\newcommand{\Simplify}[1]{\OptFN{\sf Simplify}{#1}}
\newcommand{\LitConj}[1]{\OptFN{\sf LiteralConjuction}{#1}}
\newcommand{\WMINB}[1]{\OptFN{{{\sf{WMI}}_{\sf{nb}}}}{#1}}
\definecolor{AC1}{HTML}{d5b3a0}
\definecolor{AC2}{HTML}{b2182b}
\definecolor{AC3}{HTML}{2166ac}
\definecolor{AC4}{HTML}{b0c1ca}

\colorlet{phi1T}{AC1!60!black}
\colorlet{phi1F}{AC2!80!black}
\colorlet{phi2T}{AC3!80!black}
\colorlet{phi2F}{AC4!70!black}

\newcommand{\br}[1]{{\ensuremath \pos #1 \vee \neg #1}}
\newcommand{\msatoption}{\texttt}

\floatstyle{boxed}
\newfloat{floatbox}{t}{box}
\floatname{floatbox}{Box}

\newcommand{\TODROP}[1]{\todo[color=violet!40,size=tiny]{{RS: drop?}}\textcolor{darkviolet}{#1}}
\renewcommand{\TODROP}[1]{}
\newcommand{\DROPINMAIN}[1]{\todo[color=violet!40,size=tiny]{{Dropped
      in main}}\textcolor{darkviolet}{#1}}
\renewcommand{\DROPINMAIN}[1]{#1}

\newcommand{\wmipa}{\textsc{WMI-PA}\xspace}
\newcommand{\wmisapa}{\textsc{SA-WMI-PA}\xspace}
\newcommand{\method}{\textsc{SAE4WMI}\xspace}

\newcommand{\latteintegrale}{\textsc{LattE Integrale}\xspace}
\newcommand{\latte}{\textsc{LattE}\xspace}
\newcommand{\volesti}{\textsc{VolEsti}\xspace}

\newcommand{\MU}{{\mathcal M}}
\newcommand{\MUA}{{\mathcal M}^{\allA}}
\newcommand{\MULARAT}{{\mathcal M}^{\larat}}

\makeatletter
\newenvironment{customlegend}[1][]{%
    \begingroup
    \pgfplots@init@cleared@structures
    \pgfplotsset{#1}%
}{%
    \pgfplots@createlegend
    \endgroup
}%

\def\addlegendentry{\pgfplots@addlegendentry}
\def\addlegendimage{\pgfplots@addlegendimage}
\makeatother

\definecolor{XSDD}{HTML}{b2182b}
\definecolor{FXSDD}{HTML}{ef8a62}
\definecolor{XADD}{HTML}{d5b3a0}
\definecolor{WMIPA}{HTML}{b0c1ca}
\definecolor{WMISAPA}{HTML}{67a9cf}
\definecolor{WMISAPASK}{HTML}{2166ac}
\definecolor{WMISAPASK_approx_100}{HTML}{79155B}
\definecolor{WMISAPASK_approx_1000}{HTML}{C23373}
\definecolor{WMISAPASK_approx_10000}{HTML}{F6635C}
\definecolor{WMISAPASK_approx_100000}{HTML}{FFBA86}

\begin{document}

\begin{frontmatter}

  \title{
Enhancing SMT-based Weighted Model Integration \\ by Structure Awareness}
\tnotetext[mytitlenote]{A preliminary version of this work was
  presented at UAI 2022~\cite{spallitta2022smt}.
}

\author[mymainaddress]{Giuseppe Spallitta}
\author[mymainaddress]{Gabriele Masina}
\author[mymainaddress]{Paolo Morettin}
\author[mymainaddress]{Andrea Passerini}
\author[mymainaddress]{Roberto Sebastiani}

\address[mymainaddress]{DISI, University of Trento}

\begin{abstract}
The development of efficient exact and approximate algorithms for probabilistic inference is a long-standing goal of artificial intelligence research. Whereas substantial progress has been made in dealing with purely discrete or purely continuous domains, adapting the developed solutions to tackle hybrid domains, characterized by discrete and continuous variables and their relationships, is highly non-trivial. Weighted Model Integration (WMI) recently emerged as a unifying formalism for probabilistic inference in hybrid domains. Despite a considerable amount of recent work, allowing WMI algorithms to scale with the complexity of the hybrid problem is still a challenge. In this paper we highlight some substantial limitations of existing state-of-the-art solutions, and develop an algorithm that combines SMT-based enumeration, an efficient technique in formal verification, with an effective encoding of the problem structure. This allows our algorithm to avoid generating redundant models, resulting in drastic computational savings. Additionally, we show how SMT-based approaches can seamlessly deal with different integration techniques, both exact and approximate, significantly expanding the set of problems that can be tackled by WMI technology. An extensive experimental evaluation on both synthetic and real-world datasets confirms the substantial advantage of the proposed solution over existing alternatives. The application potential of this technology is further showcased on a prototypical task aimed at verifying the fairness of probabilistic programs.
\end{abstract}


\end{frontmatter}

\section{Introduction}%
\label{sec:intro}

There is a growing interest in the artificial
intelligence community in extending probabilistic reasoning
  approaches to deal with hybrid domains, characterized by both
  continuous and discrete variables and their relationships. Indeed,
  hybrid domains are extremely common in real-world scenarios,
  from transport modelling~\citep{hensher2007handbook} to
  probabilistic robotics~\citep{Thrun:2005:PR:1121596} and
  cyber-physical systems~\citep{Lee2008}.

Early approaches to hybrid probabilistic modelling relied on
discretization of the continuous
variables~\citep{FriedmanG96,KozlovK97} or 
made strong distributional assumptions, not allowing to
condition discrete variables on continuous ones~\citep{lauritzen1992propagation}.
Later approaches, like hybrid Markov random fields~\citep{YangBRAL14} and hybrid sum-product networks~\citep{Molina18,BueSB18}, were still unable to support deterministic relations, i.e., hard constraints, between variables, severely limiting their usability in real-world applications.

Weighted Model Integration  (WMI)~\citep{Belle:2015aa, morettin-wmi-ijcar17,ijcai_surv2021} recently emerged as a unifying formalism for probabilistic inference in hybrid
domains, thanks to its expressivity, flexibility and its natural support for both deterministic and probabilistic relations. 
The paradigm extends Weighted Model Counting
(WMC)~\citep{Chavira2008}, which is the task of computing the weighted
sum of the set of satisfying assignments of a propositional formula,
to deal with SMT formulas~(e.g.,~\citep{barrettsst21}) consisting of combinations of Boolean atoms and
  connectives with symbols from a background theory, like linear
  arithmetic over rationals (\larat).
%

Although WMC can be made extremely efficient by leveraging component
caching techniques~\citep{sang2004cc,bacchus2009solving}, these
strategies are difficult to apply for WMI due to the tight coupling
induced by arithmetic constraints. Indeed,
existing component caching approaches for WMI are
restricted to fully factorized densities 
with few dependencies among continuous variables~\citep{BelleAAAI16}. Another
direction specifically targets acyclic~\citep{ZengB19,ZengMYVB20} or
loopy~\citep{ZengMYVB20b} pairwise
models. Approximations with guarantees can be computed
  for problems in disjunctive normal
  form~\citep{abboud2020approximability} or, in restricted cases,
  conjunctive normal form~\citep{BelleUAI15}.

Exact solutions for more general classes of densities and
constraints mainly take advantage of advances in SMT
technology \citep{barrettsst21} or knowledge
compilation (KC)~\citep{DarwicheM02}.
%
%
%
\wmipa~\citep{morettin-wmi-ijcar17,morettin-wmi-aij19}
relies on SMT-based Predicate Abstraction
(PA)~\citep{allsmt} to {both} reduce the number of models
to be generated and integrated over {and efficiently enumerate them},
and was shown to achieve substantial improvements over previous
solutions.
%
Nevertheless, {in this paper} we show how \wmipa{} has the major
drawback of ignoring the conditional structure of the weight function,
which prevents from pruning away lots of redundant models.
The use of KC for hybrid probabilistic inference
was pioneered by~\cite{sanner2012symbolic} and further refined in a series of
later
works~\citep{KolbMSBK18,dos2019exact,kolb2020exploit,feldstein2021lifted}. By
compiling a formula into an algebraic circuit, KC
techniques can exploit the structure of the problem to reduce the size
of the resulting circuit, and are at the core of many state-of-the-art
approaches for WMC~\citep{Chavira2008}.
Nevertheless, {in this paper we
show that} even the most recent  {KC}
solutions for WMI~\citep{dos2019exact,kolb2020exploit} have serious
troubles in dealing with densely coupled problems, resulting in
exponentially large circuits {that are impossible to
  store or prohibitively expensive to evaluate}.

To address these problems, in this paper we introduce Structure-Aware Enumeration for WMI (\method{}),
{a novel algorithm for WMI that aims} to
combine the best of both worlds, by introducing weight-structure
awareness into PA-based WMI.\@
The main idea is to build a formula, which we call the \emph{conditional skeleton}, which mimics the
conditional structure of the weight function, to drive the
SMT-based enumeration algorithm preventing it from generating
redundant models. 
%
An extensive experimental evaluation of synthetic
and real-world datasets show substantial computational advantages of
the proposed solution over existing alternatives for the most
challenging settings.

Parallel to the introduction of weight-structure awareness, we highlight how PA-based algorithms are \textit{agnostic} to the integration algorithm chosen to compute the volume of each enumerated assignment. To this extent, we extend \method{} to support both \textit{exact numerical integration} (the one implicitly embedded in~\cite{morettin-wmi-ijcar17, morettin-wmi-aij19}) and \textit{approximate integration}. The advantages of using approximate integration are twofold: it positively affects the scalability of \method{} with complex instances when integration is the major bottleneck; (ii) it allows applying \method{} to problems with non-polynomial weight functions (e.g., Gaussian densities), increasing the applicability of WMI to real-world scenarios.

Our  main contributions can be summarized as follows:

\begin{itemize}
\item We analyse existing state-of-the-art of WMI and we identify major efficiency issues for  both PA and
  KC-based approaches.
{\item We introduce \method{}, a novel WMI algorithm
       that combines PA with weight-structure awareness by generating explicitly
       a \textit{conditional skeleton}} of the weight function to drive
       the search of assignments.
    \item We show how \method{} achieves substantial
       improvements over existing solutions in both synthetic and real-world settings. 
    \item We demonstrate how PA-based approaches can deal
      with different integration techniques, both exact and
      approximate, substantially expanding the set of problems that
      can be tackled by WMI technology.
    {\item We present a prototypical application of \method{} to verifying the fairness of probabilistic programs.} 
    \end{itemize}

The rest of the paper is organized as follows. {We start by presenting the related work in \sref{sec:related}}. In~\sref{sec:background} we introduce the background, focusing on the formulation of Weighted
Model Integration. In~\sref{sec:wmipa-issues} {we analyse the WMI
approaches based on knowledge compilation and on Predicate
Abstraction, identifying some weaknesses for either of them}. In \sref{sec:wmipa-improved} we address the gaps mentioned in the
previous section, showing theoretical ideas to make \wmipa{}
structure aware and their implementation into our novel algorithm, \method{}. \sref{sec:expeval} is devoted to an empirical evaluation of
\method{} with respect to the other existing approaches, considering
both synthetic and real-world settings. In \sref{sec:approximated-wmipa}, we highlight how PA-based approaches are agnostic to the integrator chosen to compute the volume of the polytopes and propose an adaptation of the algorithm to deal with approximate integration. Our conclusion and
final considerations are presented in \sref{sec:concl}. 

%

\section{Related work}%
\label{sec:related}
Traditionally, inference algorithms in hybrid graphical models either
disallowed complex algebraic and logical
constraints~\citep{lauritzen2001stable,yang2014mixed,molina2018mixed},
or solved the problem
approximately~\citep{AfsharSW16,gogate2005approximate}.
%
%
The use of piecewise polynomial densities when dealing with algebraic
constraints was pioneered by \citet{sanner2012symbolic}.
The algorithm, called Symbolic Variable Elimination (SVE), first
used a compact representation called eXtended Algebraic Decision Diagrams (XADD)~\citep{xadd2012} in probabilistic inference tasks.  As
described in detail in~\sref{sec:wmipa-issues}, these representations
lie at the core of modern compilation-based WMI techniques.
In broad terms, knowledge compilation refers to a
  family of approaches that seek to convert problem instances in a
  target representation that, thanks to certain structural properties,
  guarantees efficient solving of a specific task.
  An offline, expensive compilation phase searches in the space of target
  representations for the most compact encoding of a problem instance.
  Then, the compiled representation of the problem instance can be
  queried efficiently with respect to its size, typically amortizing
  the computational cost of the compilation over multiple queries.
  This approach is quite popular in WMC, where propositional logic
  formulas are compiled in \emph{deterministic, decomposable normal
  negation form} (d-DNNF) circuits. Thanks to these properties, WMC
  tasks can be computed in linear time in the size of the d-DNNF.\@
  Since the problem structure has to be explicitly encoded in a target
  representation, KC techniques fail by running out of memory on
  instances where this compact representation does not exist.
Probabilistic inference modulo
theories~\citep{desalvobraz2016probabilistic} used representations
similar to XADDs together with a modified DPLL procedure. The
resulting algorithm, called PRAiSE, showed superior performance with
respect to both SVE and the original WMI~\citep{Belle:2015aa} solver,
which didn't exploit any SMT enumeration technique. Follow-up works on
SMT-based WMI capitalized on the advanced features of SMT solvers,
obtaining state-of-the-art results in many
benchmarks~\citep{morettin-wmi-ijcar17,morettin-wmi-aij19}.

The substantial efficiency gains obtained by caching partial results
in WMC~\citep{sang2004cc,bacchus2009solving} motivated their
applications in the hybrid case. When the dependencies induced by the
algebraic constraints are limited, similar ideas can be applied to
WMI~\citep{BelleAAAI16}.
Alternatively to SMT- or compilation-based approaches, WMI can be
reduced to Belief Propagation when the algebraic dependencies involve
at most two variables. Then, an efficient message passing algorithm on
an acyclic pairwise factor graph can solve the problem
exactly~\citep{ZengMYVB20}. This idea was later extended for
approximating problems with circular dependencies~\citep{ZengMYVB20b}.
In contrast with the work above, this paper focuses on the general WMI
problem, where the above assumptions cannot be leveraged.

Approximate WMI algorithms exploit both SMT-based~\citep{BelleUAI15}
and knowledge compilation-based~\citep{dos2019exact} approaches. While generally
applicable, the latter does not provide any guarantee on the
approximation error. The former provides practical statistical
guarantees in limited scenarios when constraints are expressed in
conjunctive form. Instead, if the constraints can be expressed in
disjunctive form, WMI admits a fully polynomial randomized
approximation scheme (FPRAS) under mild
assumptions~\citep{abboud2020approximability}.

The quantitative evaluation of programs by counting SMT solutions (see
e.g.~\cite{fredrikson2014satisfiability,chistikov2015approximate}) is
equivalent to the notion of \emph{unweighted} model integration.
This approach was later extended to fairness verification of decision
procedures based on small ML models.
This approach, dubbed FairSquare~\cite{albarghouthi2017fairsquare},
encodes the verification problem in \SMT($\LRA{}$) and reduces
probabilistic inference to axis-aligned volume computations.
FairSquare ultimately addresses a decision problem, answering whether
the fairness measure is above a user-defined threshold or not. This is
achieved by iteratively computing tighter lower and upper bounds using
increasingly accurate approximations of the measure until convergence.
Moreover, the distribution considered in FairSquare are probabilistic
programs where random variables have univariate Gaussian or piecewise
constant distributions. The notion of weight function considered in
this work is more general, including multivariate piecewise
polynomials of arbitrary degree and Gaussians.

\section{Background}%
\label{sec:background}

In this section, we introduce the theoretical background which is
needed for the comprehension of this paper. For the sake of
readability, we have reduced the technical details to the minimum. A
more detailed description with several SMT-related technicalities is
reported in~\ref{sec:detailedbackground}.

\subsection{SMT Notation, Syntax and Semantics }%
\label{sec:background-smt}

We assume the reader is familiar with the basic syntax, semantics, and
results of propositional and first-order logic.
We adopt the notation and definitions in~\cite{morettin-wmi-aij19} ---including some terminology and concepts from Satisfiability Modulo Theories
(SMT)--- which we summarize below.

\paragraph{Notation and terminology}
Satisfiability Modulo Theories (SMT) (see~\cite{barrettsst21}  for details) consists in deciding the
satisfiability of quantifier-free
first-order formulas over some given background theory \T. For the
context of this paper,
we restrict to the theory of linear real arithmetic (\larat{}).
Thus,  ``$\models$'' 
denotes entailment in \larat{} (e.g., \mbox{$(x\ge 2)\models (x\ge 1)$})
and we say that $\vi_1$, $\vi_2$ are \emph{equivalent}
iff they are \larat-equivalent (e.g., $(x-y\le 3)$ and $(y\ge x-3)$).

%
We use $\Bool \defas \{\top, \perp\}$ to indicate the set of Boolean
values and $\mathbb{R}$ to indicate the set of real
values.
We use the sets $\allA\defas\set{A_i}_i$,
$\allB\defas\set{B_i}_i$ to denote Boolean atoms and
the sets $\allx\defas\set{x_i}_i$,
$\ally\defas\set{y_i}_i$ to denote real variables.
SMT(\larat{}) formulas combine Boolean atoms $A_i \in
\Bool$ and \larat{} atoms in the form $(\sum_i c_i x_i\ \bowtie c)$
(where $c_i$, $c$ are rational values, $x_i$ are real variables in
$\mathbb{R}$ and $\bowtie$ is one of the standard arithmetic operators
$\{=, \neq, <, >, \leq, \geq\}$)  by using standard Boolean operators
$\{\neg,\wedge,\vee,\rightarrow,\leftrightarrow\}$.
\atoms{\vi} denotes the set of atoms occurring in $\vi$,
both Boolean and
$\larat$
ones.
%
%

  A 
  \emph{literal} is either an 
atom (a \emph{positive literal}) or its negation (a \emph{negative
  literal}).
A \emph{clause}
$\bigvee_{j=1}^{K} l_{j}$ is a disjunction of literals.
A formula is in \emph{Conjunctive Normal Form (CNF)} if it is
  written as a conjunction of clauses
  $ \bigwedge_{i=1}^{L}\bigvee_{j_i=1}^{K_i} l_{j_i} $.
Some shortcuts of frequently-used expressions (marked as
``\shortcut{\ldots}'') are provided to simplify the reading:
the formula $(x_i
\geq l) \wedge (x_i \leq u)$ is shortened into
$\inside{x_i}{[l,u]}$; if $\phi\defas\bigwedge_i C_i$ is a CNF formula
and $C\defas\bigvee_j l_j$ is a clause, then  the formula
$\bigwedge_i(C\vee C_i)$ is shortened into $\shortcut{C\vee\phi}$.      

\paragraph{Semantics}

Given a set of Boolean and \larat{} atoms
 $\allpsi\defas\set{\psi_1,\ldots,\psi_K}$, we 
call a \emph{total \resp{partial} truth assignment} $\mu$ for \allpsi{}
any total \resp{partial} map from \allpsi{} to $\Bool$.
(These definitions can be used also for sets $\allpsi$ of non-atomic
\larat{} formulas.)  {We denote by $\Bool^K$ the set of all total
  truth assignments over \allpsi.}
With a little abuse of notation, 
we represent $\mu$
either as a set or as a conjunction of literals:
$\mu \defas \{\psi\ |\ \psi \in \allpsi{},\ \mu(\psi)=\top \} 
\cup \{\neg\psi\ |\ \psi \in \allpsi{},\ \mu(\psi)=\bot \}$, or
$
\mu \defas \bigwedge_{\psi \in \allpsi{},{\mu(\psi)=\top}} \pos\psi
\wedge
\bigwedge_{\psi \in \allpsi{},{\mu(\psi)=\bot}} \neg\psi
$,
and we write ``$\psi_i\in\mu_1$'' and ``$\mu_1\subseteq\mu_2$'' as if 
$\mu_1,\mu_2$ were represented as sets (i.e., we write
``$\psi_1\wedge\neg \psi_2\subseteq \psi_1\wedge\neg \psi_2\wedge \psi_3$'' 
meaning 
``$\set{\psi_1,\neg \psi_2}\subseteq \set{\psi_1,\neg \psi_2,
  \psi_3}$'').
We denote by \mua{} and \mularat{} its two components on the Boolean atoms in \allA{} and on the \larat{} atoms, respectively, so that 
$\mu\defas\mua\wedge\mularat$.
\begin{example}
    If $\mu\defas \set{A_1=\top,A_2=\bot,(x \ge 1)=\top,(x \ge 3)=\bot}$,
then we represent  it as the set $\mu\defas \set{A_1, \neg A_2 , (x \ge
1) , \neg (x \ge 3)}$
or as the formula $\mu\defas A_1\wedge \neg A_2 \wedge (x \ge
1) \wedge \neg (x \ge 3)$.
Hence $\mu=\mua\wedge \mularat$ where
$\mua\defas A_1\wedge \neg A_2$ and
$\mularat\defas (x \ge 1) \wedge \neg (x \ge 3)$.\hfill $\diamond$
\end{example}

Given a (partial) truth assignment $\mu$ to \atoms{\vi},  we call the
\emph{residual} of $\vi$ w.r.t.\ $\mu$, written ``$\vimuagen{\vi}{\mu}$'',
the formula obtained from $\vi$ by substituting all the atoms assigned
in $\mu$ with their respective truth value, and by recursively propagating truth
values through Boolean operators in the usual way.\footnote{I.e.,
  $\neg\top\mapsto \bot$, 
  $\neg\bot\mapsto \top$, 
  $\vi\wedge\top \mapsto \vi$, 
  $\vi\wedge\bot \mapsto \bot$, 
  $\vi\iff \top \mapsto \vi$,
  $\vi\iff \bot \mapsto \neg\vi$,
  \ldots}
%
We say that $\mu$ \emph{propositionally satisfies} $\vi$ iff $\vimuagen{\vi}{\mu}$
reduces to $\top$.%
\footnote{%
  The definition of satisfiability by partial assignment may
    present some ambiguities for non-CNF and existentially-quantified
    formulas~\cite{sebastianipartial2020,mohlesb20}. Here we adopt the
    above definition because it is 
    the one typically used by state-of-the-art solvers.}
Thus, 
if $\mu$ propositionally satisfies $\vi$, then all total
assignments $\eta$ s.t.\ $\mu\subseteq \eta$ propositionally satisfy $\vi$.\@
%
Since $\mu$
 can be \larat-unsatisfiable due to its \larat{} component
(e.g., $\mu\defas\neg A_1\wedge(x\ge 2)\wedge\neg(x \ge 1)$),
 \smtlarat{} checks the existence
 of an \larat-satisfiable assignment $\mu$ propositionally satisfying
   $\vi$.
%
%
  \begin{example}
    Let
 $ 
 \vi \defas(A_1\vee (x\le 1))\wedge(\neg
 A_1\vee(x\ge 2))$,
 $\mu\defas A_1\wedge(x \ge 2)$ so that
 $\muA\defas A_1$ and $\mularat\defas(x \ge 2)$.
 Then $\vimuagen{\vi}{\muA}=(x\ge 2)$ and $\vimuagen{\vi}{\mu}=\top$,
 s.t.\ $\mu$ propositionally satisfies $\vi$.
 Since $\mularat$ is $\larat$-satisfiable, then $\vi$ is
 $\larat$-satisfiable. \hfill $\diamond$
  \end{example}%

 
 Given \vixab, we say that a Boolean (partial) assignment $\mua$ on
$\allA$ \emph{propositionally satisfies} 
 $\exists \allx. \exists \allB. \vi$
 iff $\mu\defas\mua\wedge\mub\wedge\mularat$ propositionally satisfies $\vi$
 for some total assignment $\mub$ on \allB{} and some total
 \larat-satisfiable assignment $\mularat$ on the \larat{} atoms of
 $\vi$.\@
To this extent, notice that $\exists\allx. \exists \allB. \vixab$ is
equivalent to some purely-Boolean formula on $\allA$ only, that is,
the dependence on $\allx$ and $\allB$ is existentially-quantified
away.  (The definition of ``$\mua$ propositionally satisfies
$\exists\allx.  \vixa$''
 follows with $\allB=\emptyset,\mub=\top$.)
We say that $\muA\wedge\mularat$ propositionally satisfies $\exists
\allB.\vixab$ iff $\muA\wedge\mub\wedge\mularat$ propositionally
satisfies $\vi$ for some total assignment $\mub$ in $\allB$.
%
   \begin{example}
   Let
 $\vi \defas(A_1 \vee (x\le 1))\wedge  B_1 \wedge(B_1\iff 
 (x\ge 2)) $.
 Then $\muA_1\defas A_1$ propositionally satisfies $\exists x. \exists B_1 .\vi$ because there
 exist $\mub_1\defas B_1$  and $\mularat_1\defas\neg(x\le 1)\wedge(x \ge 2)$
  s.t.  $\mu_1\defas\muA_1\wedge\mularat_1\wedge\mub_1$ propositionally
  satisfies $\vi$ and $\mularat_1$ is \larat-satisfiable.
%
Instead, $\muA_2\defas \neg A_1$ does not satisfy $\exists x. \exists
B_1 .\vi$, because no such \mub{} and  \mularat{} exist.
Indeed, $\exists x. \exists B_1 .\vi$ is equivalent to the Boolean
formula $A_1$.
Also, $\mua_1\wedge\mularat_1$ propositionally satisfies $\exists B_1.\vi$.
 \hfill $\diamond$  
   \end{example}%

\subsection{Weighted Model Integration (WMI)}%
\label{sec:background-wmi}

 Let $\allx \defas\{x_1,\dots, x_N\} \in\mathbb{R}^N$ and 
 $\allA\defas \{A_1,\dots, A_M\} \in \Bool^M$ for some integers $N$ and
 $M$,
 {let $\vi(\allx,\allA)$ denote an \larat{} formula over \allx{} and \allA{},}
 and let $\w{} (\allx,\allA)$ denote a non-negative weight
function s.t.\ $\wxa: \mathbb{R}^N\times\Bool^M \longmapsto \mathbb{R}^+$.
Intuitively, $\w{}$ encodes a (possibly unnormalized) density function
over $\allA \cup \allx$. 
Given a total assignment $\mua$ over \allA, $\wmua(\allx)\defas\w{}(\allx,\mua)$ is
\w{} restricted to 
the truth values of \mua.

The {\bf Weighted Model
  Integral} of $\w{}(\allx,\allA)$  over $\vi(\allx,\allA)$
is defined as~\citep{morettin-wmi-aij19}:
\begin{eqnarray}
  \label{eq:wmi12}
  \WMIviwxa
  &\defas&
\!\!\!\!\!
 \sum_{\mua\in \Bool^M}^{} \WMINBgen{\vimua}{\wmua}{\allx},
  \\
\hspace{-.3cm}\label{eq:wminb1}
    \WMINBviwx\
 &\defas&
  \myint{\vi(\allx)}{\allx}{\w(\allx)}
\end{eqnarray}
where 
the $\mua$s are all the
total truth assignments on \allA,
\WMINBviwx\ is the integral of $\w{}(\allx)$ over the 
set \set{\allx\ |\ \vi(\allx)\ \mathit{is\ true}} (``${\sf _{nb}}$'' means ``no-Booleans'').

We call a \emph{support} of a weight
function \wxa{} any subset of $\mathbb{R}^N\times\Bool^M$ outside which
$\wxa=0$ and we represent it as a \larat{} formula
$\supportwff(\allx,\allA)$.
We recall that, consequently,
$
  \WMIgen{\vi\wedge\supportwff}{\w}{\allx}{\allA}=
  \WMIgen{\vi}{\w}{\allx}{\allA}.
$

We consider the class of \emph{feasibly integrable on \larat{} (\FI)}
functions $\w(\allx)$, which contain no conditional component and for
which there exists some procedure able to compute
\WMINBgen{\mularat}{\w}{\allx} for every set of \larat{} literals on
\allx. (E.g., polynomials are \FI.)
Then we call a weight function \wxa{}, \emph{feasibly integrable under \larat{}
  conditions (\FIUC)} iff it can be described in terms of 
  a support \larat{} formula $\supportwff(\allx,\allA)$
($\top$ if not provided) and
  a set 
  $\conditionset\defas\set{\psi_i(\allx,\allA)}_{i=1}^K$
  of 
 \emph{\larat{} {conditions}}, 
in such a way that,
for every total 
truth assignment $\mupsi$ to $\conditionset$, 
\wmuagen{\mupsi}(\allx) is total and \FI in the domain given by the values of
\tuple{\allx,\allA} which satisfy
$(\supportwff\wedge\mupsi)$.%
\footnote{{Notice
  that the conditions in \conditionset{} can be non-atomic, and can
  contain atoms in \allA.} }
%
%
%
Intuitively, each \mupsi{} describes a portion of the domain of \w{} inside which \wmuagen{\mupsi}(\allx) is \FI, and we say that \mupsi{}  \emph{identifies} \wmuagen{\mupsi} in \w{}.

In practice, we assume w.l.o.g.~that \FIUC{} functions are described as 
combinations  of constants, variables, standard arithmetic operators
$+,-,\cdot$,
condition-less real-valued functions (e.g.,
$\exp(.), \dots$), conditional expressions
in the form
$({\sf If}\ \psi_i\ {\sf Then}\ t_{1i}\ {\sf Else}\ {t_{2i}})$ whose conditions
$\psi_i$ are \larat{} formulas and terms $t_{1i},t_{2i}$ are \FIUC{}.%
(E.g., piecewise polynomials are
  \FIUC{}.)


\subsection{WMI via Projected AllSMT}%
\label{sec:background-wmipa}

\paragraph{Assignment Enumeration}
$\TTA{\vi} \defas \{\mu_1, \ldots, \mu_j \}$ denotes the
set of \larat{}-satisfiable \textit{total} truth assignments over $Atoms(\vi)$ that propositionally satisfy $\vi$;
$\TA{\vi} \defas \{\mu_1, \ldots, \mu_j \}$ represents one set of \larat{}-satisfiable
{\em partial} truth assignments over $Atoms(\vi)$ that
propositionally satisfy $\vi$, s.t.\ 
\begin{inparaenum}[(i)]  
\item\label{item:ta:cover} for every total assignment $\eta$ in $\TTA{\vi}$ there is some partial one $\mu$ in $\TA{\vi}$ s.t.\ $\mu\subseteq\eta$ and
\item\label{item:ta:disjoint} every pair $\mu_i,\mu_j\in \TA{\vi}$ assigns
opposite truth values to at least one atom.
\end{inparaenum}
We remark that $\TTA{\vi}$ is
 unique (modulo reordering), whereas multiple $\TA{\vi}$s are admissible for the same formula $\vi$ (including $\TTA{\vi}$).
{The disjunction of the truth assignments in \TTA{\vi}, and that of
  \TA{\vi}, are 
  equivalent to \vi.}
%
Thus, given $\vixab$, $\TTA{\exists \allx. \exists \allB. \vi}$ denotes the set of all
total truth assignment $\mua{}$ on $\allA$ s.t.\ $\mua$
propositionally satisfies $\exists
\allx. \exists \allB. \vi$, and $\TA{\exists \allx. \exists
  \allB. \vi}$ denotes one set of partial assignments \mua{} on
\allA{} s.t.\ $\mua$ propositionally satisfies $\exists \allx. \exists \allB. \vi$ 
complying with conditions~(\ref{item:ta:cover}) and~(\ref{item:ta:disjoint}) above.\@
(The definition of
$\TTA{\exists \allx. \vixa}$ or $\TA{\exists \allx. \vixa}$
follows with
$\allB=\emptyset$.)
  Similarly, \TTA{\exists\allB.\vixab} denotes the set of all
  \larat-satisfiable total truth assignments $\muA\wedge\mularat$ that
  propositionally satisfy $\exists\allB.\vixab$, and
  \TA{\exists\allB.\vixab} denotes one set of \larat-satisfiable
  partial truth assignments $\muA\wedge\mularat$ that propositionally satisfy $\exists\allB.\vixab$ complying with conditions~(\ref{item:ta:cover}) and~(\ref{item:ta:disjoint}) above.
   \begin{example}
   Let
 $\vi \defas(A_1 \vee (x\le 1))\wedge  (\neg A_1 \vee B_1) \wedge(B_1\iff 
 (x\ge 2)) $. Then:\\
 $\TTA{\vi}=\set{A_1\wedge B_1\wedge \neg (x\le 1)\wedge (x\ge 2),
                 \neg A_1\wedge  \neg B_1\wedge (x\le 1)\wedge \neg
                 (x\ge 2) }$,\\
 $\TA{\vi}\ \ =\set{A_1\wedge B_1\wedge \phantom{\neg (x\le 1)\wedge\
     \ \!} (x\ge 2),
                 \neg A_1\wedge  \neg B_1\wedge (x\le 1)\wedge \neg
                 (x\ge 2) }$.\\               
 All other total assignments propositionally satisfying $\vi$
 assign both $(x\le 1)$ and $(x\ge 2)$ to true, and are thus
 \larat-unsatisfiable.
 \\We also have that
 $\TTA{\exists x.\exists B_1.\vi}=\TA{\exists x.\exists
   B_1.\vi}=\set{A_1,\neg A_1}$. 
 In fact, \\
 for $\muA=\pos A_1$, let $\mub=\pos B_1$ and $\mularat=\neg (x\le
 1)\wedge (x\ge 2)$;\\
 for $\muA=\neg A_1$, let $\mub=\neg B_1$ and $\mularat=(x\le
 1)\wedge \neg (x\ge 2)$;\\
 in both cases, which correspond to the two elements in \TTA{\vi}, we have that 
 $\muA\wedge\mub\wedge\mularat$ propositionally satisfies $\vi$
 and $\mularat$ is \larat-satisfiable.
 Also:\\
 $\TTA{\exists B_1.\vi}=\set{A_1\wedge \neg (x\le 1)\wedge (x\ge 2),
                 \neg A_1\wedge (x\le 1)\wedge \neg
                 (x\ge 2) }$,\\
 $\TA{\exists B_1.\vi}\ \ \!=\set{A_1\wedge \phantom{\neg (x\le 1)\wedge\
     \ \!} (x\ge 2),
                 \neg A_1\wedge (x\le 1)\wedge \neg
                 (x\ge 2) }$.             
\hfill $\diamond$  
   \end{example}%

\TTA{\ldots}/\TA{\ldots}  can be 
computed efficiently by means of {\emph{Projected AllSMT}}, 
a technique used in formal verification to compute \emph{Predicate Abstraction}~\cite{allsmt}.
All these functionalities are provided by the SMT solver
\mathsat{}~\cite{mathsat5_tacas13}.
(A brief description about how \mathsat{} works when computing
\TTA{\ldots}/\TA{\ldots} is described in~\ref{sec:detailedbackground}.)

\paragraph{\wmipa{}}
\wmipa{} is an efficient WMI algorithm presented in~\cite{morettin-wmi-ijcar17,morettin-wmi-aij19}
which exploits SMT-based projected enumeration.
Let \wxa{} be a \FIUC{} function as above.
\wmipa{} is based on the fact that~\cite{morettin-wmi-aij19}:
\begin{eqnarray} 
\label{eq:wmi3}
\hspace{-.2cm}\WMIviwxa   
\hspace{-.2cm}&=& \hspace{-.9cm}
\sum_{\muastar\in\TTA{\exists \allx. \vistar}} \hspace{-.7cm}
  \WMINBgen{\vistarmuastar}{\wstarmuastar}{\allx}
  \\
  \vistar&\defas&\vi\wedge\supportwff\wedge\bigwedge_{k=1}^{K}(B_k\iff \psi_k)
  \label{eq:wmi4}
  \\
\hspace{-.3cm}
    \WMINBviwx\
\label{eq:wminb2}
  &\!\!=\!\!& \!\!\!\!
      \sum_{\mularat\in\TA{\vi}}
 \myint{\mularat}{\allx}{\w(\allx)}.      
\end{eqnarray}
s.t.
$\allAstar\defas\allA\cup\allB$ s.t.\ 
$\allB\defas\set{B_1,\ldots,B_K}$ are fresh propositional atoms and 
\wstarof{\allx}{\allA\cup\allB} is the weight function obtained by
substituting in \wxa{} each condition
$\psi_k$ in \conditionset{} with $B_k$
(except for these which are 
      Boolean literals on \allA).

The pseudocode of \wmipa{} is reported in Algorithm~\ref{algo:wmipa}.
First, the problem is transformed
by
labelling all conditions in \allpsi{} occurring in \wxa{} with fresh
Boolean atoms \allB, as above. After this preprocessing stage, the set
$\calm^{\allAstar}\defas\TTA{\exists \allx. \vistar}$ is computed by
projected AllSMT~\cite{allsmt}.
  Then, the algorithm iterates over each
  Boolean assignment \muastar{} in $\calm^{\allAstar}$.
\vistarmuastar{} is simplified by the ${\sf Simplify}$
procedure, which propagates truth values and applies
logical and arithmetical simplifications.
Then, if \vistarmuastar{} 
is already a conjunction of literals, 
 the algorithm directly computes its
  contribution to the volume by calling
  $\WMINBgen{\vistarmuastar}{\wstarmuastar}{\allx}$.  Otherwise,
  \TA{\vistarmuastar} is computed by AllSMT to produce partial assignments, and the algorithm iteratively computes
  contributions to the volume for each \mularat.
See~\cite{morettin-wmi-aij19} for more details. 

\begin{remark}%
  \label{remark:enumeration}
 Importantly, in the implementation of \wmipa{} the potentially large sets
$\MUASTAR$ and $\MULARAT$ are not generated
explicitly; rather, their elements are
generated, integrated, and then dropped one-by-one, so as to avoid
space blow-up. For example, steps like ``$ \MUASTAR \leftarrow \TTA{\dots}$; ${\bf for}\ \muastar
  \in \MUASTAR$\ldots'' in Algorithm~\ref{algo:wmipa} should be read as ``${\bf for}\ \muastar
  \in \TTA{\ldots}$\ldots''. (We keep the first representation for readability.)
  The same applies to all our novel algorithms in this
  paper. 
\end{remark}

\begin{algorithm}[t]
\caption{{\sf WMI-PA}$(\vi$, \w$, \allx, \allA)$%
\label{algo:wmipa}}

\begin{algorithmic}[1]
\STATE$\tuple{\vistar, \wstar, \allAstar} \gets  {\sf LabelConditions}(\vi, \w, \allx, \allA)$%

\STATE$\MUASTAR \gets \TTA{\exists \allx. \vistar}$%
\label{alg:wmipa:tta}
\STATE$\vol \gets 0$
\FOR{$\muastar \in \MUASTAR$}
    \STATE${\sf Simplify}(\vistarmuastar)$
    \IF{${\sf LiteralConjunction}(\vistarmuastar)$}
    \STATE$\vol \gets \vol + \WMINBgen{\vistarmuastar}{\wstarmuastar}{\allx}$
    \ELSE%
    \STATE$\MULARAT{} \gets  \TA{\vistarmuastar}$ 
    \FOR{$\mularat{} \in \MULARAT$}
       \STATE$\vol \gets \vol + \WMINBgen{\mularat}{\wstarmuastar}{\allx}$
    \ENDFOR%
    \ENDIF%
\ENDFOR%
\RETURN$\vol$
\end{algorithmic}
\end{algorithm}

\subsection{Hybrid Probabilistic Inference via WMI}
The main application of WMI is marginal inference on weighted SMT
theories. Similarly to WMC, inference can be reduced to the computation of two weighted model integrals:
\begin{align}
  \Pr(\Delta \:|\: \supportwff, w) = \frac{\WMI(\Delta \land \supportwff, w)}{\WMI(\supportwff, w)}
\end{align}
The denominator above is akin to computing the partition function on
probabilistic models with unnormalized factors.
Crucially, the formulas $\Delta$ and $\supportwff$ are arbitrary,
possibly encoding complex non-convex regions of a hybrid space. This
goes beyond the kind of queries that are supported by more traditional
algorithms like Belief Propagation, being particularly beneficial in
contexts where it is necessary to compute probabilities of complex
properties, like those arising in (probabilistic) formal verification
domains. Furthermore, the use of specialized software to deal with constraints yields state-of-the-art results on standard queries when
the support of the distribution is highly structured, as shown by the
competitive results obtained by reducing inference on discrete models
to WMC~\citep{sang2005performing}.

\section{Analysis of State-Of-The-Art WMI Techniques}%
\label{sec:wmipa-issues}

We start by presenting an analysis of the KC and \wmipa{} approaches, which represent the current state-of-the-art in WMI.\@

\subsection{Knowledge Compilation}%
\label{sec:dd_issues}

\begin{figure}
  \centering
  \includegraphics[width=0.95\textwidth]{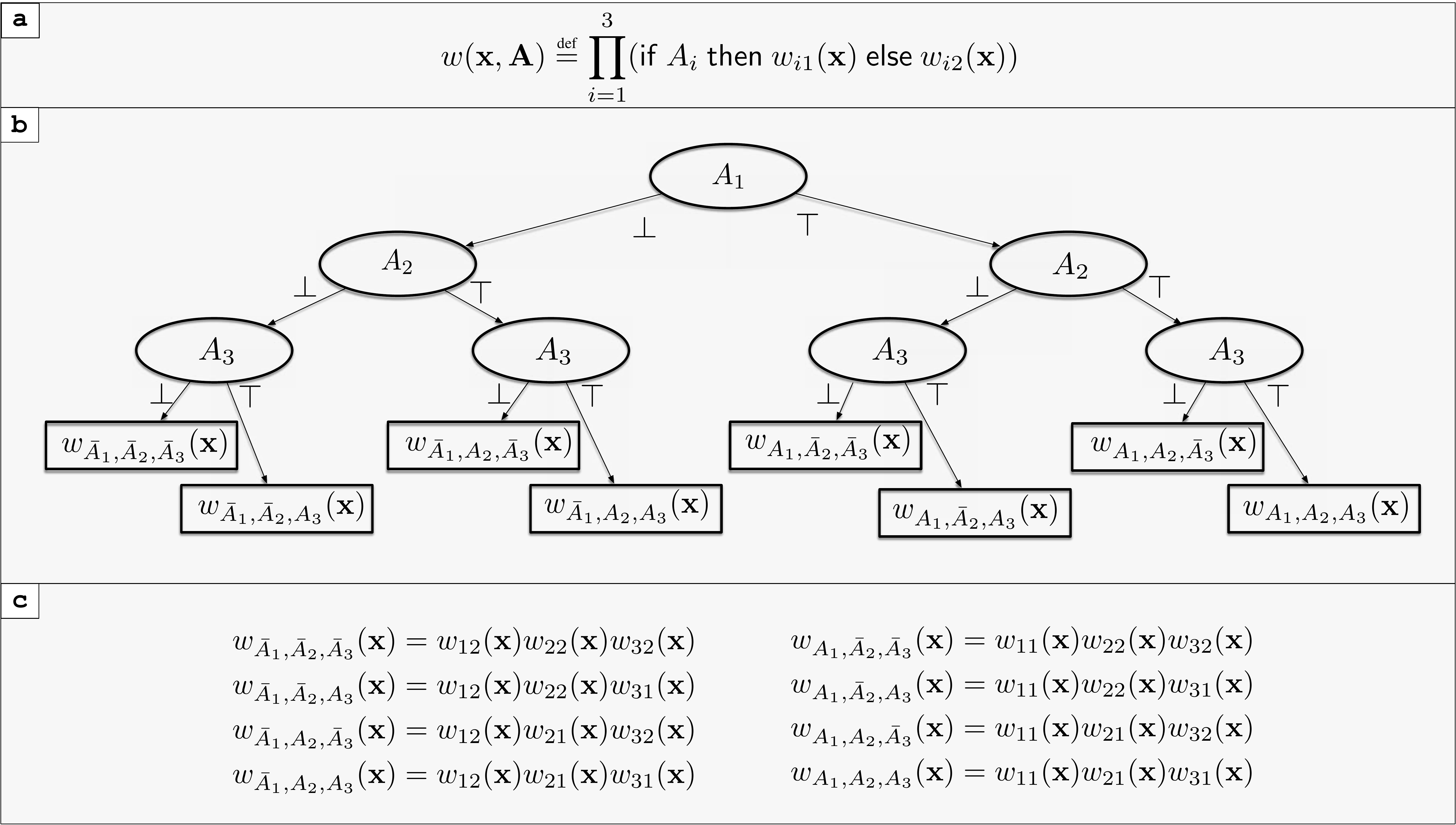}
  \caption{\label{fig:ex_dd_issues} Example highlighting the
    efficiency issues of the knowledge compilation algorithms for WMI.\@ ({\bf a}) definition of a weight function consisting of a product of conditional statements. 
   ({\bf b}) decision diagram generated by knowledge compilation approaches over the weight function. Round nodes
    indicate if-then-else conditions, with true and false cases on the
    right and left outgoing edges respectively. Squared nodes indicate
    \FI{} weight functions. Note how the diagram has a number of
    branches which is exponential in the number of conditions, and no
    compression is achieved. ({\bf c}) definition of the weight
    functions at the leaves of the diagram. In naming weight
    functions, we use $\bar{A}$ for $\lnot A$ for the sake of
    compactness.}
  %
    \centering
    \includegraphics[width=.63\textwidth]{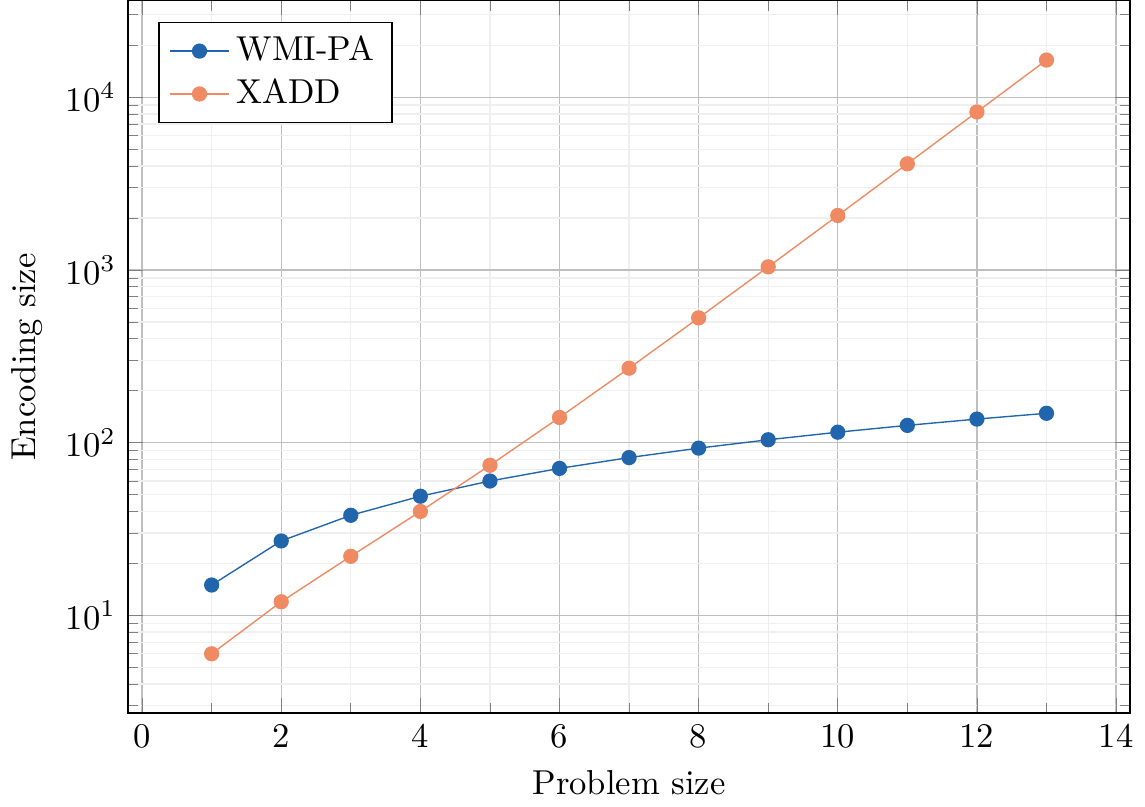}
    \caption{Size of XADD diagram (number of nodes) and \wmipa{} formula as the problem size increases, in logarithmic scale. Whereas the size of the diagram grows exponentially,
    \wmipa{} encodes the problem into a formula of linear size.
  }%
  \label{fig:xasdd}
\end{figure}

We start our analysis 
by noticing a major problem
with existing KC approaches for
WMI~\citep{dos2019exact,kolb2020exploit}, in that they can easily 
blow up in space even with simple weight functions.  Consider, e.g.,
the case in which
\begin{eqnarray}
  \label{eq:prodite}
  \wxa\defas\prod_{i=1}^N ({\sf If}\ \psi_i\ {\sf Then}\ w_{i1}(\allx)\ {\sf Else}\ w_{i2}(\allx))
\end{eqnarray}
where the $\psi_i$s are \larat{} conditions on \set{\allx,\allA} and the
$w_{i1},w_{i2}$ are generic functions on $\allx$.
First, the decision diagrams do not interleave arithmetic and
conditional operators; rather, they push all the arithmetic operators
below the conditional ones.    
Thus, with~\eqref{eq:prodite} the resulting decision diagrams consist of $2^N$ branches on the $\psi_i$s, each corresponding to a distinct unconditioned weight function of the form $\prod_{i=1}^N w_{i{j_i}}(\allx)$ s.t.\ $j_i\in\set{1,2}$.
{See Figure~\ref{fig:ex_dd_issues} for an example for $N=3$ and
$\psi_i\defas A_i$, $i\in\set{1,2,3}$.}
%
Second, the decision diagrams are built on the Boolean abstraction of
\wxa, s.t.\ they do not eliminate a priori the useless
branches consisting of \larat-unsatisfiable combinations of $\psi_i$s, which can be up to exponentially many. 

With \wmipa{}, instead, 
the representation of~\eqref{eq:prodite} does not grow in size,
because \FIUC{} functions allow for interleaving arithmetic and conditional
operators. Also, the SMT-based enumeration algorithm does not generate
\larat-unsatisfiable assignments on the $\psi_i$s. 
%

This fact has been empirically confirmed and is shown in
Figure~\ref{fig:xasdd}, 
where we plot in logarithmic scale the number of nodes of a KC-based
encoding of~\eqref{eq:prodite} using XADD
for problems of increasing size, compared with the size of the encoding
used by \wmipa{}. This graph clearly shows the exponential
blow-up of XADD size, whereas the size of the encoding
used by \wmipa{} grows linearly.
We stress the fact that~\eqref{eq:prodite} is not an
artificial scenario: rather, e.g., this is the case of the real-world
logistics
problems in~\cite{morettin-wmi-aij19}. 

\subsection{\wmipa}

\begin{figure}[th]
  \centering
  \includegraphics[width=\textwidth]{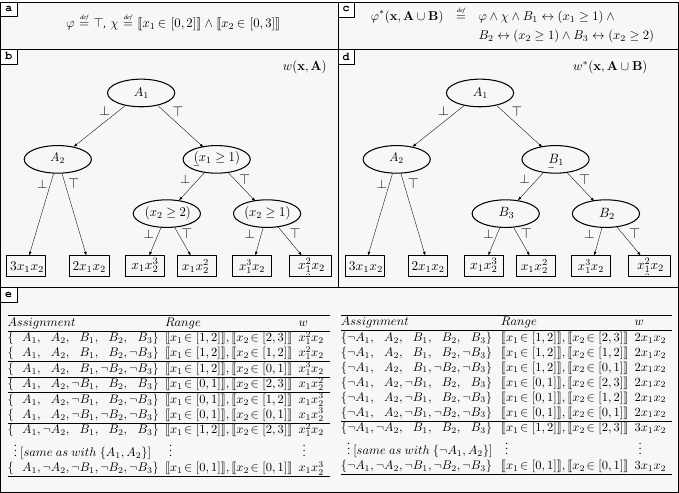}
  \caption{\label{fig:ex_pa_issues}
    Example highlighting the
    efficiency issues of the \wmipa{} algorithm. ({\bf a}) definition
    of formula $\vi$ (trivially true) and support $\supportwff$. ({\bf
      b}) definition of the weight function $\wxa$. Round nodes
    indicate if-then-else conditions, with true and false cases on the
    right and left outgoing edges respectively. Squared nodes indicate
    \FI{} weight functions. ({\bf c}) novel version of the formula
    $\vistarof{\allx}{\allA\cup\allB}$ after the application of the
    ${\sf LabelConditions(\ldots)}$ step of \wmipa{}. ({\bf d}) novel
    version of the weight function $\wstarof{\allx}{\allA\cup\allB}$,
    where all \larat{} conditions have been replaced with the fresh
    Boolean atoms introduced in
    $\vistarof{\allx}{\allA\cup\allB}$. ({\bf e}) List of assignments
    obtained by \wmipa{} on $\allA\cup\allB$ (split on two columns for the sake of compactness). Notice the amount of
    assignments sharing the same \FI{} weight function.}
\end{figure}

We continue our analysis by noticing a major problem also for the
\wmipa{} algorithm: 
unlike with the KC approaches, \emph{it fails to leverage the conditional
 structure of the weight function to prune the set of models to
integrate over}.
We illustrate the issue by means of a simple example
(see Figure~\ref{fig:ex_pa_issues}).

\begin{example}%
\label{ex:issue1}
Let $\vi\defas\top$,
$\supportwff\defas\inside{x_1}{[0,2]}\wedge\inside{x_2}{[0,3]}$ (Figure~\ref{fig:ex_pa_issues}(a)) and let $\wxa$ be a tree-structured weight function defined as in Figure~\ref{fig:ex_pa_issues}(b).
To compute $\WMIgen{\vi\wedge\supportwff}{\w}{\allx}{\allA}$, only six integrals need to be computed:\\
\noindent
\phantom{1}$\myf{11}$ on $\inside{x_1}{[1,2]}\wedge\inside{x_2}{[1,3]}$ (if $A_1=\top$)\\
\phantom{1}$\myf{12}$ on $\inside{x_1}{[1,2]}\wedge\inside{x_2}{[0,1]}$ (if $A_1=\top$)\\
\phantom{1}$\myf{21}$ on $\inside{x_1}{[0,1]}\wedge\inside{x_2}{[2,3]}$ (if $A_1=\top$)\\
\phantom{1}$\myf{22}$ on $\inside{x_1}{[0,1]}\wedge\inside{x_2}{[0,2]}$ (if $A_1=\top$)\\
$\myf{3}$ on $\inside{x_1}{[0,2]}\wedge\inside{x_2}{[0,3]}$ (if $A_1=\bot,A_2=\top$)\\
$\myf{4}$ on $\inside{x_1}{[0,2]}\wedge\inside{x_2}{[0,3]}$ (if $A_1=\bot,A_2=\bot$)\\
\noi When \wmipa{} is used (Algorithm~\ref{algo:wmipa}), by applying ${\sf
  LabelConditions(\ldots)}$ we obtain
$\vistarof{\allx}{\allA\cup\allB} \defas \vi\wedge\supportwff\wedge
B_1\iff(x_1\ge 1)\wedge
 B_2\iff(x_2\ge 1)\wedge
 B_3\iff(x_2\ge 2)$
  (Figure~\ref{fig:ex_pa_issues}(c)),
and the weight function $\wstarof{\allx}{\allA\cup\allB}$ shown in
Figure~\ref{fig:ex_pa_issues}(d). Then, by applying $\TTA{\exists \allx.\vistar}$
(line 2)
we obtain 24 total
assignments $\MUASTAR{}$ on $\allA\cup\allB$, as shown in
Figure~\ref{fig:ex_pa_issues}(e).
{(Notice that all
  assignments containing $\set{\neg B_2, B_3}$, and hence
  $\set{\neg(x_2\ge 1), (x_2\ge 2)}$,
  are missing because they are
  \larat{}-unsatisfiable.%
)}
{As a result,
  \wmipa{}  computes 24 integrals instead of 6.} 
In particular,
\wmipa{} computes twice each of the first 6 
integrals, for \set{A_1,A_2,\ldots} and  \set{A_1,\neg  A_2,\ldots};~%
%
also, it 
splits into 2
parts each the integrals of $\myf{11}$ and $\myf{22}$ 
{(on the irrelevant truth values of $B_3$ and $B_2$ respectively)}
and into 6
parts each the integrals of $\myf{3}$ and  $\myf{4}$
{(on the 8 irrelevant truth
values of $B_1,B_2,B_3$ minus the 2 \larat-unsatisfiable ones containing
$\set{\neg B_2, B_3}$)}.
\hfill $\diamond$
\end{example}

The key issue about \wmipa{} is that the enumeration
of \TTA{\exists\allx. \vistar} in~\eqref{eq:wmi3}
(line~\ref{alg:wmipa:tta} 
in Algorithm~\ref{algo:wmipa})
\emph{is not aware of the conditional structure of the weight function
  \wstar{}},
in particular, it is not aware of the fact that often
\emph{partial}
assignments to the set of conditions in \wstar{} (both in \allA{} and
\allB, i.e., both original Boolean and  \larat{} conditions in \w{})
are sufficient to identify the
value of \wstar{},
so that it is forced
to  enumerate all \larat-satisfiable total assignments extending them.
(E.g., in Example~\ref{ex:issue1}, \set{ A_1,B_1,B_2} suffices to identify $\myf{11}$, regardless
the values of $A_2$ and $B_3$, but \wmipa{} enumerates
\set{A_1,A_2,B_1,B_2,B_3},
\set{A_1,A_2,B_1,B_2,\neg B_3},
\set{A_1,\neg A_2,B_1,B_2,B_3},
\set{A_1,\neg A_2,B_1,B_2,\neg B_3}.)
This has two consequences.

First, to make the enumerator split also on the conditions
in $\conditionset$, \wmipa{} renames them with fresh Boolean atoms
\allB{},
conjoining $\bigwedge_{k=1}^K(B_k\iff \psi_k)$ to
$\vi\wedge\supportwff$  (line 1).
Unfortunately, the above equivalences force the enumerator to assign a truth value to \emph{all} the
$B_i$s in \allB{}  (hence, to all \larat-conditions $\psi_i$s in $\conditionset{}\setminus\allA$) in every
assignment, even when the conditional structure of \w{} does not need
it. This is what forces the unnecessary partitioning of integrals into
subregions.
(E.g., in Example~\ref{ex:issue1}, the integral of $x_1^2x_2$ is
unnecessarily split into two integrals for $(x_2\ge 2)$ and $\neg(x_2\ge
2)$ due to the unnecessary branch on $\set{B_3,\neg B_3}$.)

Second, not knowing \wstar{}, the enumerator is forced to always
assign also \emph{all} original Boolean atoms in \allA,
even when the combination $\mua\cup\mub$ of a \emph{partial}
assignment $\mua$ on \allA{} and a total assignment \mub{} on \allB{}
suffices to satisfy \vistar{} and to make \wstar{} \FI{},
that is, 
$\TTA{\exists \allx. \vistar}$ is used instead of $\TA{\exists
  \allx. \vistar}$ in~\eqref{eq:wmi3} and in line~\ref{alg:wmipa:tta} of Algorithm~\ref{algo:wmipa}.
This is what causes the unnecessary duplication of integrals. 
(E.g., in Example~\ref{ex:issue1}, 
 each of the first 6 
integrals is computed twice, for $\set{A_1,A_2,\ldots}$ and  $\set{A_1,\neg
  A_2,\ldots}$, due to the unnecessary branching on $\set{A_2,\neg A_2}$.)
Although the latter problem could in principle be fixed by caching all
the values of   the integrals, this could be expensive in terms of both time and memory.

To cope with these issue, we need to modify \wmipa{} to
make it aware of the conditional structure of \w{}. 
%
{(One further issue, dealing with the fact that the
  conditions in \conditionset{} could not be literals, will be
  addressed in \sref{sec:nonatomic-conditions}.)}

\section{Making \wmipa{} Weight-Structure Aware}%
\label{sec:wmipa-improved}

In order to address the issues described in \sref{sec:wmipa-issues},
we aim to combine the best of KC approaches ---i.e., weight-structure
awareness--- with the best of PA-based approaches ---i.e., SMT-based enumeration--- by introducing
weight-structure awareness into PA-based WMI.\@
In \sref{sec:skeleton} we present the general idea for  making \wmipa{}
weight-structure aware,
by introducing and exploiting the notion of \emph{conditional skeleton} for \w{}.
%
%
%
%
%
A first preliminary approach and algorithm (namely ``\wmisapa{}''),
based on an implicit enumeration of a skeleton, was proposed
in~\cite{spallitta2022smt}.  For the sake of compactness, we do not
report it here; a detailed description is reported
in~\ref{sec:implicitskeleton}.  In \sref{sec:explicitskeleton} we
propose our novel and current best approach and algorithm (namely
``Structure-Aware Enumeration for WMI'', in short ``\method{}''),
which is both much simpler and more effective than that
from~\cite{spallitta2022smt}, in which the skeleton is generated
explicitly.

\subsection{General Idea: Exploiting a Conditional
    Skeleton of \w{}}%
    \label{sec:skeleton}
%
We start by introducing the following concept.

\begin{definition}[Conditional skeleton of \w{}, \skw{}]%
  \label{def:skeleton}
  {Let \vi{}, \supportwff{} be as 
  in~\sref{sec:background-wmi}%
;}
  let \wxa{} be \FIUC{} on the set of conditions $\conditionset$.
  We call a {\bf Conditional Skeleton of \w{}}, written \skw{}, any
  \larat{} formula s.t.:
  \begin{enumerate}[(a)]
  \item\label{item:skw:a}  its atoms are all and only {those occurring in} the conditions in $\conditionset$;
  \item\label{item:skw:b} \skw{} is 
  valid, so that $\vi\wedge\supportwff$ is 
  equivalent to
    $\vi\wedge\supportwff\wedge\skw{}$;
%
  \item\label{item:skw:c}  every \emph{partial} truth value assignment $\mu$ to (the
    atoms occurring in) the conditions
$\conditionset$ which makes \skw{} true is such that \wmuagen{\mu} is \FI{}.
  \end{enumerate}
\end{definition}

\begin{example}%
\label{ex:encsk}
Consider the problem in Example~\ref{ex:issue1}.
Then the following formula is a conditional skeleton \skw{} for \wxa{}:

\noindent
$\def\arraycolsep{2.5pt}
\begin{array}{ll|l}
  (\br{A_1}) & \land & \mbox{split on $A_1$}\\
   \big(\neg A_1 \vee \br{(x_1 \ge 1)}\big) & \land & \mbox{if $A_1$ split on $(x_1 \ge 1)$}\\
  \big(\neg A_1 \vee \neg (x_1 \ge 1) \vee \br{(x_2\ge 1)}\big) & \land & \mbox{if $A_1,\pos(x_1 \ge 1)$ split on $(x_2 \ge 1)$}\\
  \big(\neg A_1 \vee \pos (x_1 \ge 1) \vee \br{(x_2\ge 2)}\big) & \land & \mbox{if $A_1,\neg(x_1 \ge 1)$ split on $(x_2 \ge 2)$}\\
  (\pos A_1 \vee \br{A_2}) & \land & \mbox{if $\neg A_1$ split on $A_2$}
                             \\
\end{array}
$

\noindent E.g., the partial assignment
  $\mu\defas\set{A_1, (x_1\ge 1), (x_2\ge 1)}$ propositionally satisfies \skw{}, and is
  such that,
   in
  Example~\ref{ex:issue1},
  $\wmuagen{\mu}=\myf{11}$,
  which is \FI{}. $\hfill \diamond$
\end{example}

Assume we have produced one such formula \skw{}.
%
Unlike with \wmipa{}, 
we do \emph{not} rename with \allB{} the conditions \conditionset{} in \wxa{},
and we conjoin to $\vi\wedge\supportwff$ the skeleton \skw{} instead of $\bigwedge_{k=1}^K(B_k\iff \psi_k)$.
Unlike with $\bigwedge_{k=1}^K(B_k\iff
\psi_k)$, which needs \emph{total} assignments to
\conditionset{} (i.e., to \allB) to be satisfied, for \skw{} it suffices to produce \emph{partial}
assignments to \conditionset{} which verify condition~(\ref{item:skw:c}), i.e.,
those assignments $\mu$ s.t. \wmuagen{\mu} is \FI{}.
Thus, we can rewrite~\eqref{eq:wmi3}--\eqref{eq:wmi4} into:
\begin{eqnarray}
\label{eq:newwmi1}
\hspace{-.5cm}\WMIviwxa   
\hspace{-.3cm}&=& 
\sum_{\mu\in\TA{\viencsk}}
2^{|\allA\setminus\mu|}\cdot
  \WMINBgen{\viencskmu}{\wmuagen{\mu}}{\allx}
  \\
  \viencsk&\defas&\vi\wedge\supportwff\wedge\skw{}
  \label{eq:newwmi2}
\end{eqnarray}
\noindent
(Notice that in~\eqref{eq:newwmi1} $\mu$ is in $\TA{\viencsk}$, not
$\TTA{\viencsk}$, that is, we enumerate \emph{partial}
assignments $\mu$ for $\viencsk$ and, in particular, \emph{partial} assignments for \conditionset.)

Condition~(\ref{item:skw:c}) in Definition~\ref{def:skeleton} guarantees that
$\WMINBgen{\viencskmu}{\wmuagen{\mu}}{\allx}$ in~\eqref{eq:newwmi1}
can be directly computed even though $\mu$ is partial, without further
partitioning due to irrelevant conditions.
The $2^{|\allA\setminus\mu|}$ factor in~\eqref{eq:newwmi1}, where $|\allA\!\setminus\!\mu|$ is the number of Boolean atoms in \allA{}
that are not assigned by $\mu$, resembles
the fact that, if 
$A_i\in\allA$ is not assigned in 
$\mu$,
then $\WMINBgen{\viencskmu}{\wmuagen{\mu}}{\allx}$ must be counted
twice, because $\mu$ represents two assignments $\mu\cup\set{A_i}$ and
$\mu\cup\set{\neg A_i}$ which would produce two identical integrals,
because their \larat{} component and the branch of the weight function they identify are identical.
When $|\allA\!\setminus\!\mu|>0$, this allows to compute only one
integral rather than $2^{|\allA\setminus\mu|}$ ones (!).

Notice that the latter is only one of the
computational advantages of~\eqref{eq:newwmi1}-\eqref{eq:newwmi2}
w.r.t.\ \eqref{eq:wmi3}-\eqref{eq:wmi4}: even more importantly,~\eqref{eq:newwmi1}-\eqref{eq:newwmi2} allow for enumerating partial
assignments also on the \larat-conditions in $\conditionset\setminus \allA$, reducing the number of
integrals to compute for each partial assignment $\mu$ by up to a $2^j$ factor, $j$ being
the number of unassigned \larat{} atoms in $\mu$. (In this case, no
multiplication factor should be considered, because an unassigned
\larat{} condition $\psi_i$ in $\mu$ causes the merging of the two
integration domains of 
 $\mu\cup\set{\psi_i}$ and $\mu\cup\set{\neg\psi_i}$.)
  \begin{example}%
  \label{ex:skenumeration}
  Consider the problem in Example~\ref{ex:issue1} and the
  corresponding \skwenc{} formula in Example~\ref{ex:encsk}.
  Let $\viencsk\defas\vi\wedge\supportwff\wedge\skwenc{}$,
  as in~\eqref{eq:newwmi2}.
  We show how 
  $\TA{\viencsk}$ in~\eqref{eq:newwmi1} can be produced by the SMT solver. 
  Assume nondeterministic choices are picked
  following the order $A_1, A_2, (x_1 \ge 1),(x_2 \ge 1),(x_2 \ge 2)$,~%
  assigning positive values first.
  Then in the first branch the following satisfying \emph{total} truth
  assignment is generated:
%
\newcommand{\supportass}{\set{(x_1\ge 0),(x_1\le 2),(x_2\ge 0),(x_2\le 3)}}
\renewcommand{\und}{}
\[
  \supportass\cup
 \{\und{A_1}, \und{A_2}, \und{(x_1\ge 1)}, \und{(x_2\ge 1)}, \und{(x_2\ge 2)}\}
\]
\noindent (where $\set{(x_1\ge 0),(x_1\le 2),(x_2\ge 0),(x_2\le 3)}$ is assigned deterministically to satisfy the
$\vi\wedge\supportwff{}$ part) from which the solver extracts the {minimal} partial truth assignment which evaluates \skwenc{} to true:
  \[
  \begin{array}{l}\mu_1\defas\supportass\cup
   \{\und{A_1}, \und{(x_1\ge 1)}, \und{(x_2\ge 1)}\}.
  \end{array}
  \]
In the next branch, the following minimal partial assignment is produced:
  \[
  \begin{array}{l}\mu_2\defas\supportass\cup
    \{\und{A_1}, \und{(x_1\ge 1)}, \neg(x_2\ge 1)\}.
  \end{array}
  \]
  s.t.\ $\mu_1,\mu_2$ assign opposite truth value to (at least) one atom.
\noindent Overall, the algorithm enumerates the following collection
of minimal partial assignments:\\
$
\begin{array}{ll}
\supportass\cup\set{\pos A_1,\pos (x_1\ge 1), \pos (x_2\ge 1)},\\
\supportass\cup\set{\pos A_1,\pos (x_1\ge 1), \neg (x_2\ge 1)},\\
\supportass\cup\set{\pos A_1,\neg (x_1\ge 1), \pos (x_2\ge 2)},\\
\supportass\cup\set{\pos A_1,\neg (x_1\ge 1), \neg (x_2\ge 2)},\\
\supportass\cup\set{\neg A_1,\pos A_2},\\
\supportass\cup\set{\neg A_1,\neg A_2}\\
\end{array}
$\\
\noindent
which correspond to the six integrals of Example~\ref{ex:issue1}.
Notice that according to~\eqref{eq:newwmi1} the first four
integrals have to be multiplied by 2, because the partial assignment
\set{A_1} covers two total assignments \set{A_1,A_2} and \set{A_1,\neg A_2}. \hfill $\diamond$
\end{example}

Notice that logic-wise \skw{} is non-informative because it is a
valid formula (condition~(\ref{item:skw:b}) in Definition~\ref{def:skeleton}).
Nevertheless, the role of \skw{} is not only to
``make  the 
enumerator aware of the presence of the conditions
$\conditionset{}$'' ---like the $\bigwedge_{k=1}^K(B_k\iff\psi_k)$ in \wmipa{}---
but also to 
mimic the conditional structure of $\w$,
forcing every assignment $\mu$ 
to assign truth values to all and only those conditions in \allpsi{}
which are necessary to make
$\wmuagen{\mu}$ \FI{}, and hence make
$\WMINBgen{\viencskmu}{\wmuagen{\mu}}{\allx}$ directly computable,
without further partitioning.

In principle, we could use as \skw{} (a \larat{} formula encoding
the conditional structure of) an XADD or (F)XSDD, which do not suffer
from the lack of structure awareness of \wmipa{}.
However,  this would 
cause a blow-up in size, as discussed in~\sref{sec:dd_issues}.
%
To this extent,  avoiding \skw{}
blow up in size is a key issue  in our approach, which we will discuss
in the following steps.

\subsection{Encoding a Conditional Skeleton Explicitly}
\label{sec:explicitskeleton}
%


\subsubsection{Encoding}
\label{sec:explicitskeleton-encoding}
\begin{algorithm}[t]
  \caption{\label{alg:convertsk}
    \convertsk{}($\term$, $\conds{}$)\ \ // \conds{} is a set of literal conditions
}
\begin{algorithmic}[1]
  \IF{(\{$\term$ is a constant or a polynomial\})}\label{alg:convertsk-line1}
  \RETURN $\top$ \label{alg:convertsk-line2}
  \ENDIF
  \IF{($\term==(\term_1 \bowtie \term_2)$, $\bowtie\ \in\set{+,-,\cdot}$)} \label{alg:convertsk-line3}
  \RETURN \convertsk{}($\term_1$, \conds{})\ $\wedge$  \convertsk{}($\term_2$, \conds{}) \label{alg:convertsk-line4}
  \ENDIF%
  {\IF{($\term==g(\term_1,\dots,\term_k)$, $g$ unconditioned)} \label{alg:convertsk-startf}
  \RETURN $\bigwedge_{i=1}^{k}\convertsk{}(\term_i, \conds{})$\label{alg:convertsk-endf}
  \ENDIF}
  \IF{($\term==({\sf If}\ \psi\ {\sf Then}\ \term_{1}\ {\sf Else}\ {\term_{2}})$)} \label{alg:convertsk-line5}
  \IF {$\psi$ is a literal}
  \STATE $\branch{} \gets \bigvee_{\psi_i\in\conds}\neg\psi_i \vee \psi \vee \neg\psi$ \label{alg:convertsk-line9}
  \STATE $\defs_1\gets\convertsk{}({\term_1, \conds{} \cup \set{\psi}})$ \label{alg:convertsk-line7}
  \STATE $\defs_2\gets\convertsk{}({\term_2, \conds{} \cup \set{\neg \psi}})$ \label{alg:convertsk-line8}
  \RETURN
   $\branch{} \land\defs{}_1 \land\defs{}_2  $ \label{alg:convertsk-line10}
  \ELSE 
  \STATE {\bf let} $B$ be a fresh Boolean
  atom \label{alg:convertsk-line12}   
\STATE
$\branch{} \gets%
  \begin{array}{l}
  (\bigvee_{\psi_i \in \conds} \neg \psi_i
    \vee B \vee \neg B )\ \wedge \\
   \shortcut{\bigvee_{\psi_i \in \conds} \neg
    \psi_i\vee \CNFplaisted{B\iff\psi}}    
  \end{array}
  $
  \STATE $\defs{}_1 \gets \convertsk{}({\term_1, \conds{} \cup \set{B}})$
  \STATE $\defs{}_2 \gets \convertsk{}({\term_2, \conds{} \cup \set{\neg B}})$
  \RETURN  
  $\branch{} \land \defs{}_1 \land \defs{}_2$  \label{alg:convertsk-line16}
  \ENDIF
  \ENDIF
\end{algorithmic}
\label{algo:convertsk}
\end{algorithm}

Algorithm~\ref{algo:convertsk} shows the procedure for
building explicitly a formula \skw{}. 
Intuitively, the output consists of a conjunction of formulas, each stating
``\emph{if all the conditions $\psi_1,..,\psi_i$ of the current sub-branch in \w{} hold, then split on the
  next condition $\psi_{i+1}$ in the branch}''.
  For example, for the problem in Example~\ref{ex:issue1},
 Algorithm~\ref{algo:convertsk}
produces as \skwenc{} the formula reported in
Example~\ref{ex:encsk}, whose satisfying partial assignments
  can be enumerated as in Example~\ref{ex:skenumeration}.

%
%
The recursive procedure \convertsk{}($\w_i$, $\conds{}$) takes as input the current subtree $\w_i$ of the weight function \w{} and the set of literals $\conds$ representing the conditions under whose scope $\w_i$ occurs, and returns the CNF representation of
{$\bigwedge_{\psi_i\in\conds}\psi_i\imp\skwencwi{}$ (i.e., it returns 
$\shortcut{(\bigvee_{\psi_i\in\conds}\neg\psi_i)\vee\ \skwencwi{}}$). Hence, \convertsk{}($\w$, $\emptyset{}$) returns \skwenc{}.}
 Notice that the set $\conds$ contains the conditions that need to be
true in order to activate the current branch.

We first focus on lines~\ref{alg:convertsk-line1}-\ref{alg:convertsk-line10}, in which we assume
that the weight conditions $\psi_i$ are literals (the behaviour in case of non-literal conditions, lines~\ref{alg:convertsk-line12}-\ref{alg:convertsk-line16}, will be explained in \sref{sec:nonatomic-conditions}).
The recursive encoding of \skwenc{} works as follows:
\begin{itemize}
  \item[(Lines~\ref{alg:convertsk-line1}-\ref{alg:convertsk-line2}):] A constant or a polynomial does not contain weight
    conditions. Thus, we can simply encode it with $\top$. Notice that here $\conds$ has no role, because $(\bigwedge_{\psi_i\in \conds}\psi_i) \imp \top$ reduces to $\top$. 
  
  \item[(Lines~\ref{alg:convertsk-line3}-\ref{alg:convertsk-line4}):] A term representing an arithmetic operator $\term_1 \bowtie \term_2$, with $\bowtie\ \in \{+, -, \cdot\}$
    must ensure that the conditions in both branches $\term_1,\term_2$
    are enumerated. Thus, we encode it by conjoining the results of the conversion procedure on $\term_1$ and $\term_2$.
      \item[(Lines~\ref{alg:convertsk-startf}-\ref{alg:convertsk-endf}):]Similarly, a term representing an unconditioned function in the form $g (\term_1,\dots,\term_k)$, must ensure that the conditions in all the branches $\term_1\dots\term_k$ are enumerated.
      Thus, we encode it by conjoining the results of the conversion procedure on $\term_1,\dots,\term_k$.
    
  \item[(Lines~\ref{alg:convertsk-line5}-\ref{alg:convertsk-line10}):] When encoding a conditional term in the form
    $ (\Ite{\psi}{\w_1}{\w_2})$, if $\psi$ is a literal,  we
    have the following:
    \begin{itemize}
      \item[(Line~\ref{alg:convertsk-line9}):] When  all \conds{} are true, then
        $\psi$ must be split upon. This fact is encoded by the
        valid clause:%
        \begin{eqnarray}
          \label{eq:branch}
          \textstyle
          \bigvee_{\psi_i \in \conds}{\neg \psi_i} \lor \psi \lor \neg \psi.
        \end{eqnarray}
      \item[(Line~\ref{alg:convertsk-line7}):] When (all \conds{} and
        are true and) $\psi$ is true, all the branches of $\w_1$ must
        be enumerated. This is encoded by recursively calling the
        conversion procedure on $w_1$ adding $\psi$ to the conditions
        \conds{} that need to be true to activate the branch of $w_1$,
        {which returns $\shortcut{ \bigvee_{\psi_i \in \conds}{\neg
            \psi_i} \lor \neg \psi \lor \skwencwone{}}$}.
          
      \item[(Line~\ref{alg:convertsk-line8}):] Similarly, when (all \conds{} are true and) $\psi$ is false, all the branches of $\w_2$ must be enumerated. This is encoded by recursively calling the conversion procedure on $w_2$ adding $\neg \psi$ to \conds{},
        {which returns $\shortcut{ \bigvee_{\psi_i \in \conds}{\neg
            \psi_i} \lor \psi \lor \skwencwtwo{}}$}.
    \end{itemize}
\end{itemize}

  \noindent
  Unlike with KC approaches,
  it is easy to see that \skw{} grows linearly in size
  w.r.t.~\w{}. E.g., with~\eqref{eq:prodite}, 
  $\encSK{}$ is
  $\bigwedge_{i=1}^N (\psi_i\vee\neg\psi_i)$, whose size grows linearly
 rather than exponentially w.r.t.~$N$.

\subsubsection{Dealing with non-literal conditions}%
\label{sec:nonatomic-conditions}
%
%

State-of-the-art SMT solvers, e.g.,
\mathsat{}~\cite{mathsat5_tacas13}, deal very efficiently with
formulas expressed in CNF.\@ Hence, if the input formula is not in
CNF, a CNF-ization step is typically performed before the solving
process.\@

  We first recall some basic notions about CNF-ization (see~\ref{sec:detailed-cnfization} for details.)
     A formula $\vixa$ can be converted into an 
     equivalent one (``\CNF{\vi}'') by recursively
     applying some rewriting rules; unfortunately the size of
     \CNF{\vi}  
     can be exponential w.r.t.\ that of $\vi$.
A typically better choice is 
  \emph{Tseitin CNF-ization}~\cite{tseitin68}  (``\CNFtseitin{\vi}''), which consists in applying recursively
    the ``labelling'' rewrite rule $\vi\thus \vi[\psi|B]\wedge
    (B\iff \psi)$ ---$\vi[\psi|B]$ being the results of substituting
    all occurrences of
    a subformula $\psi$ with a fresh Boolean atom $B$--- until
    all conjuncts can be CNF-ized classically without space blow-up.
    Alternatively, one can apply the rule $\vi\thus \vi[\psi|B]\wedge
    (B\imp \psi)$
    under some restrictions  (``\CNFplaisted{\vi}'')~\cite{plaisted1986structure}.
%
    With both cases, the resulting formula $\phi(\allx,\allA\cup\allB)$
    is s.t.\ $\vixa$ is 
    equivalent to $\exists
    \allB.\phi(\allx,\allA\cup\allB)$, 
%
    \allB being the set of fresh
    atoms introduced, and the size of $\phi$ is linear w.r.t.\ that of
    $\vi$. Thus
\TTA{\vi}, \TA{\vi}, \TTA{\exists \allx.\vi} and \TA{\exists
  \allx.\vi} can be computed as
\TTA{\exists \allB. \phi}, \TA{\exists \allB. \phi}, \TTA{\exists
  \allx.\exists \allB. \phi } and \TA{\exists \allx.\exists \allB. \phi},  respectively.


 Algorithm \ref{alg:convertsk} in
 \sref{sec:explicitskeleton-encoding} assumes that the
conditions in \allpsi{} are single literals, either Boolean  or
\larat{} ones. 
In general, this is not necessarily the case, for instance it is common to have 
\emph{range conditions} in the form $(x_i \ge l) \land (x_i \le u)$ or even more complex
conditions. In these cases, the skeleton formula is not in
CNF form.
In fact, $\CNF{\bigwedge_i \psi_i \imp \varphi}$ 
can be straightforwardly computed out of $\CNF{\varphi}$ only if the $\psi_i$s are literals,
because this reduces to augment each clause in $\CNF{\varphi}$ with $\bigvee_i\neg\psi_i$.
If this is not the case, then the CNF-ization either may cause a
blow-up in size if equivalence-preserving CNF-ization is applied,
or it may require some
variant of Tseitin CNF-ization~\cite{tseitin68}.
%
\SMT{} solvers typically rely on the second option for
satisfiability checks.
For what enumeration is concerned, however,
introducing new atomic labels that define complex subformulas can
force the enumeration procedure to produce more partial assignments
than necessary \cite{masinass23}. This is what happens with \wmipa{}, and also
with 
\wmisapa{}~\cite{spallitta2022smt}, when conditions in $\conditionset$ are non-literals.

  \begin{figure}[tp]
    \begin{subfigure}{0.495\textwidth}
      \includegraphics[width=\textwidth]{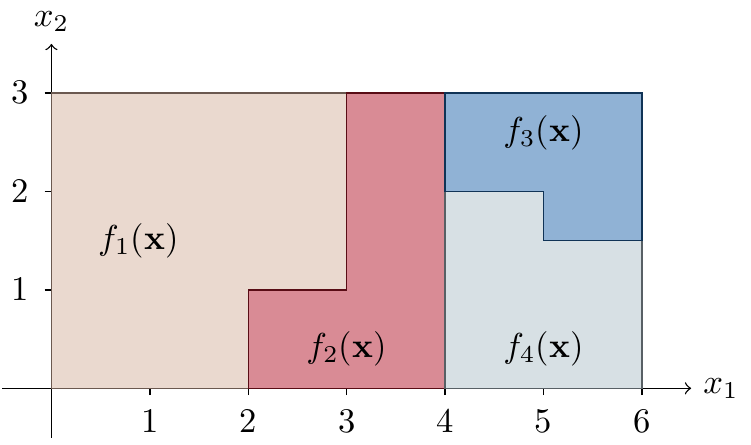}
      \caption{WMI problem.}%
      \label{fig:example_skv2_cnf}
    \end{subfigure} \hfill
    \begin{subfigure}{0.495\textwidth}
      \includegraphics[width=\textwidth]{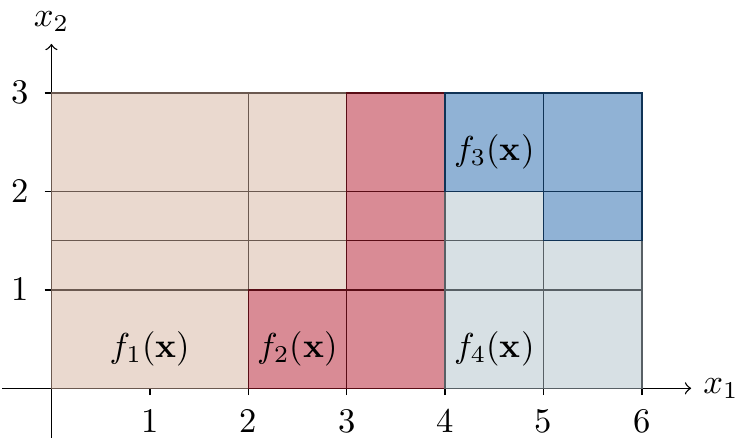}
      \caption{\wmipa}%
      \label{fig:example_skv2_cnf_pa}
    \end{subfigure} \hfill
    \begin{subfigure}{0.495\textwidth}
      \includegraphics[width=\textwidth]{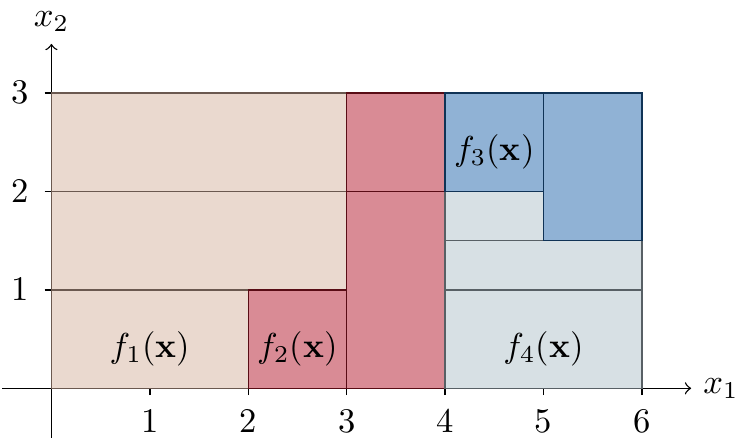}      
      \caption{\wmisapa}%
      \label{fig:example_skv2_cnf_sapa}
    \end{subfigure} \hfill
    \begin{subfigure}{0.495\textwidth}
      \includegraphics[width=\textwidth]{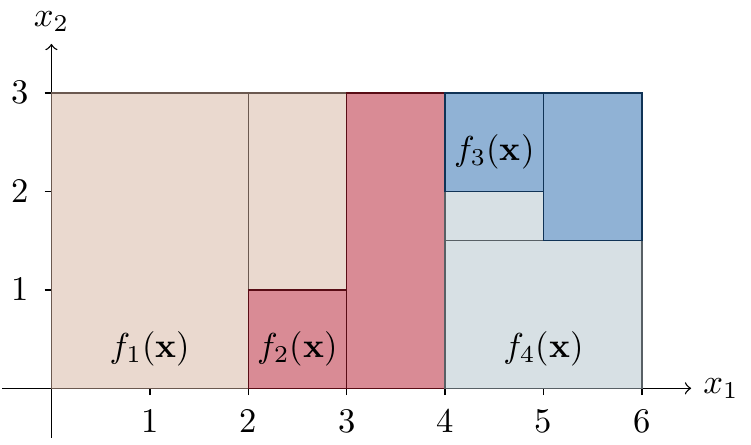}
      \caption{\method}%
      \label{fig:example_skv2_cnf_sapask}
    \end{subfigure}
    \caption{{Graphical representation of the effectiveness
        of \method{} as compared to the alternatives: (a) WMI
        problem from Example~\ref{exmp:example_skv2_cnf}, with 4 non-convex areas
        of \FI{} weights. Notice that the weight function includes
        non-literal conditions; (b) \wmipa{} computes \emph{20 distinct
          integrals};
        (c) \wmisapa{}, using both the implicit and
        explicit non-CNF skeleton versions, computes \emph{11 distinct
          integrals};
        (d) \method{}, that uses the ``local'' CNF
        explicit skeleton, computes \emph{8 distinct integrals}, the
        minimum number of integrals that is needed to retain convexity
        of their areas (i.e., \FI{} weights).} 
      }%
    \label{fig:example_skv2_cnf_full}
  \end{figure}

We illustrate this problem in Example~\ref{exmp:example_skv2_cnf}.
Our proposed solution will be eventually illustrated in Example~\ref{exmp:example_skv2_cnf_solved}.

 
\begin{example}%
  \label{exmp:example_skv2_cnf}
  Consider the following WMI problem:
  \begin{align*}
    \vi &= \top\\
    \wmis &= (x_1 \ge 0) \land (x_1 \le 6) \land (x_2 \ge 0) \land (x_2 \le 3)\\
    w &= \Ite{(x_1 \le 4)}{   
    \\&(\Ite{(x_1 \le 2) \lor ((x_1 \le 3) \land (x_2 > 1))}{f_1(\allx)}{f_2(\allx)})
    }{
      \\&(\Ite{(x_2 > 2) \lor ((x_1 > 5) \land (x_2 > \tfrac{3}{2}))}{f_3(\allx)}{f_4(\allx)})
    }
  \end{align*}
  {s.t.\ the $f_i(\allx)$s are condition-less functions,}
  which is graphically represented in Figure~\ref{fig:example_skv2_cnf}.
   Notice that some conditions of the weight function
  are not literals. Thus, if we wrote
  $\viencsk\defas\vi\wedge\supportwff\wedge\skwenc{}$ without concerning about this
  fact, then we
  would obtain the following non-CNF formula to feed to the SMT solver:
  \[ \arraycolsep=1pt
    \begin{array}{ll}
      (x_1 \ge 0) \land (x_1 \le 6) \land (x_2 \ge 0) \land (x_2 \le 3) & \land\\
    ((x_1 \le 4) \lor \neg(x_1 \le 4))                    & \land \\
    (\neg (x_1 \le 4) \lor
        \psi_1 \lor \neg \psi_1)  & \land \\
    (\pos (x_1 \le 4) \lor
        \psi_2 \lor \neg \psi_2)\\
  \end{array}
  \]
  where\\$\psi_1\defas \overbrace{(x_1 \le 2) \lor \underbrace{((x_1 \le 3) \land (x_2 > 1))}_{B_3}}^{B_1}$,
         $\psi_2\defas \overbrace{(x_2 > 2) \lor \underbrace{((x_1 > 5) \land (x_2 > \tfrac{3}{2}))}_{B_4}}^{B_2}  $  
  
  %
  \noindent The solver would apply Tseitin CNF-ization,
  labelling the subformulas by fresh Boolean
  atoms $\{B_1, B_2, B_3, B_4\}$ as above, producing the CNF formula
  $\phi$\footnote{{Here we colour the labelling clauses introduced by
      \CNFtseitin{\dots} as the corresponding regions in
      Figure~\ref{fig:example_skv2_cnf_full}.
      We recall that $\CNFtseitin{B_i\iff\psi_i}$ is
    $\CNFtseitin{B_i\imp\psi_i}\wedge\CNFtseitin{B_i\limp\psi_i}$.}}:
    \begin{eqnarray}
    \label{eq:tscnfsk} \def\arraystretch{0.8}
      \begin{array}{lc|l}
        (x_1 \ge 0) \land (x_1 \le 6) \land (x_2 \ge 0) \land (x_2 \le 3) & \land & \pos\vi\wedge\supportwff \\
      \big(\pos(x_1 \le 4) \lor \neg(x_1 \le 4)\big)                       & \land & \pos(x_1 \le 4) \lor \neg(x_1 \le 4) \\  
      \big(\neg (x_1 \le 4) \lor \pos  B_1 \lor \neg B_1 \big)            & \land  & \neg (x_1 \le 4) \lor  B_1 \lor \neg B_1 \\
      \textcolor{phi1T}{\big(\neg B_1  \lor \pos (x_1 \le 2) \lor \pos B_3\big)}                & \textcolor{phi1T}{\land} & \pos \textcolor{phi1T}{\CNFtseitin{B_1 \imp \psi_1}} \\  
      \textcolor{phi1T}{\big(\neg B_3  \lor \pos (x_1 \le 3)\big)                           \land 
                       \big(\neg B_3  \lor \pos (x_2 > 1)\big)}                             & \textcolor{phi1T}{\land} & \\
      \textcolor{phi1T}{\big(\pos B_3 \lor \neg (x_1 \le 3) \lor \neg (x_2 > 1)\big)}         & \land & \\
      \textcolor{phi1F}{\big(\pos B_1  \lor \neg (x_1 \le 2) \big)                         \land  
                       \big(\pos B_1  \lor \neg B_3 \big)}                                 & \textcolor{phi1F}{\land} & \pos \textcolor{phi1F}{\CNFtseitin{B_1 \limp \psi_1}} \\
      \big(\pos (x_1 \le 4) \lor \pos B_2 \lor \neg B_2\big)                   & \land & \pos (x_1 \le 4) \lor \pos B_2 \lor \neg B_2 \\
      \textcolor{phi2T}{\big(\neg B_2  \lor \pos (x_2 > 2) \lor \pos B_4\big)}                    & \textcolor{phi2T}{\land} & \pos \textcolor{phi2T}{\CNFtseitin{B_2 \imp \psi_2}}\\
      \textcolor{phi2T}{\big(\neg B_4  \lor \pos (x_1 > 5)\big)                             \land  
      \big(\neg B_4  \lor \pos (x_2 > \frac{3}{2})\big)}                   & \textcolor{phi2T}{\land} & \\
      \textcolor{phi2T}{\big(\pos B_4 \lor \neg (x_1 > 5) \lor \neg (x_2 > \frac{3}{2})\big)} & \textcolor{phi2T}{\land} &  \\
      \textcolor{phi2F}{\big(\pos B_2  \lor \neg (x_2 > 2) \big)                            \land 
      \big(\pos B_2  \lor \neg B_4 \big)}                                &  & \pos \textcolor{phi2F}{\CNFtseitin{B_2 \limp \psi_2}} \\
    \end{array}
    \end{eqnarray}
%
and
then compute \TA{\viencsk} in~\eqref{eq:newwmi1} as
\TA{\exists\allB.\phi}, s.t.\ the $B_i$s
will not occur in the enumerated models
(see \sref{sec:background-smt}).
    
    Consider, e.g.,  the partial truth assignment:
  \begin{eqnarray}
    \label{eq:musk}
    \mu\defas & \{(x_1 \ge 0), \pos (x_1 \le 6), \pos (x_2 \ge 0), \pos (x_2 \le 3)\} & \cup \\\nonumber
              & \{(x_1 \le 4), \neg (x_1 \le 2), \neg (x_1 \le 3), \neg B_1, \neg B_3\}&
  \end{eqnarray}
  which would be sufficient to make $\fgen{\w}{\mu}$ \FI{}
  because it suffices to identify $f_2(\allx)$.
  Unfortunately, $\mu$ does not suffice to evaluate~\eqref{eq:tscnfsk}
  to true  because of the last six clauses, which represent 
  $B_2\iff((x_2 > 2) \lor {((x_1 > 5) \land (x_2 > \frac{3}{2}))})$.
  Thus, when computing \TA{\exists\allB.\phi}, the solver is forced
  to assign truth values also to other atoms, in
  particular to $B_2$ and
  $B_4$. Whereas $B_4$ can be consistently assigned only to false due to the clause
  $(\neg B_4 \lor (x_1 > 5))$ and the fact that $(x_1 \le 4)$ is 
  true, $B_2$ can assume consistently both truth values, so that the solver is
  forced to branch on $B_2$ and $\neg B_2$, which respectively force
  $(x_2 > 2)$ and  $\neg(x_2 > 2)$, causing the unnecessary generation of two distinct integrals on $f_2(\allx)$, for $x_1\in[3,4],x_2\in[0,2]$ and
  $x_1\in[3,4],x_2\in]2,3]$ respectively, instead of only one on
  $x_1\in[3,4],x_2\in[0,3]$.%
  %
  (Notice that the atom $(x_2 > 2)$ belongs to a condition in the
  ``$(x_1 \le 4)=\bot$''
  branch of \w{}, whereas with $\mu$ we are
  actually considering the ``$(x_1 \le 4)=\top$'' branch.)
%
%

  This issue is graphically represented in
  Figure~\ref{fig:example_skv2_cnf_sapa}, where we show the regions
  enumerated by \wmisapa{}~\cite{spallitta2022smt}  on the above problem,
  by using either the implicit version of \skwenc{} as in the original
  algorithm or the non-CNF explicit version of \skwenc{}: 11 convex regions are enumerated. Notice that the brown, red and grey areas are each
  split into tree regions, whereas it is possible to split each
  into two convex regions only. 
  \hfill$\diamond$
\end{example}
The source of the problem is that, with Tseitin CNF-ization, each ``labelling
definition'' $\CNFtseitin{B_i\iff\psi_i}$ is conjoined to the rest of the
formula,
so that the SMT solver is forced to enumerate the different ways
to satisfy it \emph{even when the rest of the assignment selects a branch
which does not involve the condition $\psi_i$}.   
%
Notice that the fact that  
$B_i$ is implicitly treated as
existentially-quantified, so that it 
does not occur in the enumerated models
(see \sref{sec:background-smt}) does not help,
because $\psi_i$  and the atoms in it 
instead do occur in the enumerated models,
s.t.\ the enumerator is forced to split on their truth values anyway.
(E.g., in Example~\ref{exmp:example_skv2_cnf}, although $B_2$ is
implicitly existentially 
quantified, the enumerator is forced to
     split on $(x_2 > 2)$ anyway.)

To cope with this problem, we propose a variant of the Tseitin
CNF-ization for the skeleton where each
``labelling definition'' $\CNFtseitin{B\iff\psi}$ and all occurrences of $B$
are ``local'', that is, they
occur in the formula only implied by the conditions in \conds{}, so that the definitions
are simplified to true unless 
the assignment selects a branch
which propositionally satisfies all conditions in $\conds$.
%
Thus, in Algorithm~\ref{algo:convertsk}, lines~\ref{alg:convertsk-line12}-\ref{alg:convertsk-line16},
we substitute each
non-literal condition $\psi$  in \branch{} and $\defs{}_i$
from lines~\ref{alg:convertsk-line9}-\ref{alg:convertsk-line10} with a fresh Boolean atom
$B$ and we add
$\textstyle\shortcut{\bigvee_{\psi_i \in \conds} \neg \psi_i  \vee
  \CNFplaisted{B\iff\psi}}$%
\footnote{{recall from \sref{sec:background-smt} that
  \shortcut{C\vee\bigwedge_i C_i}\defas $\bigwedge_i (C\vee C_i)$.}}
instead of
$\CNFtseitin{B\iff\psi}$ to the main
formula, where
  %
$\CNFplaisted{\ldots}$ is 
the CNF-ization in~\cite{plaisted1986structure}%
\footnote{{Here we use \CNFplaisted{}
    instead of \CNFtseitin{} because the former generates fewer
    clauses and behaves generally better, although 
    we could use  \CNFtseitin{} without affecting
    correctness. }},
so that \branch{} is 
equivalent to
$(\bigwedge_{\psi_i \in \conds}\psi_i)
  \imp \left((B \vee \neg B) \wedge \CNFplaisted{B\iff\psi}\right)$.
(We recall that all $\psi_i$s in $\conds$ are literals
by construction, see Algorithm~\ref{algo:convertsk}.)

\begin{example}%
    \label{exmp:example_skv2_cnf_solved}
  With the problem of Example~\ref{exmp:example_skv2_cnf},
  $\viencsk\defas\vi\wedge\supportwff\wedge\skwenc$ is\footnote{{Here
      we colour the literals deriving from the labelling clauses
      introduced by \CNFplaisted{\dots} as the corresponding regions
      in Figure~\ref{fig:example_skv2_cnf_full}. We recall that $\CNFplaisted{B_i\iff\psi_i}$ is
    $\CNFplaisted{B_i\imp\psi_i}\wedge\CNFplaisted{B_i\limp\psi_i}$.}}:
  \[\def\arraystretch{0.8}
  \begin{array}{lc|l}
     (x_1 \ge 0) \land (x_1 \le 6) \land (x_2 \ge 0) \land (x_2 \le 3) & \land
    & \pos\vi\wedge\supportwff\\
\big(\pos (x_1 \le 4) \lor \neg (x_1 \le 4)\big)                                         & \land &
\pos (x_1 \le 4) \lor \neg (x_1 \le 4)\\
\big(\neg (x_1 \le 4) \lor \pos B_1 \lor \neg B_1\big)                                 & \land &
\neg (x_1 \le 4) \lor B_1 \lor \neg B_1\\
\big(\neg (x_1 \le 4) \lor \textcolor{phi1T}{\neg B_1 \lor \pos(x_1 \le 2) \lor \pos B_3}\big)                  & \land &
\shortcut{\neg(x_1 \le 4) \lor \textcolor{phi1T}{\CNFplaisted{B_1 \imp \psi_1}}}\\
\big(\neg (x_1 \le 4) \lor \textcolor{phi1T}{\neg B_1 \lor \neg B_3 \lor \pos(x_1 \le 3)}\big)       & \land &\\
\big(\neg (x_1 \le 4) \lor \textcolor{phi1T}{\neg B_1 \lor \neg B_3 \lor \pos(x_2 > 1)}\big)         & \land &\\ 
\big(\neg (x_1 \le 4) \lor \textcolor{phi1F}{\pos B_1 \lor \neg (x_1 \le 2)}\big)                        & \land &
\shortcut{\neg(x_1 \le 4) \lor \textcolor{phi1F}{\CNFplaisted{B_1 \limp \psi_1}}}\\
\big(\neg (x_1 \le 4) \lor \textcolor{phi1F}{\pos B_1 \lor \neg (x_1 \le 3) \lor \neg (x_2 > 1)}\big) & \land & \\
\big(\pos (x_1 \le 4) \lor \pos B_2 \lor \neg B_2\big)                                      & \land &
\pos (x_1 \le 4) \lor B_2 \lor \neg B_2\\
\big(\pos (x_1 \le 4) \lor \textcolor{phi2T}{\neg B_2 \lor \pos(x_2 > 2) \lor \pos B_4}\big)                       & \land &
\shortcut{\pos (x_1 \le 4) \lor \textcolor{phi2T}{\CNFplaisted{B_2 \imp \psi_2}}}\\
\big(\pos (x_1 \le 4) \lor \textcolor{phi2T}{\neg B_2 \lor \neg B_4 \lor \pos(x_1 > 5)}\big)                  & \land &\\
\big(\pos (x_1 \le 4) \lor \textcolor{phi2T}{\neg B_2 \lor \neg B_4 \lor \pos(x_2 > \frac{3}{2})}\big)        & \land &\\
\big(\pos (x_1 \le 4) \lor \textcolor{phi2F}{\pos B_2 \lor \neg (x_2 > 2)}\big)                                & \land &
\shortcut{\pos (x_1 \le 4) \lor \textcolor{phi2F}{\CNFplaisted{B_2 \limp \psi_2}}}\\

\big(\pos (x_1 \le 4) \lor \textcolor{phi2F}{\pos B_2 \lor \neg (x_1 > 5) \lor \neg (x_2 > \frac{3}{2})}\big)  &  & \\
\end{array}
\]
(Here $B_1,B_2$ are labels for conditions
  $\psi_1\defas(x_1 \le 2) \lor ((x_1 \le 3) \land (x_2 > 1))$,
   $\psi_2\defas(x_2 > 2) \lor ((x_1 > 5) \land (x_2 > \frac{3}{2}))$
respectively, as in
Example~\ref{exmp:example_skv2_cnf}, whereas $B_3,B_4$ are labels
introduced by $\CNFplaisted{\ldots}$, which do not produce valid clauses in the
form $(\ldots \vee B_i\vee\neg B_i)$.%
)
For instance, the assignment $\mu$ in~\eqref{eq:musk}
evaluates to true all the clauses of \viencsk{}, so that the solver
considers it as satisfying assignment for the formula.

In Figure~\ref{fig:example_skv2_cnf_sapask} we show the regions
enumerated by the \method{} procedure
(\sref{sec:revised-enumeration}), using as skeleton the ``local'' CNF version of \skwenc{}.
We see that the regions enumerated are 8, instead of the 11 of Figure~\ref{fig:example_skv2_cnf_sapa}. In this
case 8 is also the minimum number of convex areas into which we can split the problem, 
because the 4 distinct coloured areas in
Figure~\ref{fig:example_skv2_cnf} are non-convex, and as such their
integral cannot be computed with single integrations. \hfill$\diamond$ 
\end{example}

We stress the importance of the locality of labelling clauses in the
enumeration of truth assignments: the \SMT{} enumerator needs
branching on $B$ (hence, on $\psi$) \emph{only in those branches in which all \conds{} are
true}, that is, only in those branches where a truth value for condition
$\psi$ is needed to define a branch of $\w$. 
Notice that, if \CNFplaisted{} introduces some labelling
definitions, then also the latter are active only when all
\conds{} are true.

\begin{algorithm}[t]
\begin{algorithmic}[1]
  \caption[A]{$\textsf{\method}(\varphi, \w, \allx, \allA)$}%
  \label{algo:sawmipask}

  \STATE $\skwenc{} \gets \ConvertSK{w, \emptyset}$\label{algo:sawmipask:enc:start}
  

  \STATE $\viencsk \gets \vi \land \wmis \land \skwenc{}$\label{algo:sawmipask:enc:end}

  \STATE $\MUA \gets \TA{\exists \allx.\viencsk}$\label{algo:sawmipask:ta}

  \STATE $\vol \gets 0$  


  \FOR{$\muA \in \MUA$}{\label{algo:sawmipask:tata:start}
  
    \STATE $k \gets | \allA \setminus \muA |$

    \STATE $\Simplify{\viencskmua}$ 
    
    \IF{$\LitConj{\viencskmua}$}{

      \STATE $\vol \gets \vol + 2^k \cdot \WMINB{\viencskmua, \fgen{\w}{\muA} | \allx}$

    } \ELSE {%
\label{algo:sawmipask:ta-setup}
      \STATE $\mathcal{M} \gets  \TA{\viencskmua}$

      \FOR {$(\mu \defeq \muA{}' \land \muLRA) \in \mathcal{M}$} {\label{algo:sawmipask:ta:start}

        \STATE $k' \gets k -| \muA{}' |$
        
        \STATE $\vol \gets \vol + 2^{k'} \cdot \WMINB{\muLRA, \fgen{\w}{\muA \land \mu} | \allx}$

      }
      \ENDFOR
    }
    \ENDIF
  }\label{algo:sawmipask:tata:end}
  \ENDFOR
  
  \RETURN {\vol}
\end{algorithmic}
\end{algorithm}

\subsubsection{The \method{} Procedure}%
\label{sec:revised-enumeration}
We
devised a new algorithm, \method,
that \emph{explicitly}
encodes the weight 
function as \encSK{} (\sref{sec:explicitskeleton-encoding}),
handles non-literal conditions effectively
(\sref{sec:nonatomic-conditions}),
and introduces some further improvements in the enumeration, which we
describe below.
The outline of 
the procedure is shown in Algorithm~\ref{algo:sawmipask}.
  (To simplify the narration, here we assume that the encoding of \encSK{}
did not cause the introduction of fresh Boolean variables \allB, see
\sref{sec:nonatomic-conditions}; if not so,
  then $\TA{\exists \allx.\viencsk}$ and $\TA{\viencskmua}$ in
  Algorithm~\ref{algo:sawmipask} should be replaced with
  $\TA{\exists \allx.\exists\allB.\viencsk}$ and $\TA{\exists\allB.\viencskmua}$ respectively.) 

%
{\begin{itemize}
  \item (Lines~\ref{algo:sawmipask:enc:start}-\ref{algo:sawmipask:ta})
    After producing \skwenc{} by Algorithm~\ref{alg:convertsk}, we first enumerate by 
    projected AllSMT a set $\MUA$
    of \emph{partial} truth assignments $\muA$ over  \allA{}, s.t.\  $\MUA\defas\TA{\exists \allx. \viencsk}$, where $\viencsk\defas\vi \land \wmis \land \skwenc{}$.
  \item
    (Lines~\ref{algo:sawmipask:tata:start}-\ref{algo:sawmipask:tata:end})
    For each $\mua\in\MUA$ we compute and simplify the residual
    formula \viencskmua.
    As with  \wmipa{} and \wmisapa{}, if  \viencskmua{} is already
    a conjunction of literals,
    then we can simply compute and return its integral,
    which is multiplied by a $2^k$ factor, $k$ being the number of
    unassigned Boolean atoms in $\mua$.\footnote{Notice that in this
      case
      the conjunction of literals \viencskmua{} does not
      contain Boolean literals, because
      they would have already been assigned
      by \mua.}
    If this is not the case, 
    we proceed by enumerating a set $\MU{}$ of \emph{partial} truth assignments $\mu = \muA{}' \land \muLRA$
    over the remaining atoms of the formula,
    s.t.\ $\MU{}\defas\TA{\viencskmua{}}$. Notice that,
    since \mua{} is partial, \viencskmua{} could still contain Boolean
    atoms (see Examples~\ref{exmp:sapa_vs_sk} and~\ref{ex:muastar}), so that $\mu$
    could contain some non-empty Boolean component $\muA{}'$.
    We then compute the integral of the \FI\ function $\fgen{\w}{\muA \land \mu}$
    over the convex area \muLRA{}. As above, 
    we need to multiply this integral by a $2^{k'}$ factor,
    $k'\defas k-|\muA{}'|$ being the number of unassigned 
    Boolean atoms in $\muA \land \mu$.
\end{itemize}}

\noi As with \wmipa{} and \wmisapa{}, in the implementation
(Lines~\ref{algo:sawmipask:ta}-\ref{algo:sawmipask:tata:start} and~\ref{algo:sawmipask:ta-setup}-\ref{algo:sawmipask:ta:start})
the sets
$\MUA$ and $\MU$ are not generated
explicitly; rather, their elements are generated, integrated and then dropped one-by-one, so as to avoid
space blow-up (see Remark~\ref{remark:enumeration}).

We remark that producing smaller partial assignments over
the Boolean atoms, as done with \method{}, has a positive effect regardless of the structure of
the weight function.
In Example~\ref{exmp:sapa_vs_sk} we give an intuition of the benefits introduced by the improvements of the enumeration
phase in \method{}.

\begin{table}[th]

 \begin{subtable}{\textwidth}
    \centering
  \begin{tabular}{rrr|rr|rr}
    \multicolumn{3}{c}{{$\muA$}} & \multicolumn{2}{c}{$\muAres$} & \multicolumn{2}{c}{$\muLRA$}\\
    \hline
    $\neg A_1$ & $\neg A_2$ & $     A_3$ &            &            & $(x \ge 0)$ & $(x \le 3)$ \\
    $     A_1$ & $\neg A_2$ &            &            & $     A_3$ & $(x \ge 1)$ & $(x \le 3)$ \\
    $     A_1$ & $\neg A_2$ &            &            & $\neg A_3$ & $(x \ge 1)$ & $(x  \le  4)$ \\
    $        $ & $     A_2$ &            & $     A_1$ & $     A_3$ & $(x \ge 2)$ & $(x \le 3)$ \\
    $        $ & $     A_2$ &            & $     A_1$ & $\neg A_3$ & $(x \ge 2)$ & $(x  \le  4)$ \\
    $        $ & $     A_2$ &            & $\neg A_1$ & $     A_3$ & $(x \ge 2)$ & $(x \le 3)$ \\
    $        $ & $     A_2$ &            & $\neg A_1$ & $\neg A_3$ & $(x \ge 2)$ & $(x  \le  4)$ \\
  \end{tabular}
  \caption{\label{tab:exmpsk-sapa}Assignments enumerated by \wmipa{} and \wmisapa{}}%
\end{subtable}
\begin{subtable}{\textwidth}
    \centering
  \begin{tabular}{rrr|rrr}
    \multicolumn{3}{c}{{$\muA$}} & \multicolumn{3}{c}{$\mu \defeq \muA{}' \land \muLRA$} \\
    \hline
    $\neg A_1$ & $\neg A_2$ & $     A_3$ &            & $(x \ge 0)$ & $(x \le 3)$ \\
    $     A_1$ & $\neg A_2$ &            & $     A_3$ & $(x \ge 1)$ & $(x \le 3)$ \\
    $     A_1$ & $\neg A_2$ &            & $\neg A_3$ & $(x \ge 1)$ & $(x  \le  4)$ \\
    $        $ & $     A_2$ &            & $     A_3$ & $(x \ge 2)$ & $(x \le 3)$ \\
    $        $ & $     A_2$ &            & $\neg A_3$ & $(x \ge 2)$ & $(x  \le  4)$ \\
  \end{tabular}
  \caption{\label{tab:exmpsk-sapask}Assignments enumerated by \method{}.}  
 \end{subtable}
    \caption{\label{tab:exmpsk-sapa-vs-sapask}Assignments enumerated by \wmipa{} and \wmisapa{}~(a) and
      by \method~(b) for the problem in
      Example~\ref{exmp:sapa_vs_sk}. {(We omit the
      \larat-literals which are implied by the others and do not contribute to the integral 
      ---e.g., $(x\ge 0)$ when $(x\ge 2)$ is true.)}}
  
\end{table}
  
\begin{example}%
\label{exmp:sapa_vs_sk}
Consider the following WMI problem:
  \begin{align*}
    \vi &= (A_1 \lor A_2 \lor A_3) \land 
      (\neg A_1 \vee (x \ge 1)) \land 
      (\neg A_2 \vee (x \ge 2)) \land 
      (\neg A_3 \vee (x \le 3)) \\
    \wmis &= (x \ge 0) \land (x \le 4)\\
    \w &= 1.0
  \end{align*}
  (\w{} is constant and $\conditionset=\emptyset$, thus $\bigwedge_i{B_i\iff \psi_i}$ 
  and  
  $\skwenc{}$ reduce to $\top$ and is thus irrelevant, and
  $\vistar
  =\viencsk\defas\vi\land\wmis$.)
 %
  \wmipa{} and \wmisapa{}~\cite{spallitta2022smt} enumerate the 7 assignments listed in
 Table~\ref{tab:exmpsk-sapa},
 whereas with \method{}  the number of
 assignments enumerated shrinks to 5, as shown in Table~\ref{tab:exmpsk-sapask}.

 With \wmipa{}, \TTA{\exists\allx.\vistar} directly generates in one step the 7 total truth
 assignments $\muA\wedge\muAres$ in Table~\ref{tab:exmpsk-sapa}. 
 
With both \wmisapa{} and \method, 
 \TA{\exists\allx.\viencsk}
 produces  the partial assignment $\muA=\set{A_2}$, so that 
 $\viencskmua$ reduces to
 $ (x \ge 0) \wedge (x \le 4) \land
 (\neg A_1 \vee (x \ge 1)) \land
 (x \ge 2) \land 
(\neg A_3 \vee (x \le 3)) $. 
Then:

with \wmisapa{},
$\TTA{\exists \allx.\viencskmua}$
enumerates all 4 total residual
      assignments on $A_1,A_3$,%
\footnote{Here $\exists
        \allx.\viencskmua$ corresponds to valid Boolean formula on \set{A_1,A_3}.}
      duplicating the two integrals on $x\in[2,3]$ and
      $x\in[2,4]$ by uselessly case-splitting on $A_1$ and $\neg A_1$,
      as in the last four rows in Table~\ref{tab:exmpsk-sapa};

with \method{},  $\TA{\viencskmua}$ enumerates only the two
partial
assignments:%
\footnote{Here the truth value of $A_1$ is irrelevant
because $(x \ge 1)$ is forced to be  true by $(x \ge 2)$.}
$\set{\pos A_3,(x \ge 0),(x  \le  4),(x \ge 1),(x \ge 2),(x \le 3)}$,\\ 
$\set{\neg A_3,(x \ge 0),(x  \le  4),(x \ge 1),(x \ge 2)}$,\\
as reported in the last two rows in Table~\ref{tab:exmpsk-sapask}.
The two corresponding integrals are multiplied by $2^{(3-|\set{A_2}|-|\set{A_3}|)}=2$ and
$2^{(3-|\set{A_2}|-|\set{\neg A_3}|)}=2$ respectively.~\hfill$\diamond$

\end{example}
   The benefit becomes more evident as the number of Boolean atoms increases, since
  if we avoid computing the \TTA{\exists\allx.\viencskmua} over the residual Boolean atoms, then
  we are potentially largely cutting the number 
  of integrals.

\section{Experimental Evaluation}%
\label{sec:expeval}

In the following experiments, we evaluate the performance gain of
\method{} over the current state of the art in WMI.\@
Specifically, we explore the effect of the novel
structure encoding using the same setting of our recent conference
paper~\citep{spallitta2022smt}. To this extent, we compare the
proposed algorithm \method{}, and its previous version
\wmisapa{}~\citep{spallitta2022smt},  with the previous SMT-based approach \wmipa{}~\citep{morettin-wmi-aij19}, and with the
state-of-the-art KC-based approaches: XADD~\citep{KolbMSBK18}, XSDD
and FXSDD~\citep{kolb2020exploit}.
For \wmipa{}, \wmisapa{} and \method{}, we use a slightly-modified version of \mathsat{}~\cite{mathsat5_tacas13}%
\footnote{We have added one option to \mathsat{} allowing to disable the
  simplification of valid clauses, which would otherwise eliminate all
  the clauses ``$(\ldots\vee\psi_i\vee\neg\psi_i)$'' from \skw{}. The binary file integrating these modifications is attached to the code of \method{}.} 
for SMT enumeration.%
\footnote{To compute \TA{\ldots}, \mathsat{} is invoked with the options:\\
\msatoption{-dpll.allsat\_allow\_duplicates=false} (all assignments
differ for at least one truth value),
\msatoption{-dpll.allsat\_minimize\_model=true} (assignments are partial and
are minimized),
\msatoption{-preprocessor.simplification=0} and
\msatoption{-preprocessor.toplevel\_propagation=false}
(disable several non-validity-preserving
preprocessing steps).
To compute \TTA{\ldots}, the second option is set to \msatoption{false}.
%
}

Convex integration is handled by
\latteintegrale{}~\cite{latte} in the SMT-based approaches, whereas
KC-based solvers use PSI Algebra for both integration and
inconsistency checking~\citep{psi}. This difference is not
penalizing the KC-based approaches, which were shown to achieve worse results when using \latte~\cite{kolb2020exploit}.

\noi{} All experiments are performed on an Intel Xeon Gold 6238R @ 2.20GHz 28
Core machine with 128 GB of RAM and running Ubuntu Linux 20.04. The
code of both \wmisapa{} and \method{} is freely available at
\url{https://github.com/unitn-sml/wmi-pa}, and the benchmarks are
available at
\url{https://github.com/unitn-sml/wmi-benchmarks}. 

\noi We report our findings using cactus plots. Each tick on the x-axis
corresponds to a single instance, and results on the y-axis are sorted
independently for each method. Steeper slopes mean lower
efficiency. 

\subsection{Synthetic experiments}%
\label{sec:exp-synth-exact}


\paragraph{Dataset description}

As in our previous paper, random formulas are obtained using the
generator introduced by~\citet{morettin-wmi-aij19}.
In contrast with other recent
works~\citep{KolbMSBK18,kolb2020exploit}, these synthetic benchmarks
do not contain strong structural regularities, offering a more neutral
perspective on how the different techniques are expected to perform on
average.
We fix to 3 the number of Boolean and real variables, while varying
the depth of the weights in the range $\lbrack4,7\rbrack$. Timeout is
set to 3600 seconds.

\begin{figure}[t]
  \centering
   \begin{subfigure}[b]{\textwidth}
    \resizebox{\textwidth}{!}{\begin{tikzpicture}
\begin{customlegend}[
    legend columns=-1,
    legend style={
      column sep=1ex,
    },
    every axis plot/.append style={ultra thick, mark=x, mark size=3},
  ]
    \addlegendimage{mark=none, black, dashed} \addlegendentry{Timeout}
    \addlegendimage{XSDD}\addlegendentry{XSDD}
    \addlegendimage{FXSDD}\addlegendentry{FXSDD}
    \addlegendimage{XADD}\addlegendentry{XADD}
    \addlegendimage{WMIPA}\addlegendentry{\wmipa}
    \addlegendimage{WMISAPA}\addlegendentry{\wmisapa}
    \addlegendimage{WMISAPASK}\addlegendentry{\method}
\end{customlegend}
\end{tikzpicture}}
  \end{subfigure}
   \begin{subfigure}[b]{0.49\textwidth}
    \includegraphics[width=\textwidth]{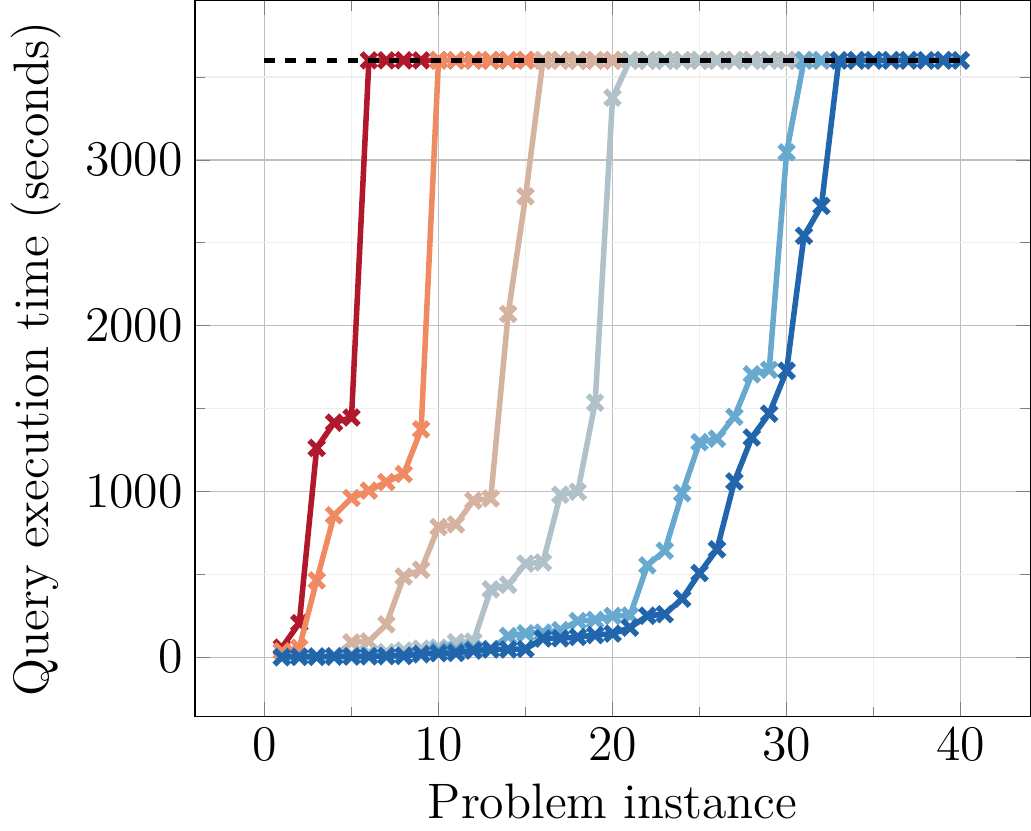}
   \end{subfigure}
   \hfill
   \begin{subfigure}[b]{0.49\textwidth}
    \includegraphics[width=0.97\textwidth]{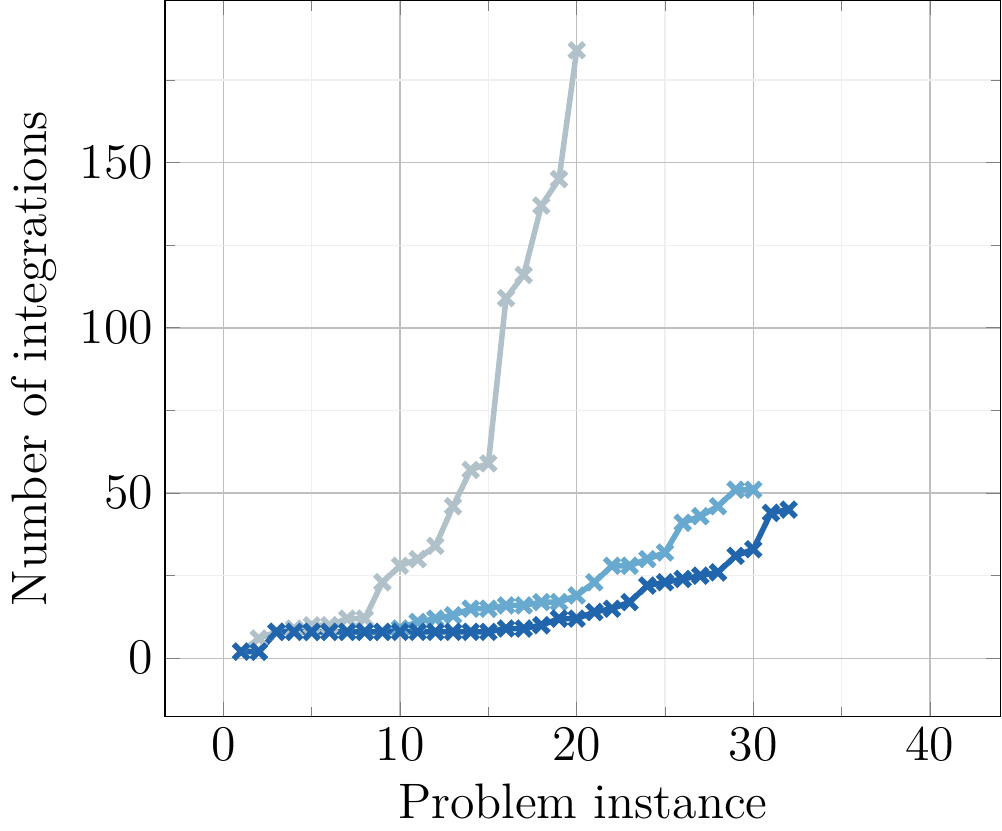}
   \end{subfigure}
   \caption{Cactus plots of the synthetic
  experiments reporting execution time (left) for all the methods; number of integrals (right) for \wmipa, \wmisapa{} and \method{}.
}\label{fig:synth_tree}
\end{figure}

\paragraph{Results}
Consistently with previous papers, in this setting SMT-based solvers
obtain superior performance with respect to KC-based ones
(Figure~\ref{fig:synth_tree} left), which suffer from the complex
algebraic dependencies between continuous variables.
Our structure-aware algorithms, \wmisapa{} and \method{}, further drastically improve over \wmipa{} in terms of time and number of computed integrals
(Figure~\ref{fig:synth_tree}).
These results are aligned with our expectations. Indeed, the advantage of structure-awareness given by the
  introduction of either form of skeleton is eye-catching. 

Additionally, \method{} is able to gain some advantage over its
  previous version \wmisapa{}, although the order of magnitude
remains quite the same. Notice that in this setting, the number of Boolean atoms
is limited to 3, so that the revised enumeration technique of \sref{sec:revised-enumeration} for the
Boolean component of the formula does not show its full potential.
The benefits of the novel algorithm \method{} with respect to \wmisapa{} (and 
\wmipa{})  will be evident in the next experiment, where more-complex logical
structures are involved.

\subsection{Inference on Density Estimation Trees}%
\label{sec:exp-det}

\paragraph{Dataset description}
We then consider the problem of computing arbitrary algebraic
queries over \emph{Density Estimation Trees}
(DETs)~\citep{ram2011density}. DETs are the density estimation
equivalent of decision or regression trees, encoding piecewise
constant distributions over mixed continuous/discrete domains. Having
only univariate conditions in the internal nodes, DETs support
efficient density estimation and marginal inference over single
variables. Answering algebraic queries over multiple variables, like
$\Pr(X \le Y)$, requires instead marginalizing over an oblique
constraint, which is reduced to: 
\begin{align}
\Pr(X \le Y) = \frac{\WMI{((X \le Y)
  \wedge \chi_\mathit{DET}, w_\mathit{DET})}}{\WMI{( \chi_\mathit{DET}, w_\mathit{DET})}}.
\end{align}
\noindent This type of query commonly arises when applying WMI-based
inference to probabilistic formal verification tasks, when properties
involve multiple continuous variables.
The support
$\chi_\mathit{DET}$ is obtained by conjoining the lower and upper
bounds of each continuous variable, while $w_\mathit{DET}$ encodes the
density estimate by means of nested if-then-elses and non-negative
constant leaves.

\noi Consistently with our previous conference
paper~\cite{spallitta2022smt}, we train DETs on a selection of
hybrid datasets from the UCI repository~\citep{Dua2019uci}. The datasets and  
resulting models are summarized in \sref{sec:det-datasets}. Following
the approach of~\cite{morettin2020learning}, discrete numerical
features are relaxed into continuous variables, while $n$-ary
categorical features are one-hot encoded with $n$ binary variables.

DETs are trained on each dataset using the standard greedy
procedure~\citep{ram2011density} with bounds on the number of
instances for each leaf set to $[100, 200]$. We investigate the
effect of algebraic dependencies by generating increasingly complex
queries involving a number of variables $k = \max(1, \lfloor H \cdot
|\allx|\rfloor)$, where $H \in [0,1]$ is the ratio of continuous
variables appearing in the inequality. For each value of $H$, 5 random
inequalities involving $k$ variables are generated, for a total of
$20$ problem instances for each dataset.
Given the larger number of problems in this setting, the timeout
has been decreased to 1200 seconds.

\paragraph{Results}

\begin{figure}[t]
  \centering
   \begin{subfigure}[b]{\textwidth}
    \resizebox{\textwidth}{!}{\begin{tikzpicture}
\begin{customlegend}[
    legend columns=-1,
    legend style={
      column sep=1ex,
    },
    every axis plot/.append style={ultra thick, mark=x, mark size=3},
  ]
    \addlegendimage{mark=none, black, dashed} \addlegendentry{Timeout}
    \addlegendimage{XSDD}\addlegendentry{XSDD}
    \addlegendimage{FXSDD}\addlegendentry{FXSDD}
    \addlegendimage{XADD}\addlegendentry{XADD}
    \addlegendimage{WMIPA}\addlegendentry{\wmipa}
    \addlegendimage{WMISAPA}\addlegendentry{\wmisapa}
    \addlegendimage{WMISAPASK}\addlegendentry{\method}
\end{customlegend}
\end{tikzpicture}}
  \end{subfigure}
   \begin{subfigure}[b]{0.49\textwidth}
   \includegraphics[width=\textwidth]{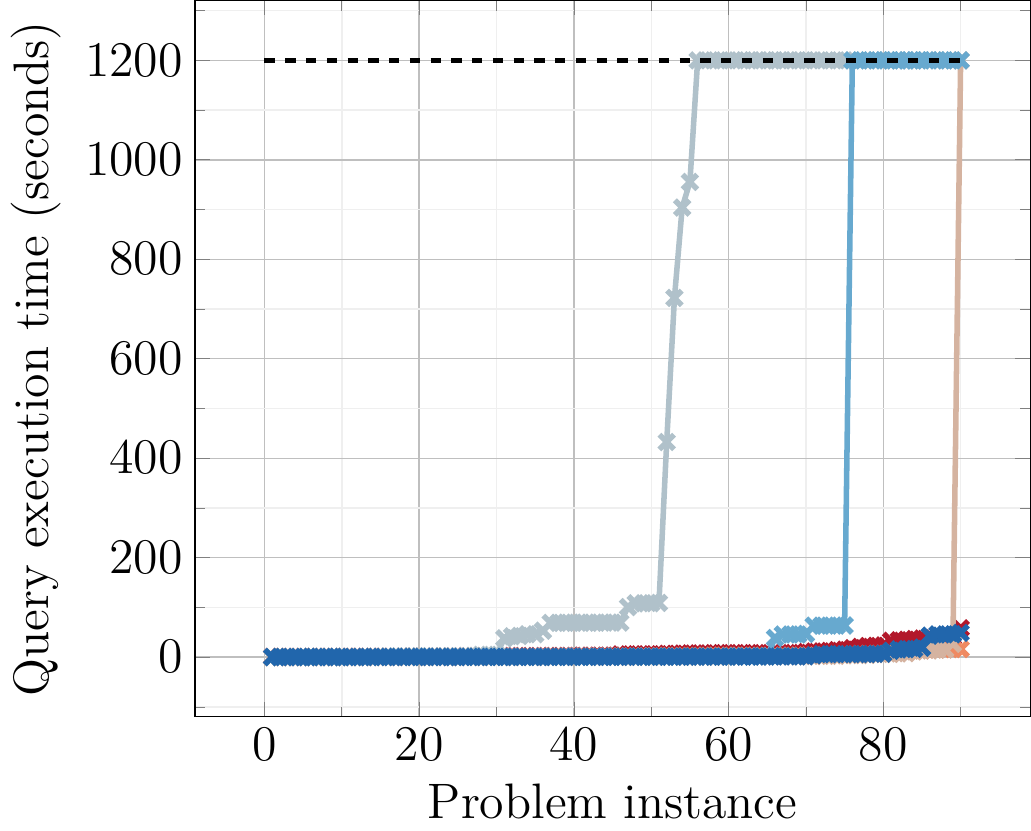}
   \caption{$H=0.25$}%
   \label{fig:plot_sapask_det_time_025}
   \end{subfigure} \hfill
   \begin{subfigure}[b]{0.49\textwidth}
   \includegraphics[width=\textwidth]{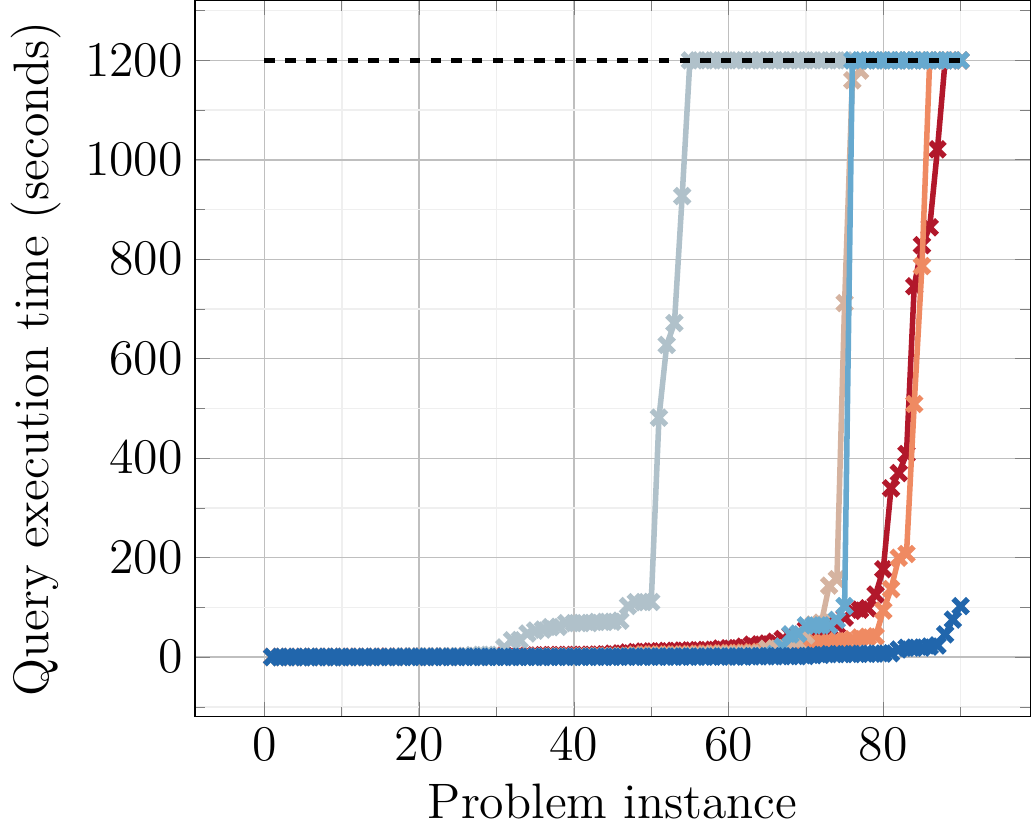}
   \caption{$H=0.5$}%
   \label{fig:plot_sapask_det_time_05}
   \end{subfigure} \par\bigskip
   \begin{subfigure}[b]{0.49\textwidth}
   \includegraphics[width=\textwidth]{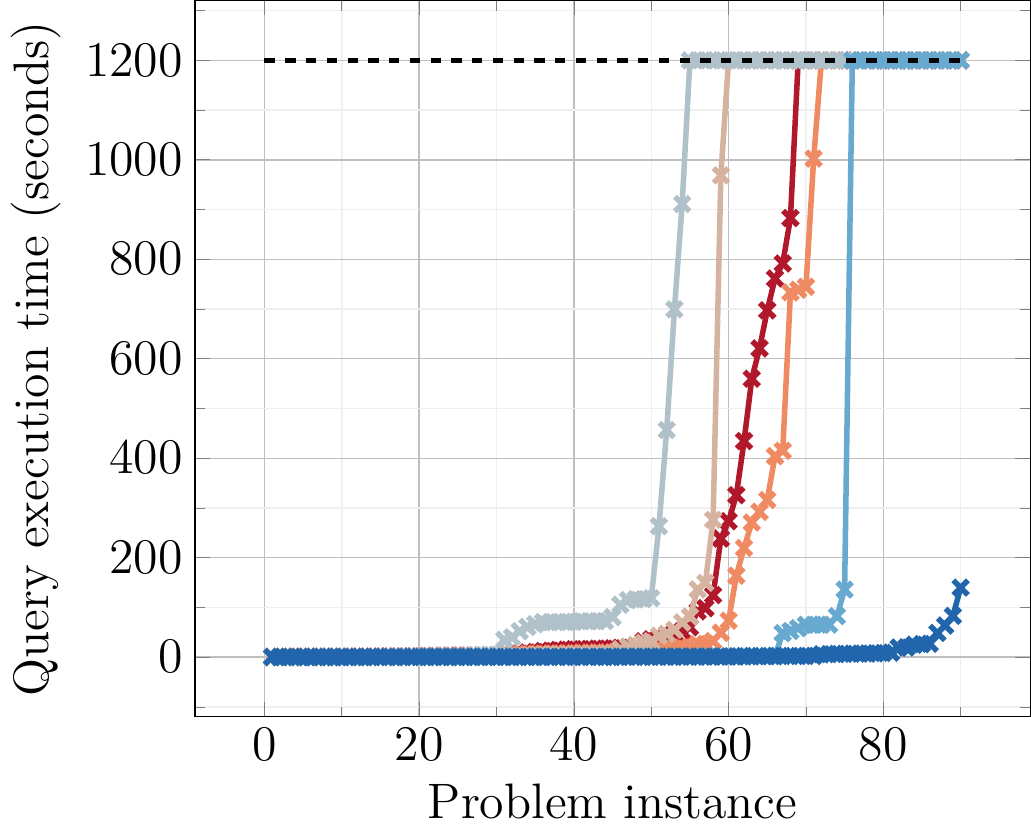}
   \caption{$H=0.75$}%
   \label{fig:plot_sapask_det_time_075}
   \end{subfigure} \hfill
   \begin{subfigure}[b]{0.49\textwidth}
   \includegraphics[width=\textwidth]{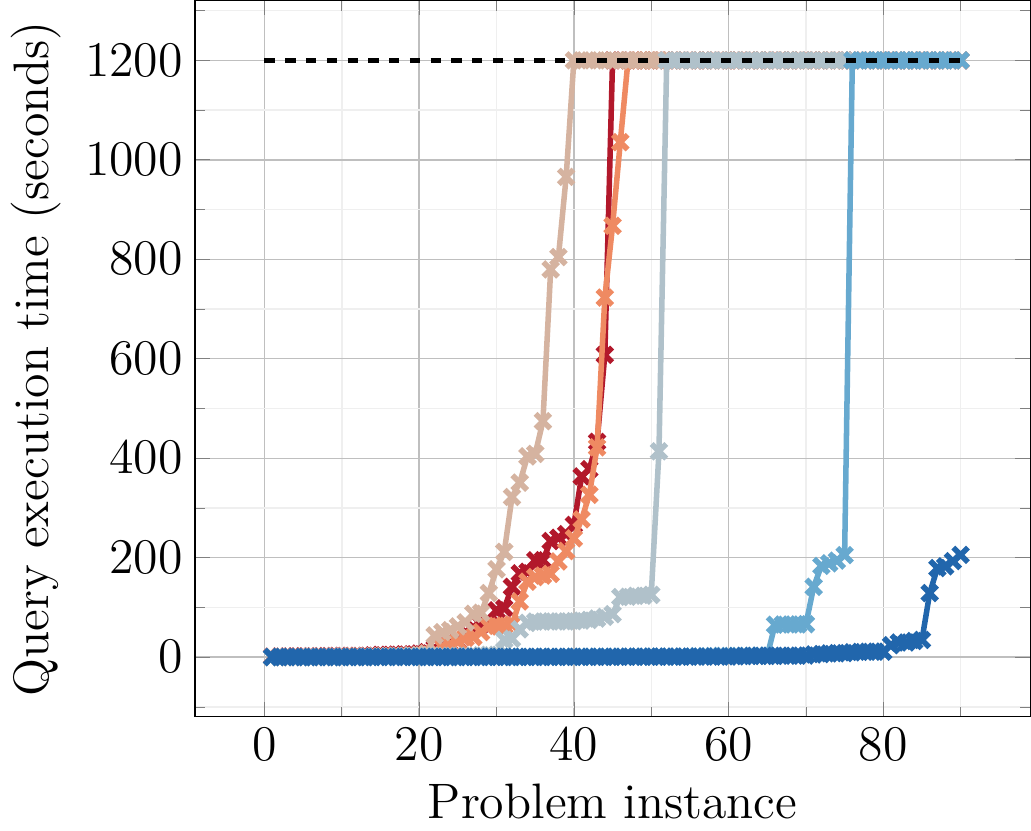}
   \caption{$H=1.0$}%
   \label{fig:plot_sapask_det_time_10}
   \end{subfigure}
   \caption{Cactus plots representing average query execution times in seconds on a set of DET problems with $H \in \{0.25, 0.5, 0.75, 1\}$.
  }\label{fig:plot_sapask_det_time}
\end{figure}

\begin{figure}[ht!]
  \centering
  \begin{subfigure}[b]{\textwidth}
    \centering
    \resizebox{0.5\textwidth}{!}{\begin{tikzpicture}
\begin{customlegend}[
    legend columns=-1,
    legend style={
      column sep=1ex,
    },
    every axis plot/.append style={ultra thick, mark=x, mark size=3},
  ]
    \addlegendimage{WMIPA}\addlegendentry{\wmipa{}}
    \addlegendimage{WMISAPA}\addlegendentry{\wmisapa{}}
    \addlegendimage{WMISAPASK}\addlegendentry{\method{}}
\end{customlegend}
\end{tikzpicture}}
  \end{subfigure}
   \begin{subfigure}[b]{0.49\textwidth}
   \includegraphics[width=\textwidth]{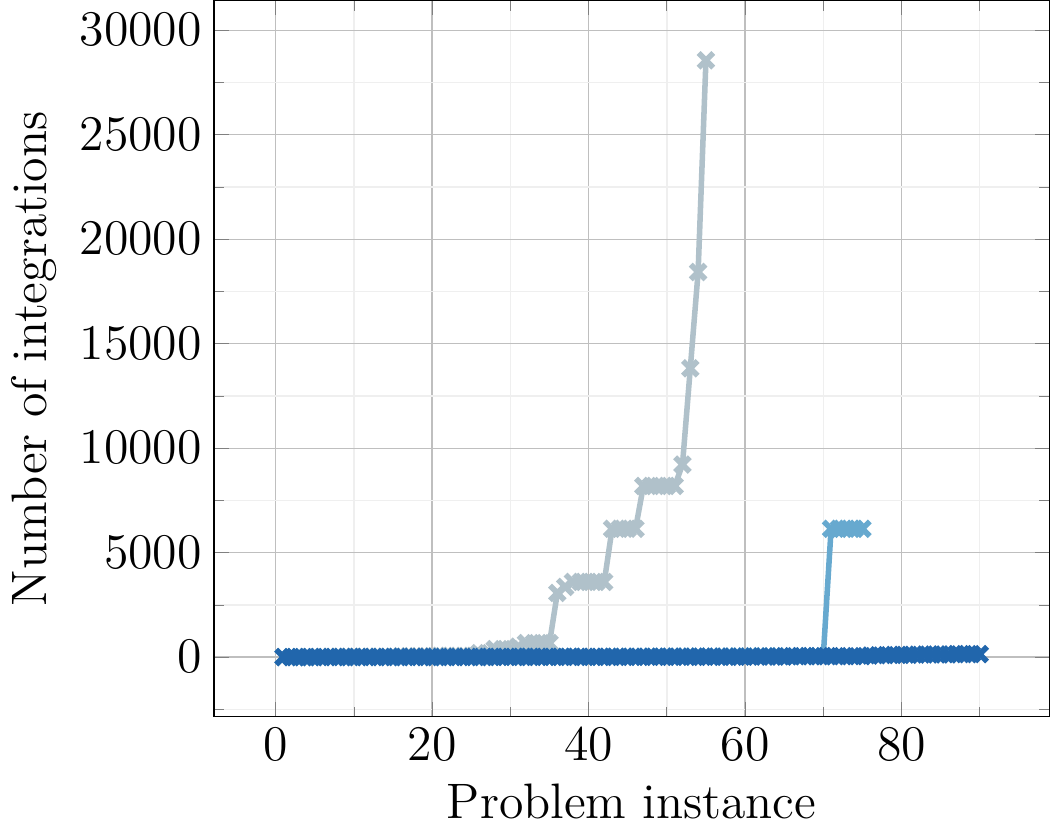}
  \caption{$H=0.25$}%
  \label{fig:plot_sapask_det_n_integrations_025}
   \end{subfigure} \hfill
   \begin{subfigure}[b]{0.49\textwidth}
   \includegraphics[width=\textwidth]{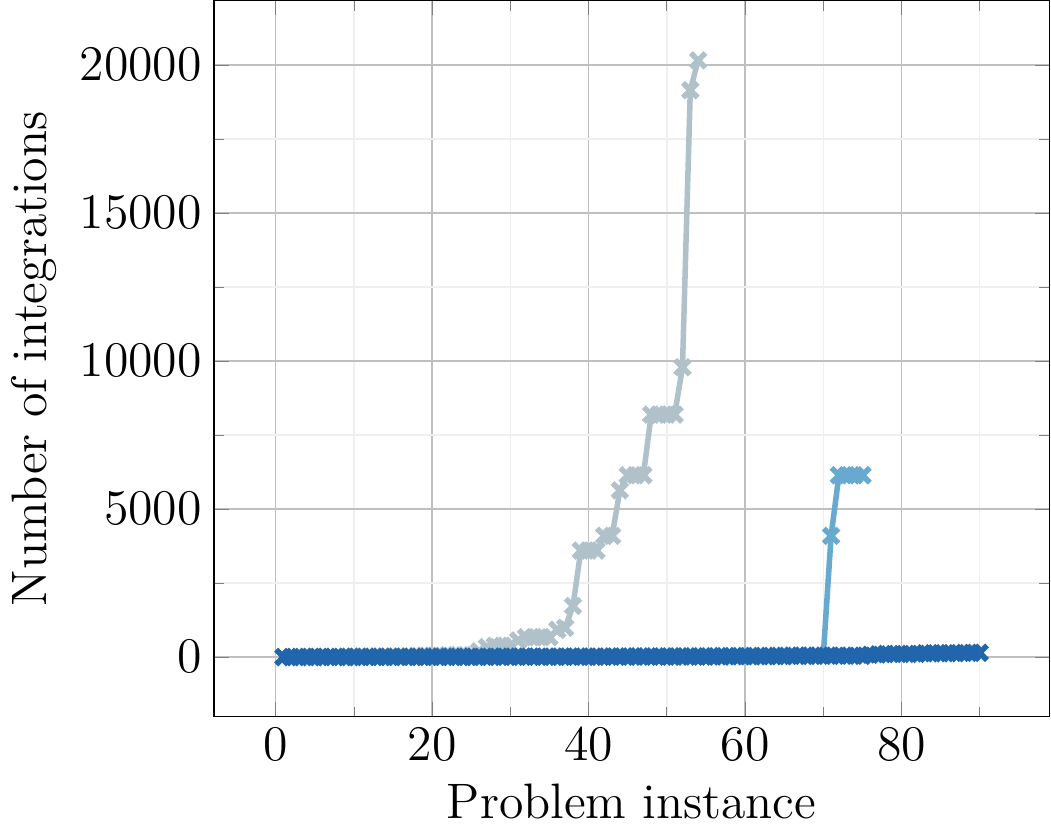}
  \caption{$H=0.5$}%
  \label{fig:plot_sapask_det_n_integrations_05}
   \end{subfigure} \par\bigskip
   \begin{subfigure}[b]{0.49\textwidth}
   \includegraphics[width=\textwidth]{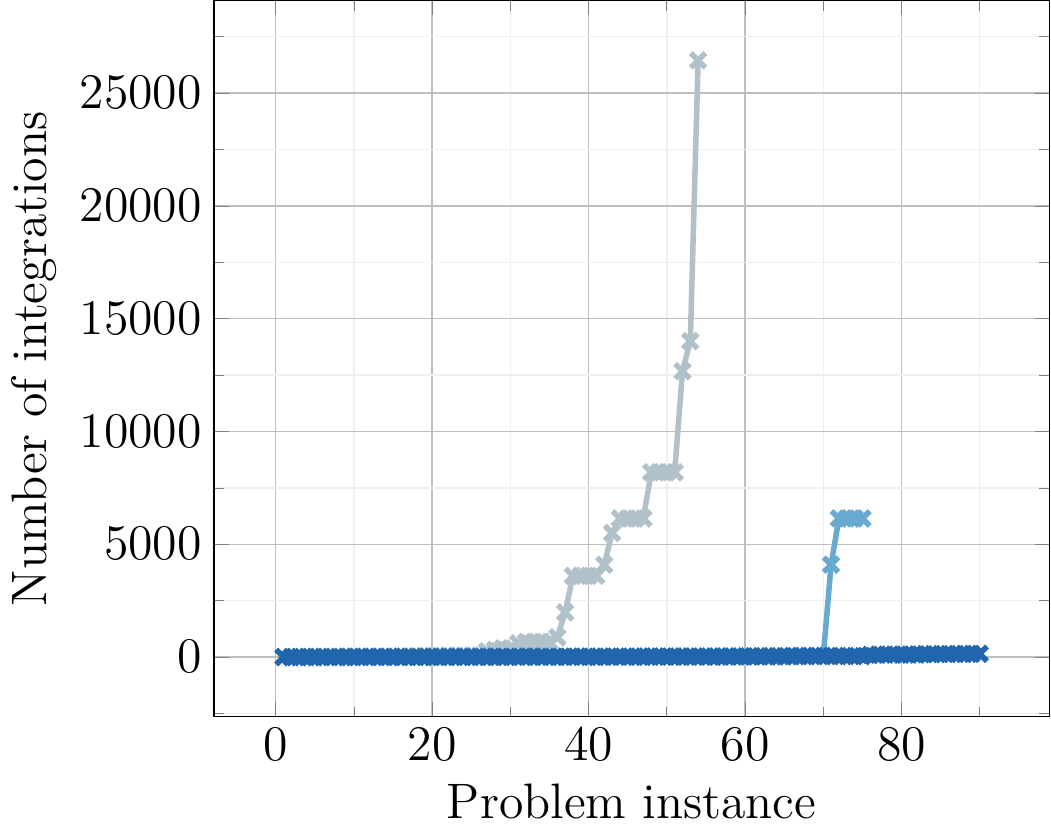}
  \caption{$H=0.75$}%
  \label{fig:plot_sapask_det_n_integrations_075}
   \end{subfigure} \hfill
   \begin{subfigure}[b]{0.49\textwidth}
   \includegraphics[width=\textwidth]{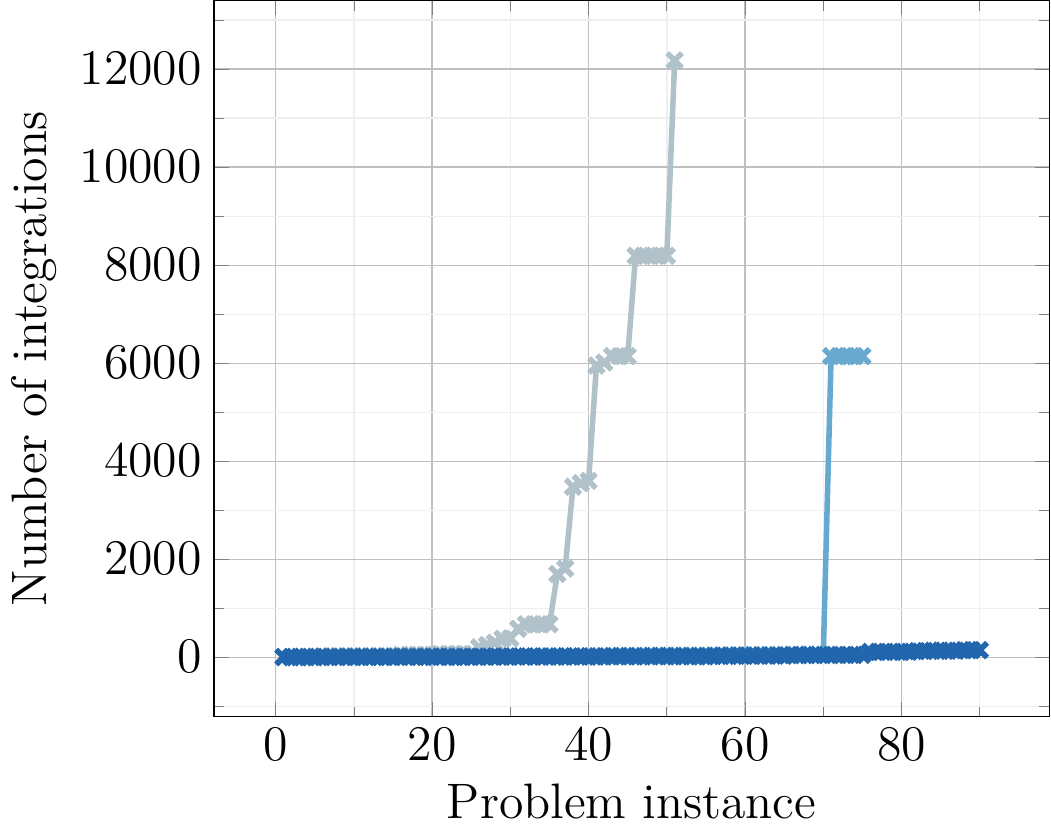}
  \caption{$H=1.0$}%
  \label{fig:plot_sapask_det_n_integrations_10}
   \end{subfigure}
   \caption{Cactus plots representing the number of computed integrals on a set of DET problems with $H \in \{0.25, 0.5, 0.75, 1\}$.
   }\label{fig:plot_sapask_det_int}
\end{figure}

Figure~\ref{fig:plot_sapask_det_time} depicts the runtime of the
algorithms for $H \in \{0.25, 0.5, 0.75, 1\}$. Thanks to the
absence of arithmetic operations in the internal nodes, DETs are more
suited for KC compared to the weight functions used
in~\sref{sec:exp-synth-exact}. While in the simplest inference cases
($H \le 0.5$) substantial factorization of the integrals is possible,
when the coupling between variables increases this advantage is
overweighted by the combinatorial reasoning capabilities of SMT-based
approaches.

The efficacy of \method{} with respect to both \wmisapa{} and \wmipa{} is particularly
evident on this batch of problems. Analysing the number of integrals
computed by PA-based algorithms (Figure~\ref{fig:plot_sapask_det_int})
we notice a drastic reduction in the number of integrals generated by
\method{}.  This behaviour is expected: the number of Boolean
atoms is typically very high, so that we are able to drastically
reduce the number of partial assignment over the Boolean component
with respect to \wmisapa{} thanks to the revised enumeration strategy.

\noindent Crucially, in contrast with the recent \wmisapa{}, the novel
algorithm is able to compete or even outperform KC-based solvers even
when the oblique queries involve a low number of variables.


\section{Extending \wmipa{} with different integration approaches}%
\label{sec:approximated-wmipa}


A key feature of PA-based solvers like \wmipa, which has been
inherited by our structure-aware procedures, is that of \emph{being
  agnostic of the integration procedure used as backend}.
%
%
%
Besides developing a novel structure encoding, we implemented a
modular approach to PA-based solvers, allowing the use of different
integration strategies, including approximate ones. This provides
multiple advantages. First, the general integration procedure can be
substituted with a specialized one, targeting specific weight families
such as piecewise-constants or leveraging structural properties like
full factorizability.
Second, approximate integration procedures can be applied when the
number of continuous variables makes exact integration of \FIUC{}
functions infeasible. Third, it broadens the use of these solvers
beyond the \FIUC{} family, to functions whose integral over convex
polytopes can be approximated with a finite sample, like multivariate
Gaussian distributions or even the output of a simulation or any other
black-box function.

We explored the practical benefits of this approach with a variant of
\method{} using Monte Carlo (MC) approximate
integration~\cite{newman1999monte}:
\begin{align}
   \myint{\mularat}{\allx}{\w(\allx)} &\approx
   \overbrace{\myint{\mularat}{\allx}{1}}^{\mathsf{Vol(\mularat)}}
   \:\cdot\: \mathbb{E}_{x \sim \mathcal{U}(\mularat)} [ \w(x) ]
\end{align}
In practice, our MC integration procedure is implemented via
\volesti~\cite{volesti}, a library for volume approximation and
sampling in convex polytopes. \volesti{} is used for both approximating
$\mathsf{Vol}(\mularat)$ and for sampling $\w$.
In stark contrast with fully approximate WMI
algorithms~\cite{BelleUAI15,abboud2020approximability}, in our case
the combinatorial reasoning is exact. This feature can be crucial in
contexts where it is acceptable to compute approximate probabilities
but the hard constraints cannot be violated. Considering the
difficulties of sampling in combinatorial spaces, this approach is
also preferable in problems where the discrete reasoning is feasible
but the integration is not.
%
%

\begin{figure}
    \centering
    \begin{subfigure}{\textwidth}
      \centering
      \resizebox{\textwidth}{!}{\begin{tikzpicture}
  \begin{customlegend}[
      legend columns=-1,
      legend style={
          column sep=1ex,
          align=center,
        },
      every axis plot/.append style={ultra thick, mark=x, mark size=3pt},
    ]
    \addlegendimage{mark=none, black, dashed} \addlegendentry{Timeout}
    \addlegendimage{WMIPA}\addlegendentry{Symbolic}
    \addlegendimage{WMISAPASK}\addlegendentry{Numerical}
    \addlegendimage{WMISAPASK_approx_100}\addlegendentry{Approx\\(N=100)}
    \addlegendimage{WMISAPASK_approx_1000}\addlegendentry{Approx\\(N=1000)}
    \addlegendimage{WMISAPASK_approx_10000}\addlegendentry{Approx\\(N=10000)}
    \addlegendimage{WMISAPASK_approx_100000}\addlegendentry{Approx\\(N=100000)}
  \end{customlegend}
\end{tikzpicture}}
    \end{subfigure}
    \begin{minipage}{.49\textwidth}      
      \begin{subfigure}{\textwidth}
      \centering
      \includegraphics[width=\textwidth]{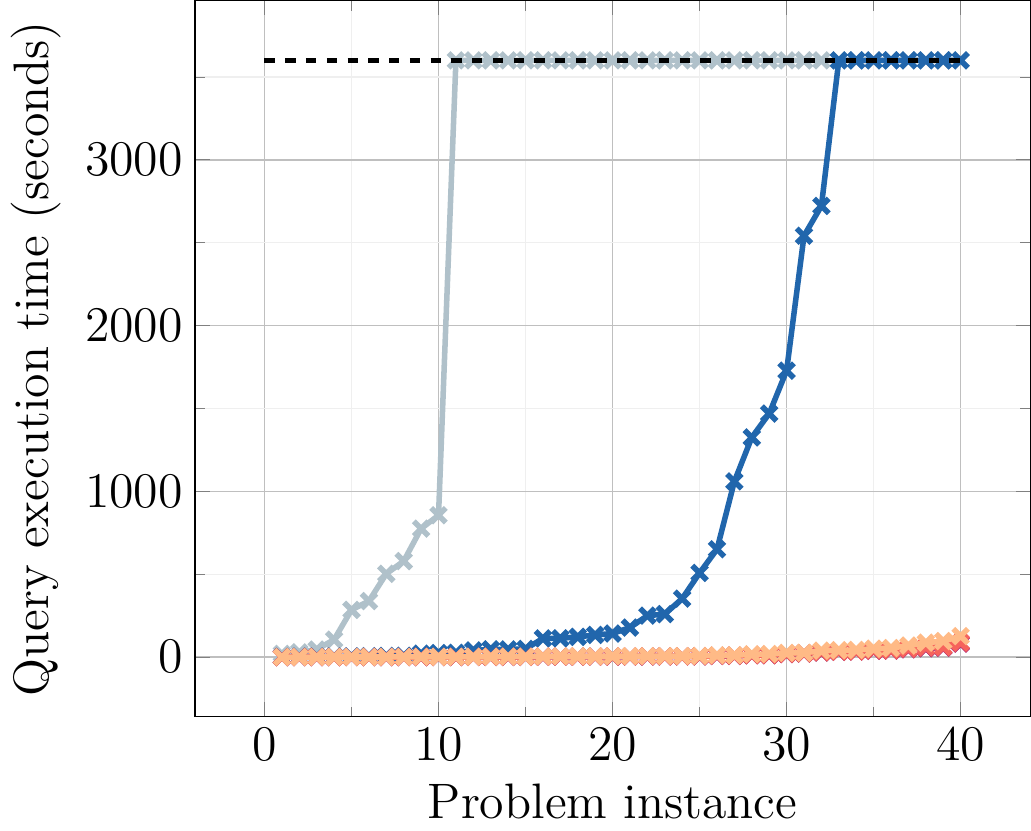}
    \end{subfigure}
    \begin{subfigure}{\textwidth}
      \centering
      \includegraphics[width=0.98\textwidth]{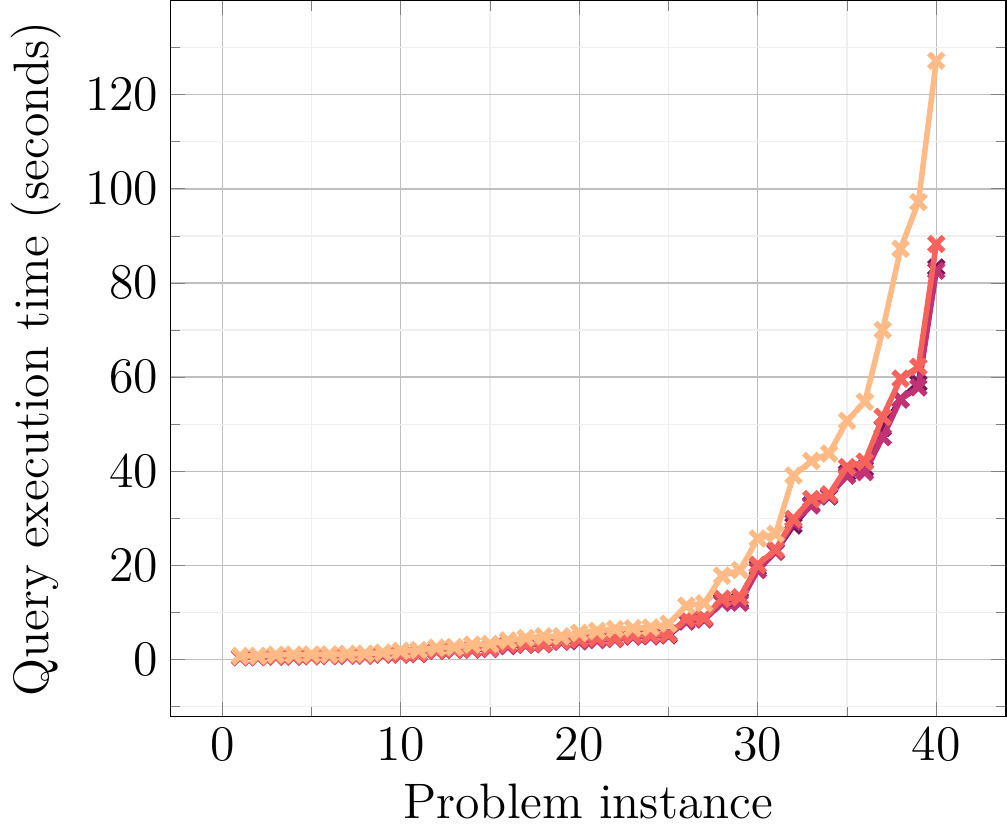}
    \end{subfigure}
  \end{minipage}
  \begin{minipage}{.49\textwidth}      
      \begin{subfigure}{\textwidth}
      \centering
      \includegraphics[width=0.985\textwidth]{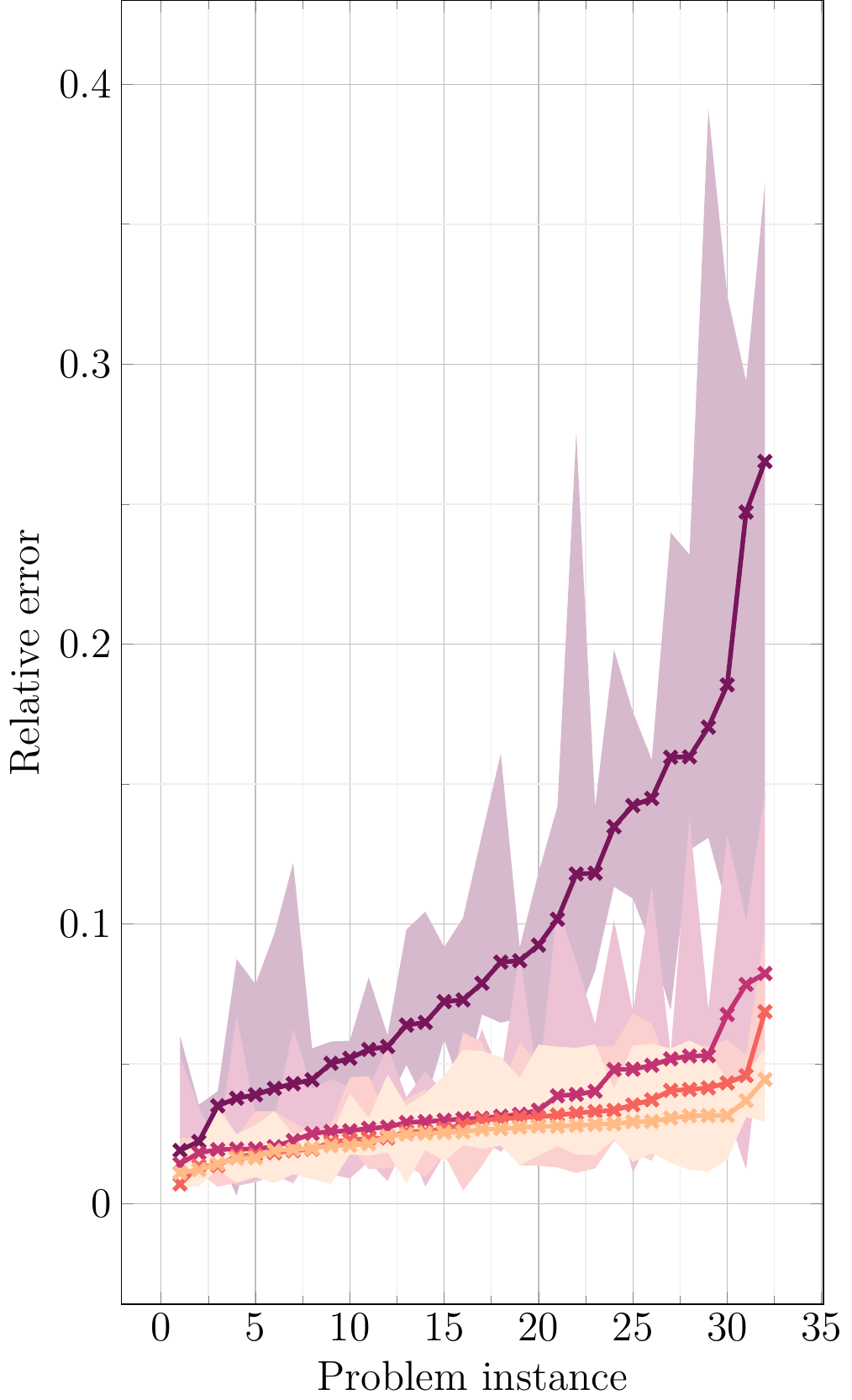}
    \end{subfigure}
  \end{minipage}
    \caption{Cactus plots comparing
      \method{} using different integration
      strategies on the synthetic datasets.
      For \method{}(\volesti), we plot a line representing the median, and we color the range between the 1st and 3rd quartiles of the outcomes obtained over 10 runs with different random seeds.
      Top-left: comparison of the execution time of \method{} when equipped with different integration strategies, i.e., symbolic, based on a XADD, numerical, based on \latte, and approximate, based on \volesti, with increasing number of samples $N\in\set{100, 1000, 10000, 100000}$. Bottom-left: zoom on the previous plot, showing only the results for \method{}(\volesti).
      Right: relative error of the approximate result computed by \method{}(\volesti) w.r.t.\ the exact result computed by \method{}(\latte), for the problems where the latter does not time out.
      }%
      \label{fig:latte-vs-volesti}
\end{figure}

We evaluated the resulting solver, dubbed \method{}(\volesti), on the
problems described in~\sref{sec:exp-synth-exact}. We compare
  it with the \method{} version using exact numerical integration,
  i.e., the one we used in \sref{sec:expeval}, which we refer to here
  as \method{}(\latte) to avoid any ambiguity.
Figure~\ref{fig:latte-vs-volesti} displays the computational
gain and the approximation error obtained with the above
\emph{approximate} MC integrator with respect to \latte's exact
\emph{numerical} approach.  We show that by increasing $N$,
  i.e., the number of samples to estimate $\mathbb{E}_{x \sim
    \mathcal{U}(\mularat)} [ \w(x) ]$, we can trade the approximation
  error with the execution time.\@ From
  Figure~\ref{fig:latte-vs-volesti} (top-left) we notice that the
  approximate method, \method{}(\volesti) with $N \in
  \set{100,1000,10000,100000}$, drastically outperforms the numerical
  one, \method{}(\latte).  By zooming into these plots, in
  Figure~\ref{fig:latte-vs-volesti} (bottom-left), we notice that the
  execution time of \method{}(\volesti) grows very little with
  $N$. This is due to the fact that sampling is very cheap, so that
  within \method{}(\volesti) the time for enumeration and for
  computing the integration volumes dominates over the time for
  sampling.  In Figure~\ref{fig:latte-vs-volesti} (right) we notice
  that the relative error reduces significantly by increasing
  $N$. Thus we have obtained relatively accurate results with very
  limited computational costs.

%
%
We additionally equipped \method{} with an
exact \emph{symbolic} integrator
based on a XADD. Noticeably, whereas symbolic
approaches were shown to be preferable in conjunction with KC-based
algorithms (see Figure 3 in~\cite{kolb2020exploit}) and do not assume
convex integration bounds, numerical integration performs substantially better with
our approach (Figure~\ref{fig:latte-vs-volesti} (top-left)).

\newcommand{\pop}{\textsf{pop}\xspace}
\newcommand{\dec}{\textsf{dec}\xspace}
\newcommand{\hire}{\textsf{hire}\xspace}
\newcommand{\minority}{\textsf{minority}\xspace}
\newcommand{\ethnicity}{\ensuremath{\mathit{ethnicity}}\xspace}
\newcommand{\colRank}{\ensuremath{\mathit{colRank}}\xspace}
\newcommand{\yExp}{\ensuremath{\mathit{yExp}}\xspace}

We additionally showcase a prototypical application of the modified
integration routine with Gaussian weights, borrowing the setting and
examples from FairSquare~\cite{albarghouthi2017fairsquare}.
In their setting, a (deterministic) \emph{decision program} \dec{} has
to decide whether to \hire{} a candidate or not, according to some
features. The input of \dec{} is distributed according to a Gaussian
probabilistic program, called the \emph{population model} \pop, which
additionally models sensitive conditions that are not accessed by
\dec, such as whether an individual belongs to a \minority{} group.
The goal is
quantifying the demographic parity of \dec{} under the distribution
\pop,
i.e., the ratio:
\begin{align}
  \label{eq:fair-ratio}
  \frac{\Pr(\hire \:|\: \minority)}{\Pr(\hire \:|\: \neg \minority)}
\end{align}
For motivational purposes,  we report our results on
the same introductory examples used in the Fairsquare paper, whose
detailed description is reported in~\ref{sec:fair-programs}.

\begin{figure}
    \centering
    \includegraphics[width=0.6\textwidth]{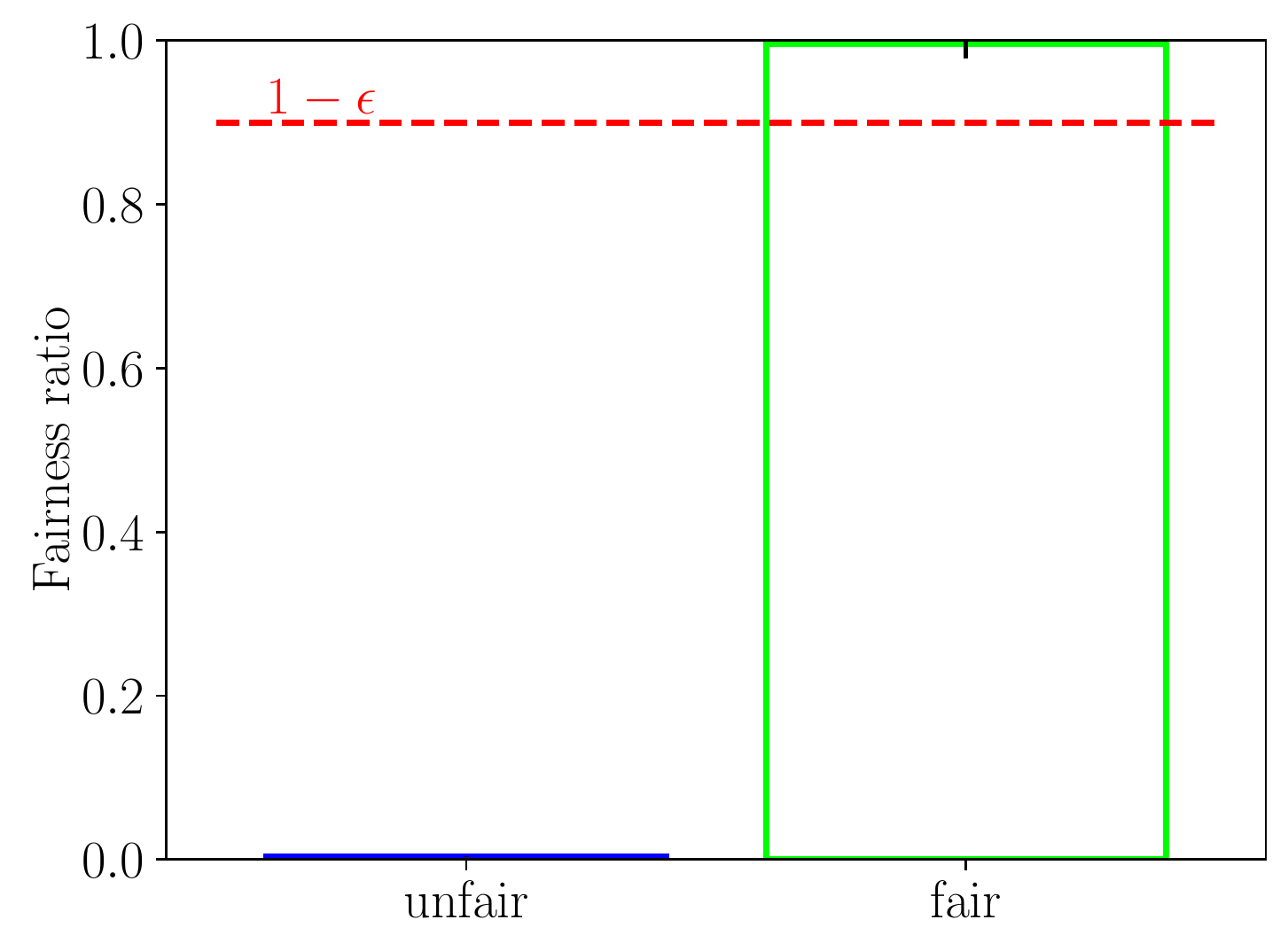}
    \caption{\label{fig:fair} The ratio in~\eqref{eq:fair-ratio}
      computed for the initial (unfair) \dec{} and the modified
      one. Average and standard deviation over $5$ runs are reported,
      confirming the results reported in the FairSquare paper with
      high confidence.}
\end{figure}

In contrast with our approach, FairSquare only certifies whether the
ratio in~\eqref{eq:fair-ratio} is greater than a user-defined
threshold $1 - \varepsilon$.
With \method(\volesti), we could indeed observe that the initial
\dec{} does not respect the fairness criterion with $\varepsilon =
0.1$. This problem was addressed by making \dec{} less sensitive to a
feature that was directly influenced by the sensitive condition. This
second version of the program is fair with high probability, as
reported in Figure~\ref{fig:fair}.

A current limitation of this approach is that \volesti{} does not directly
offer statistical guarantees on the relative error. In problems like
volume approximation~\cite{chalkis2019practical} and
sampling~\cite{chalkis2020geometric}, practical statistical bounds on
the error were obtained using diagnostics like the empirical sample
size (ESS) or the potential scale reduction factor
(PSRF)~\cite{gelman1992inference}. The adoption of similar mechanisms
in our MC integrator, a fundamental step towards the application of
WMI in formal verification, and the evaluation of WMI-based verifiers
on more realistic programs, is left for future work.

\section{Conclusion}%
\label{sec:concl}


The difficulty of dealing with densely coupled problems has been a
major obstacle to a wider adoption of hybrid probabilistic inference
technology beyond academia. In this paper, we made a significant step
towards removing this obstacle. Starting from the identification of
the limitations of existing solutions for WMI, we proposed a novel
algorithm combining predicate abstraction with weight-structure
awareness, thus unleashing the full potential of SMT-based
technology. An extensive experimental evaluation showed the advantage
of the proposed algorithm over current state-of-the-art solutions on
both synthetic and real-world problems. These include inference tasks
over constrained density functions learned from data, which highlight
the potential of the approach as a flexible solution for probabilistic
inference in hybrid domains.

Whereas the improvements described in this work drastically reduce the number of integrals required to perform WMI, the integration itself can be a bottleneck for scalability. To deal with this issue, we showed how SMT-based approaches can seamlessly
incorporate multiple integration strategies, including symbolic,
numeric and approximate ones, allowing to choose the most appropriate
strategy for the problem at hand. In particular, e.g., using approximate MC integration
techniques could drastically improve performances at the cost of
limited losses in terms of precision.

We further showcased the potential utility of our approach on a prototypical application aimed at verifying the fairness of probabilistic programs, as a first step towards the application of WMI technology to the probabilistic verification of hybrid systems and machine learning models.

\section*{Acknowledgments}

This research was supported by TAILOR, a project funded
by the EU Horizon 2020 research and innovation program under GA No
952215.
We acknowledge the support of the MUR PNRR project FAIR -- Future AI Research (PE00000013) , under the NRRP MUR program funded by the NextGenerationEU.
The work was also partially supported by the
  project ``AI@TN'' funded by the Autonomous Province of Trento. 

\FloatBarrier%
\bibliography{rs_refs}

\newpage
\appendix
\section{Detailed Technical Background}%
\label{sec:detailedbackground}

\subsection{SMT:\@ Notation, Syntax, and Semantics}%
\label{sec:detailed-background-smt}

We assume the reader is familiar with the basic syntax, semantics, and
results of propositional and first-order logic.
We adopt the notation and definitions in~\cite{morettin-wmi-aij19} ---including some terminology and concepts from Satisfiability Modulo Theories
(SMT)--- which we summarize below.

\paragraph{Notation and terminology}
Satisfiability Modulo Theories (SMT) (see~\cite{barrettsst21}  for details) consists in deciding the
satisfiability of quantifier-free
first-order formulas over some given background theory \T. For the
context of this paper,
we restrict to the theory of linear real arithmetic (\larat{}), possibly combined with
 uninterpreted function symbols ($\larat{}\cup\euf{}$), s.t.\ hereafter we assume $\T\in\set{\larat,\larat\cup\euf}$.
%
We use $\Bool \defas \{\top, \perp\}$ to indicate the set of Boolean
values and $\mathbb{R}$ to indicate the set of real
values.
We use the sets $\allA\defas\set{A_i}_i$,
$\allB\defas\set{B_i}_i$ to denote Boolean atoms and
the sets $\allx\defas\set{x_i}_i$,
$\ally\defas\set{y_i}_i$ to denote real variables.
SMT(\larat{}) formulas combines Boolean atoms $A_i \in
\Bool$ and \larat{} atoms in the form $(\sum_i c_i x_i\ \bowtie c)$
(where $c_i$, $c$ are rational values, $x_i$ are real variables in
$\mathbb{R}$ and $\bowtie$ is one of the standard arithmetic operators
$\{=, \neq, <, >, \leq, \geq\}$)  by using standard Boolean operators
$\{\neg,\wedge,\vee,\rightarrow,\leftrightarrow\}$.
In \smt($\larat\cup\euf$), \larat{} terms can be interleaved with uninterpreted function symbols.
\atoms{\vi} denotes the set of atomic formulas occurring in $\vi$, both Boolean%
\ and \T-atoms.
%

%

  A 
  \emph{literal} is either an 
atom (a \emph{positive literal}) or its negation (a \emph{negative
  literal}).
A \emph{clause}
$\bigvee_{j=1}^{K} l_{j}$ is a disjunction of literals.
A formula is in \emph{Conjunctive Normal Form (CNF)} if it is
  written as a conjunction of clauses
  $ \bigwedge_{i=1}^{L}\bigvee_{j_i=1}^{K_i} l_{j_i} $.
Some shortcuts of frequently-used expressions (marked as
``\shortcut{\ldots}'') are provided to simplify the reading:
the formula $(x_i
\geq l) \wedge (x_i \leq u)$ is shortened into
$\inside{x_i}{[l,u]}$; if $\phi\defas\bigwedge_i C_i$ is a CNF formula
and $C\defas\bigvee_j l_j$ is a clause, then  the formula
$\bigwedge_i(C\vee C_i)$ is shortened into $\shortcut{C\vee\phi}$.      

\paragraph{Semantics}
 ``$\Tmodels$'' 
denotes entailment in $\T$ (e.g., \mbox{$(x\ge 2)\laratmodels (x\ge 1)$}),
\DROPINMAIN{%
and 
``$\pmodels$'' denotes tautological entailment,
i.e., purely-Boolean entailment
treating \T{} atoms as Boolean
propositions  (e.g.
$A_1\wedge(x \ge 2)\pmodels(A_1\vee (x\le 1))\wedge(\neg
A_1\vee(x\ge 2))$, since 
$A_1\wedge A_2\models(A_1\vee A_3)\wedge(\neg
A_1\vee A_2)$.)}
\DROPINMAIN{Notice that $\pmodels$ is strictly stronger than $\Tmodels$, that
is, if $\vi_1\pmodels\vi_2$ then $\vi_1\Tmodels\vi_2$, but not
vice versa.}
We say that $\vi_1,\vi_2$ are \T-equivalent
\DROPINMAIN{written $\vi_1 \Tequiv{} \vi_2 $,}
iff both $ \vi_1\Tmodels\vi_2$
and $ \vi_2\Tmodels\vi_1$,
\DROPINMAIN{and \emph{tautologically
  equivalent}
\DROPINMAIN{, written $\vi_1 \boolequiv{} \vi_2 $,}
iff both $ \vi_1\pmodels\vi_2$
and $ \vi_2\pmodels\vi_1$.}
\DROPINMAIN{(We use ``$\models$'' rather than ``$\pmodels$'' and
 ``$\Tmodels$'' when the distinction is clear from the context or
 is irrelevant.)}

%
%
%

Given a set of Boolean and \T-atoms $\allpsi\defas\set{\psi_1,\ldots,\psi_K}$, we 
call a \emph{total \resp{partial} truth assignment} $\mu$ for \allpsi{}
any total \resp{partial} map from \allpsi{} to $\Bool$.
(These definitions can be used also for sets \allpsi{} of non-atomic \T{}-formulas.)
{We denote by $\Bool^K$ the set of all total truth assignments over \allpsi.}
With a little abuse of notation, 
we represent $\mu$
either as a set or a conjunction of literals:
$\mu \defas \{\psi\ |\ \psi \in \allpsi{},\ \mu(\psi)=\top \} 
\cup \{\neg\psi\ |\ \psi \in \allpsi{},\ \mu(\psi)=\bot \}$, or
$
\mu \defas \bigwedge_{\psi \in \allpsi{},{\mu(\psi)=\top}} \pos\psi
\wedge
\bigwedge_{\psi \in \allpsi{},{\mu(\psi)=\bot}} \neg\psi
$,
and we write ``$\psi_i\in\mu_1$'' and ``$\mu_1\subseteq\mu_2$'' as if 
$\mu_1,\mu_2$ were represented as sets (i.e., we write
``$\psi_1\wedge\neg \psi_2\subseteq \psi_1\wedge\neg \psi_2\wedge \psi_3$'' 
meaning 
``$\set{\psi_1,\neg \psi_2}\subseteq \set{\psi_1,\neg \psi_2,
  \psi_3}$'').
\DROPINMAIN{In the latter case, we say that $\mu_1$ is a \emph{sub-assignment} of
$\mu_2$,
and that $\mu_2$ is a \emph{super-assignment} of $\mu_1$.}
We denote by \mua{} and \muT{} its two components on the Boolean atoms in \allA{} and on the \T atoms, respectively, so that 
$\mu\defas\mua\wedge\muT$.

\begin{example}
    If $\mu\defas \set{A_1=\top,A_2=\bot,(x \ge 1)=\top,(x \ge 3)=\bot}$
then we represent  it as the set $\mu\defas \set{A_1, \neg A_2 , (x \ge
1) , \neg (x \ge 3)}$
or as the formula $\mu\defas A_1\wedge \neg A_2 \wedge (x \ge
1) \wedge \neg (x \ge 3)$.
Hence,
$\mua\defas A_1\wedge \neg A_2$ and
$\mularat\defas (x \ge 1) \wedge \neg (x \ge 3)$.\hfill $\diamond$
\end{example}

Given a (partial) truth assignment $\mu$ to \atoms{\vi},  we call the
\emph{residual} of $\vi$ w.r.t.\ $\mu$, written ``$\vimuagen{\vi}{\mu}$'',
the formula obtained from $\vi$ by substituting all the atoms assigned
in $\mu$ with their respective truth value, and by recursively propagating truth
values through Boolean operators in the usual way.\footnote{E.g.,
  $\neg\top\thus \bot$, 
  $\vi\wedge\top \thus \vi$, 
  $\vi\vee\top \thus \top$,
  $\vi\iff \top \thus \vi$,
  $\vi\iff \bot \thus \neg\vi$,
  \ldots}
%
We say that $\mu$ \emph{propositionally satisfies} $\vi$ iff $\vimuagen{\vi}{\mu}$
reduces to $\top$.%
\footnote{%
  The definition of satisfiability by partial assignment may
    present some ambiguities for non-CNF and existentially-quantified
    formulas~\cite{sebastianipartial2020,mohlesb20}. Here we adopt the
    above definition because it is 
    the one typically used by state-of-the-art solvers.}
Thus, 
if $\mu$ propositionally satisfies $\vi$, then all total
assignments $\eta$ s.t. $\mu\subseteq \eta$ propositionally satisfy $\vi$.
\DROPINMAIN{  (Notice that if $\vimuagen{\vi}{\mu}=\top$, then
  $\mu\pmodels\vi$, but not vice versa, see~\cite{sebastianipartial2020,mohlesb20} for details.)}
Since $\mu$
 can be \T-unsatisfiable due to its \T{} component
(e.g., $\mu\defas\neg A_1\wedge(x_1+x_2=3)\wedge\neg(x_1+x_2\ge 2)$\ is \larat{}-unsatisfiable),
 \smtT{} checks the existence of an \T-satisfiable assignment $\mu$
 propositionally satisfying $\vi$.


  \begin{example}
    Let
 $ 
 \vi \defas(A_1\vee (x\le 1))\wedge(\neg
 A_1\vee(x\ge 2))$, $\mu\defas\muA\wedge\mularat$ where
 $\muA\defas A_1$ and $\mularat\defas(x \ge 2)$.
 Then $\vimuagen{\vi}{\muA}=(x\ge 2)$ and $\vimuagen{\vi}{\mu}=\top$,
 s.t.\ $\mu$ propositionally satisfies $\vi$.
 Since $\mularat$ is $\larat$-satisfiable, then $\vi$ is
 $\larat$-satisfiable. \hfill $\diamond$
  \end{example}%

 
 Given \vixab, we say that a Boolean (partial) assignment $\mua$ on
$\allA$ propositionally satisfies 
 $\exists \allx. \exists \allB. \vi$
 iff $\mu\defas\mua\wedge\mub\wedge\mularat$ propositionally satisfies $\vi$
 for some total assignment $\mub$ on \allB{} and some total
 \larat-satisfiable assignment $\mularat$ on the \larat{} atoms of
 $\vi$.
To this extent, notice that $\exists\allx. \exists
   \allB. \vi$ is equivalent to some purely-Boolean formula on $\allA$ only.
 The definition of ``$\mua$ propositionally satisfies $\exists\allx.  \vixa$'' follows with $\allB=\emptyset,\mub=\top$.
Similarly, we also say that $\muA\wedge\mularat$ propositionally
satisfies $\exists \allB.\vixab$ iff $\muA\wedge\mub\wedge\mularat$
propositionally satisfies $\vi$ for some total assignment $\mub$ in
$\allB$.
%
Finally, given \vixya{} and a (partial) truth assignment $\munoy$ on
the \ally-free atoms in \vi{}, we say that $\munoy$ propositionally
satisfies $\exists\ally.\vi$ iff there exists a total truth assignment
$\muy$ on the remaining atoms of $\vi$ s.t $\munoy\wedge\muy$ is
\T-satisfiable and it propositionally satisfies $\vi$.
%

\begin{example}
   Let
 $\vi \defas(A_1 \vee (x\le 1))\wedge  B_1 \wedge(B_1\iff 
 (x\ge 2)) $.
 Then $\muA_1\defas A_1$ propositionally satisfies $\exists x. \exists B_1 .\vi$ because there
 exist $\mub_1\defas B_1$  and $\mularat_1\defas\neg(x\le 1)\wedge(x \ge 2)$
  s.t.  $\mu_1\defas\muA_1\wedge\mularat_1\wedge\mub_1$ propositionally
  satisfies $\vi$ and $\mularat_1$ is \larat-satisfiable.
%
Instead, $\muA_2\defas \neg A_1$ does not satisfy $\exists x. \exists
B_1 .\vi$, because no such \mub{} and  \mularat{} exist.
Indeed, $\exists x. \exists B_1 .\vi$ is equivalent to the Boolean
formula $A_1$.
Also, $\mua_1\wedge\mularat_1$ propositionally satisfies $\exists B_1.\vi$.
 \hfill $\diamond$  
   \end{example}%

\subsection{Assignment Enumeration}
$\TTA{\vi} \defas \{\mu_1, \ldots, \mu_j \}$ denotes the
set of \T{}-satisfiable \textit{total} truth assignments over $Atoms(\vi)$ that propositionally satisfy $\vi$;
$\TA{\vi} \defas \{\mu_1, \ldots, \mu_j \}$ represents one set of \T{}-satisfiable
\emph{partial} truth assignments over $Atoms(\vi)$ that
propositionally satisfy $\vi$, s.t.:
\begin{enumerate}[(i)]  
\item\label{detailedbackground:item:ta:cover} for every total assignment $\eta$ in $\TTA{\vi}$ there is some partial one $\mu$ in $\TA{\vi}$ s.t.\ $\mu\subseteq\eta$ and
\item\label{detailedbackground:item:ta:disjoint} every pair $\mu_i,\mu_j\in \TA{\vi}$ assigns
opposite truth values to at least one atom.
\end{enumerate}
We remark that $\TTA{\vi}$ is
 unique (modulo reordering), whereas multiple $\TA{\vi}$s are admissible for the same formula $\vi$ (including $\TTA{\vi}$).
{The disjunction of the truth assignments in \TTA{\vi}, and that of
  \TA{\vi}, are \T-equivalent to \vi.}
Thus, given $\vixab$, $\TTA{\exists \allx. \exists \allB. \vi}$ denotes the set of all
total truth assignment $\mua{}$ on $\allA$ s.t.\ $\mua$ propositionally satisfies $\exists
\allx. \exists \allB. \vi$, and $\TA{\exists \allx. \exists
  \allB. \vi}$ denotes one set of partial assignments \mua{} on
\allA{} s.t.\ $\mua$ propositionally satisfies $\exists \allx. \exists \allB. \vi$ 
complying with conditions~(\ref{detailedbackground:item:ta:cover}) and~(\ref{detailedbackground:item:ta:disjoint}) above where $\vi$ is replaced
by $\exists
\allx. \exists \allB. \vi$. 
\DROPINMAIN{The disjunction of the assignments in
  \TTA{\exists \allx. \exists \allB. \vi} and that of
\TA{\exists \allx.  \exists \allB. \vi} are \T{}-equivalent to $\exists \allx.  \exists \allB. \vi$.}
(As above, the definition of
$\TTA{\exists \allx. \vixa}/\TA{\exists \allx. \vixa}$ follows with $\allB=\emptyset$.)
  Similarly, \TTA{\exists\allB.\vixab} denotes the set of all
  \T-satisfiable 
  total truth assignments $\muA\wedge\mularat$ that
  propositionally satisfy $\exists\allB.\vixab$, and
  \TA{\exists\allB.\vixab} denotes one set of
  \T-satisfiable 
  partial truth assignments $\muA\wedge\mularat$ that propositionally satisfy $\exists\allB.\vixab$ complying with conditions~(\ref{detailedbackground:item:ta:cover}) and~(\ref{detailedbackground:item:ta:disjoint}) above.\@
Finally \TTA{\exists\ally.\vixya} denotes the set of all total truth
assignments $\munoy$ on the \ally-free atoms in \vi{} which
propositionally satisfy $\exists\ally.\vixya$, and
\TA{\exists\ally.\vixya} denotes a set of partial such assignments
complying with conditions~(\ref{detailedbackground:item:ta:cover})
and~(\ref{detailedbackground:item:ta:disjoint}) above.  The
disjunction of the assignments in \TTA{\exists\ally.\vixya} and that
of \TA{\exists\ally.\vixya} are \T{}-equivalent to
$\exists\ally.\vixya$.

\TTA{\ldots}/\TA{\ldots}  can be 
computed efficiently by means of {\emph{Projected AllSMT}}, 
a technique used in formal verification to compute \emph{Predicate Abstraction}~\cite{allsmt}.
All these functionalities are provided by the SMT solver
\mathsat{}~\cite{mathsat5_tacas13}.
\DROPINMAIN{%
In a nutshell, \mathsat{} works as follows.
Given $\vi$ and a subset of ``relevant'' atoms
$\allpsi\subseteq\atoms{\vi}$, \TA{\ldots} generates
one-by-one \T-satisfiable partial assignments
$\mu_i\defas\mu_i^{\overline{\Psi}}\wedge\mu_i^{\Psi}$,
s.t.\ $\mu_i^{\overline{\allpsi}}$ is a total assignment on $\atoms{\vi}\!\setminus\!\allpsi$
and $\mu_i^{\allpsi}$ is a \emph{minimal}~%
\footnote{i.e., no literal can be further dropped from $\mu_i^{\allpsi}$
  without losing properties~(\ref{item:msatenum:sat}) and~(\ref{item:msatenum:disjoint}).}
partial assignment on
\allpsi{} s.t.:
\begin{enumerate}[(i)]
  \item\label{item:msatenum:sat} $\vimuagen{\vi}{\mu_i}=\top$, and 
  \item\label{item:msatenum:disjoint} for every
  $j\in\set{1,\dots,i-1}$, $\mu_j^{\allpsi},\mu_i^{\allpsi}$ assign
  opposite 
  truth values to at least one atom in \allpsi.
\end{enumerate}
Finally, the set
$\set{\mu_i^{\allpsi}}_i$ is returned.
(We say that the enumeration is \emph{projected over \allpsi}.)
  Thus:
  \begin{itemize}
    \item \TA{\vi} can be computed by setting $\allpsi\defas\atoms{\vi}$;
    \item \TA{\exists\allB.\vixab} can be computed by setting $\allpsi\defas\atoms{\vi}\setminus\allB$;
    \item \TA{\exists \allx.\exists\allB.\vixab}, possibly with $\allB=\emptyset$, can be computed by setting $\allpsi\defas\allA$;
    \item \TA{\exists\ally.\vixya} can be computed by setting
      $\allpsi\defas\atoms{\vi}\setminus\allpsi^\ally$,
      $\allpsi^\ally$ being
      the set of atoms containing \ally's.
  \end{itemize}
%
\noindent
\TTA{\dots} works in the same way, but forcing the $\mu_i$s  to be total.}

\subsection{CNF-ization}%
\label{sec:detailed-cnfization}
     A formula $\vixa$ can be converted into CNF as follows:
  implications and bi-implications are rewritten by applying the
  rewrite rules
  $(\alpha\imp\beta) \thus (\neg\alpha\vee \beta)$ and
  $(\alpha\iff\beta) \thus (\neg\alpha\vee \beta)\land(\neg\beta\vee\alpha)$;
  negations are pushed down to the literal level by recursively applying the
  rewrite rules
  $\neg(\alpha\wedge\beta)\thus(\neg\alpha\vee\neg\beta)$,
  $\neg(\alpha\vee\beta)\thus(\neg\alpha\wedge\neg\beta)$, and
  $\neg\neg\alpha\thus\alpha$. Then we have a few alternatives:\\
  \emph{Classic CNF-ization} (``\CNF{\vi}'') consists in applying recursively the DeMorgan rewrite
    rule $\alpha\vee(\beta\wedge\gamma)\thus
    (\alpha\vee\gamma)\wedge(\beta\vee\gamma)$ until the result is in
    CNF.\@ The resulting formula $\phi(\allx,\allA)$ is tautologically equivalent to $\vi$, but its size
    can be exponential w.r.t.\ that of $\vi$;\\
  \emph{Tseitin CNF-ization}~\cite{tseitin68}  (``\CNFtseitin{\vi}'') consists in applying recursively
    the ``labelling'' rewrite rule $\vi\thus \vi[\psi|B]\wedge
    (B\iff \psi)$ ---$\vi[\psi|B]$ being the results of substituting
    all occurrences of
    a subformula $\psi$ with a fresh Boolean atom $B$--- until
    all conjuncts can be CNF-ized classically without space blow-up.
    Alternatively, one can apply the rule $\vi\thus \vi[\psi|B]\wedge
    (B\imp \psi)$~\cite{plaisted1986structure}
    (``\CNFplaisted{\vi}''). 
    With both cases, the resulting formula $\phi(\allx,\allA\cup\allB)$
    is s.t.\ $\vixa$ is tautologically equivalent to $\exists
    \allB.\phi(\allx,\allA\cup\allB)$%
    ---that is, $\mu\models\vi$ iff there exists a total truth
    assignment $\eta$ on \allB s.t.\ $\mu\cup\eta\models\phi$---
    \allB being the set of fresh
    atoms introduced, and the size of $\phi$ is linear w.r.t.\ that of
    $\vi$.

    Notice that, if $\vi(\allx,\allA\cup\allB)$ is the result of applying one
of the ``labelling'' CNF-izations~\cite{tseitin68,plaisted1986structure} to $\phi(\allx, \allA)$, then
\TTA{\phi}, \TA{\phi}, \TTA{\exists \allx.\phi} and \TA{\exists
  \allx.\phi} can be computed as
\TTA{\exists \allB. \vi}, \TA{\exists \allB. \vi}, \TTA{\exists
  \allx.\exists \allB. \vi } and \TA{\exists \allx.\exists \allB. \vi}  respectively.

  \section{Implicit Enumeration of a Conditional Skeleton}%
  \label{sec:implicitskeleton}
{Here we report in detail the first preliminary approach we proposed
in~\cite{spallitta2022smt}.}
\subsection{Encoding}
We describe the encoding of a conditional skeleton that we proposed in~\cite{spallitta2022smt}. In the paper, the conditional skeleton \skw{} is not built explicitly, rather it is generated
\emph{implicitly} as a disjunction of
partial assignments over \conditionset{}, which we enumerate progressively.
To this extent, we first define $\skw{}\defas\exists\ally.\wenc{}$ where 
\wenc{} is a formula on $\allA,\allx,\ally$
s.t.\ $\ally\defas\set{y,y_1,\dots,y_k}$ is a set of
fresh \larat{} variables.
Thus, 
$\TA{\exists \allx.(\vi\wedge\supportwff\wedge\exists\ally.\wenc{})}$
can be computed as 
$\TA{\exists \allx\ally.\vistarstar}$, where $\vistarstar\defas(\vi\wedge\supportwff\wedge\wenc{})$,
because the \ally{}s do not occur in $\vi\wedge\supportwff$,
with no need to generate \skw{} explicitly.
The enumeration of 
\TA{\exists\allx\ally. \vistarstar} is performed by
the very same SMT-based procedure used in~\cite{morettin-wmi-aij19}.

\wenc{} is obtained by taking $(y=\w)$, s.t.\ $y\in\ally$ is fresh, and
recursively substituting bottom-up every
conditional term $({\sf If}\ \psi_i\ {\sf Then}\ t_{i1}\ {\sf Else}\
{t_{i2}})$ 
in it with a fresh variable $y_i\in\ally$, adding the
definition of 
$(y_i=({\sf If}\ \psi_i\ {\sf Then}\ t_{i1}\ {\sf Else}\
{t_{i2}}))$:
\begin{eqnarray}
  \label{eq:ite}
 (\neg \psi_i\vee y_i=t_{i1})\wedge (\psi_i\vee
y_i=t_{i2}).
\end{eqnarray}
%
This labelling\&rewriting process, which is inspired to Tseitin's
labelling CNF-ization 
\citep{tseitin68}, guarantees that the size of \wenc{} is linear w.r.t.\ that of \w{}.
E.g., with~\eqref{eq:prodite}, 
$\wenc$ is $(y=\prod_{i=1}^N
y_i)\wedge\bigwedge_{i=1}^N((\neg\psi_i\vee
y_i=w_{i1}(\allx))\wedge(\psi_i\vee y_i=w_{i2}(\allx)))$. 

\begin{algorithm}[t!]
  \caption{
    \convert{}$(\term, \conds{})$\\
    returns \tuple{\w',\defs,\newvars} \\
    \w': the term \w{} is rewritten into\\
    \conds{}: the current partial assignment to conditions \conditionset, \\representing
    the set of conditions which $\w$ depends on\\
    \defs{}: a conjunction of definitions in the form $y_i=\w_i$ needed to
    rewrite \w{} into \w{}'\\
    \ally{}: newly-introduced variables labelling if-then-else terms\\
  $f^g, f^{\bowtie}$: uninterpreted function naming the
  function (operator) $g$ ($\bowtie$)
    \\[.5em]
    called as $\tuple{w', \eufwenc{}, \ally} \gets \convert{}(w, \emptyset)$
}\label{algo:convert}

\begin{algorithmic}[1]
  \IF{(\{$\term$ constant or variable\})}
  \RETURN\tuple{\term,\top{},\emptyset{}}
  \ENDIF%
  \IF{($\term==(\term_1 \bowtie \term_2)$, $\bowtie\ \in\set{+,-,\cdot}$)}
  \STATE $\tuple{\term_i',\defs_i,\newvars_i}=\convert{}(\term_i,\conds{})$, $i\in{\{1,2\}}$
  \RETURN\tuple{f^{\bowtie} (\term_1',\term_2'),\defs_1\wedge\defs_2,\newvars_1\cup\newvars_2}
  \ENDIF%
  \IF{($\term==g(\term_1,\ldots,\term_k)$, $g$ unconditioned)}
  \STATE $\tuple{\term_i',\defs_i,\newvars_i}=\convert{}(\term_i,\conds{})$, $i\in{1,\ldots,k}$
  \RETURN\tuple{f^g (\term_1',\ldots,\term_k'),\bigwedge_{i=1}^k \defs_i,\cup_{i=1}^k \newvars_i}
  \ENDIF%
  \IF{($\term==({\sf If}\ \psi\ {\sf Then}\ \term_{1}\ {\sf Else}\ {\term_{2}})$)}
  \STATE $\tuple{\term_1',\defs_1,\newvars_1}=\convert{}(\term_1,\conds{}\cup\set{\pos\psi})$
  \STATE%
  $\tuple{\term_2',\defs_2,\newvars_2}=\convert{}(\term_2,\conds{}\cup\set{\neg\psi})$
  \STATE {\bf let} $y$ be a fresh variable
  \STATE $\defs=\defs_1\wedge\defs_2\ \wedge$ 
  \STATE \ \ \ \ $(\bigvee_{\psi_i\in\conds}\neg\psi_i\vee \neg \psi\vee (y=\term_1'))\ \wedge$
  \STATE \ \ \ \ $(\bigvee_{\psi_i\in\conds}\neg\psi_i\vee\pos \psi\vee (y=\term_2'))\ \wedge$
  \STATE \ \ \ \ $(\bigvee_{\psi_i\in\conds}\neg\psi_i\vee\neg(y=\term_1')\vee\neg(y=\term_2'))$
\STATE $\newvars = \newvars_1\cup\newvars_2\cup\set{y}$ 
  \RETURN\tuple{y,\defs,\newvars}
  \ENDIF%
\end{algorithmic}

\end{algorithm}

One problem with the above definition of \wenc{} is that it is not a
\larat{} formula, because $\w$ may include multiplications or even
transcendental functions out of the conditions \conditionset\footnote{The
  conditions in \conditionset{} contain only Boolean atoms or linear terms by definition of \FIUC{}.}, which makes SMT reasoning over it
dramatically hard or even undecidable. 
We notice, however, that when computing 
$\TA{\exists \allx\ally.\vistarstar}$
the arithmetic functions and operators
occurring in \w{} out of the conditions \conditionset{} have no
role, since the only aspect we need to guarantee for the validity
of \skw{} is that they are indeed functions, so that $\exists
y.(y=f(\ldots))$ is always valid.%
\footnote{This propagates down to the recursive structure of
  \wenc{} because  
$\vi$ is equivalent to $\exists y.(\vi[t|y]\wedge(y=t))$,
and, if $y$ does not occur in  $\psi$, then
$\exists y.(y=({\sf If}\ \psi\ {\sf Then}\ t_{1}\ {\sf Else}\
{t_{2}}))$ is equivalent to
$(({\sf If}\ \psi\ {\sf Then}\ \exists y.(y=t_{1})\ {\sf Else}\
{\exists y.(y=t_{2})}))$.
}
(In substance, during the enumeration we are interested only in the
truth values  of the conditions $\conditionset$ in $\mu$ which make
$\wmuagen{\mu}$ \FI{},
regardless of the actual values of $\wmuagen{\mu}$).
Therefore, we can safely substitute condition-less arithmetic
functions and operators with fresh \emph{uninterpreted function
symbols}, obtaining a $\larat\cup\euf$-formula \eufwenc{}, 
 which
is relatively easy to solve by standard SMT solvers
\citep{barrettsst21}.
It is easy to see that a partial assignment $\mu$ evaluating \wenc{} to
true is satisfiable in the theory iff its corresponding assignment $\mu_\euf$
is $\larat\cup\euf$-satisfiable.\footnote{This boils down to the fact
  that $y$ occurs only in the top equation and as such it is free to assume
  arbitrary values, and that all arithmetic functions are total in the
  domain so that, for every possible values of $\allx$ a value for 
$\ally$ always exists iff there exists in the \euf{} version.}
Thus, we can safely define $\vistarstar\defas(\vi\wedge\supportwff\wedge\eufwenc{})$,
without affecting the correctness of the procedure,
but making it much more efficient.

Finally, we enforce the fact that the two branches of an if-then-else
are alternative by adding a mutual-exclusion
constraint $\neg(y_i=t_{i1})\vee\neg(y_i=t_{i2})$ to~\eqref{eq:ite}, so that
the choice of the function is univocally associated to the set of
decisions on the $\psi_i$s.
The procedure generating \eufwenc{} is
described in detail in Algorithm~\ref{algo:convert}.



\begin{example}%
\label{ex:encoding}%
\label{ex:enumeration}
Consider the problem in Example~\ref{ex:issue1}.
Figure~\ref{fig:sa-wmi-pa-labeling} shows the relabelling process
applied to the weight function $\w$.  The resulting \eufwenc{} formula
is:

{
  \hspace{-.2cm}$\begin{array}{ll}
   \phantom{\wedge} 
   \wencxya{}\defas\\
   \phantom{\wedge}
   (\neg A_1 \vee \neg (x_1\ge 1)\vee \neg (x_2\ge 1) &\hspace{-.9cm}\vee \pos(y_1=\myeuf{11})) \\
\wedge(\neg A_1 \vee \neg (x_1\ge 1)\vee \pos (x_2\ge 1) &\hspace{-.9cm}\vee \pos(y_1=\myeuf{12})) \\
\wedge(\neg A_1 \vee \neg (x_1\ge 1)\vee \neg (y_1=\myeuf{11})&\hspace{-.9cm}\vee\neg (y_1=\myeuf{12})) \\
\wedge(\neg A_1 \vee \pos (x_1\ge 1)\vee \neg (x_2\ge 2) &\hspace{-.9cm}\vee \pos(y_2=\myeuf{21})) \\
\wedge(\neg A_1 \vee \pos (x_1\ge 1)\vee \pos (x_2\ge 2) &\hspace{-.9cm}\vee \pos(y_2=\myeuf{22})) \\
\wedge(\neg A_1 \vee \pos (x_1\ge 1)\vee \neg (y_2=\myeuf{21})&\hspace{-.9cm}\vee\neg (y_2=\myeuf{22})) \\   
\wedge(\neg A_1 \vee \neg (x_1\ge 1) &\hspace{-.9cm}\vee \pos(y_3=y_1)) \\
\wedge(\neg A_1 \vee \pos (x_1\ge 1) &\hspace{-.9cm}\vee \pos(y_3=y_2)) \\
\wedge(\neg A_1 \vee \neg (y_3=y_1)\vee\neg (y_3=y_2)) &\\   
\wedge(\pos A_1 \vee \neg A_2 &\hspace{-.9cm}\vee \pos(y_4=\myeuf{3})) \\
\wedge(\pos A_1 \vee \pos A_2 &\hspace{-.9cm}\vee \pos(y_4=\myeuf{4})) \\
\wedge(\pos A_1 \vee \neg (y_4=\myeuf{3})\vee\neg (y_4=\myeuf{4})) &\\   
\wedge(\neg A_1 &\hspace{-.9cm}\vee \pos(y_5=y_3)) \\
\wedge(\pos A_1 &\hspace{-.9cm}\vee \pos(y_5=y_4)) \\
\wedge(\neg (y_5=y_3)\vee\neg (y_5=y_4)) &\\   
\wedge(\pos (y=y_5)) &\\   
\end{array}$
}

\begin{figure}[th]
  \centering
  \includegraphics[width=\textwidth]{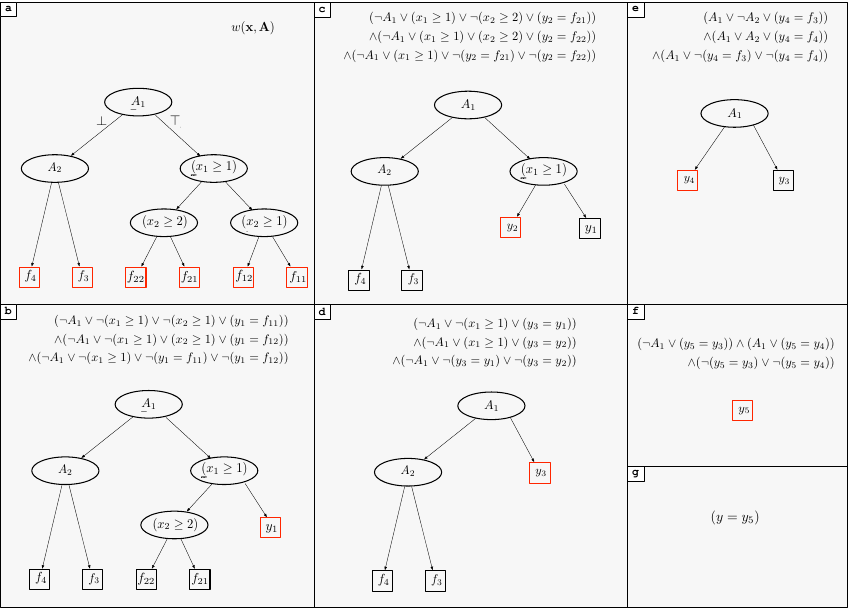}
  \caption{\label{fig:sa-wmi-pa-labeling} Example of bottom-up procedure for computing the relabelling function \eufwenc{}. ({\bf a}) Replacement of 
\FI{} weight functions (the leaves of the tree, highlighted in red) with \euf{} function symbols (we dropped the dependency on $x$ for compactness). ({\bf b-g}) Sequence of relabelling steps. At each step, a conditional term is replaced by a fresh \larat{} variable $y_i$. The encoding of the variable in shown in the upper part, while the lower part shows the weight function with the branch of the conditional term replaced with $y_i$ (highlighted in red). The last step consists in renaming the top variable as $y$, so that $y=\wxa{}$. The relabelling function \eufwenc{} is simply the conjunction of the encodings in the different steps.}
\end{figure}
\begin{figure}[th]
  \includegraphics[width=\textwidth]{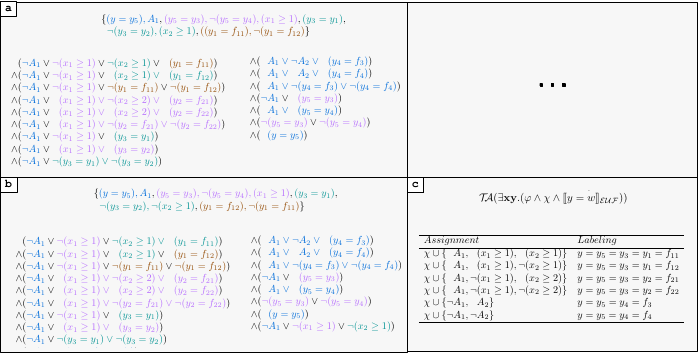}
  \caption{\label{fig:sa-wmi-pa-enumeration} Example of
    structure-aware enumeration performed by \wmisapa{} on the
    problem in Example~\ref{ex:issue1}. ({\bf a}) Generation of the
    first assignment. The assignment is on top, while the bottom part
    shows the \eufwenc{} formula. Colors indicate the progression of
    the generation, in terms of atoms added (top) and parts of the
    formula to be removed as a consequence (bottom). For the sake of
    simplicity, all atoms until the next atom in
    $\Psi\defas\set{A_1,A_2,(x_1\ge 1),(x_2\ge 1),(x_2\ge 2)}$ (if
    any) are given the same colour.  ({\bf b}) Generation of the second
    assignment. Note how the \eufwenc{} formula is enriched with the
    blocking clause $\neg A_1\vee\neg (x_1\ge 1)\vee \neg (x_2\ge 1)$
    preventing the first assignment to be generated again. ({\bf c})
    Final result of the enumeration (which contains six assignments in
    total). The partial assignments are obtained by restricting the
    generated assignments on the conditions in $\Psi$ (and combining
    them with the atoms of $\supportwff$, which here are assigned
    deterministically.).
    For each assignment, the
    corresponding chain of equivalences of the $\ally$s with the
    identified leaf \FI{} function is displayed.}
\end{figure}

Figure~\ref{fig:sa-wmi-pa-enumeration}
illustrates a possible enumeration process.
%
%
The algorithm
enumerates partial assignments satisfying
$\vistarstar\defas\vi\wedge\supportwff\wedge\eufwenc{}$, restricted to the atoms
$\Psi\defas\set{A_1,A_2,(x_1\ge 1),(x_2\ge 1),(x_2\ge 2)}$, which is
equivalent to enumerate
$\TA{\exists \ally. \vistarstar}$.
Assuming that 
the enumeration procedure picks nondeterministic choices  
following the order of the above set\footnote{Like in Algorithm~\ref{alg:wmipaeuf},
 we pick Boolean conditions first.}
and
assigning positive
values first, then in the first branch the following satisfying partial
assignment is generated, in order: 
\footnote{Here nondeterministic choices are \und{underlined}. The
  atoms in $\supportwff$ are assigned deterministically.}

\begin{center}
$
\begin{array}{l}\supportwff\cup
 \{(y=y_5),
  \und{A_1},(y_5=y_3),\neg(y_5=y_4),
  \und{(x_1\ge 1)},\\(y_3=y_1),\neg(y_3=y_2),
  \und{(x_2\ge 1)},(y_1=\myeuf{11}),\\\neg(y_1=\myeuf{12})\}.
\end{array}
$
\end{center}

(Notice that, following the chains of true equalities, we have $y=y_5=y_3=y_1=\myeuf{11}$.)
Then the SMT solver extracts from it the subset \set{A_1,(x_1\ge 1), (x_2\ge 1)}
restricted to the atoms in $\Psi$.\@
Then  the blocking clause $\neg
A_1\vee\neg (x_1\ge 1)\vee \neg (x_2\ge 1)$ is added to the formula, which prevents
to enumerate the same subset again.
This forces the algorithm to backtrack and generate:
\begin{center}
    $
\begin{array}{l}\supportwff\cup
  \{(y=y_5),
  \und{A_1},(y_5=y_3),\neg(y_5=y_4),
  \und{(x_1\ge 1)},\\(y_3=y_1),\neg(y_3=y_2),
  {\neg(x_2\ge 1)},(y_1=\myeuf{12}),\\\neg(y_1=\myeuf{11})\}.
\end{array}
$
\end{center}
\noindent
producing the assignment: \set{A_1,(x_1\ge 1), \neg(x_2\ge 1)}.%
\footnote{We refer the reader to~\cite{allsmt} for more details on the
  SMT-based enumeration algorithm.}

Overall, the algorithm
enumerates the following ordered collection of partial assignments
restricted to $\atoms{\vi\wedge\supportwff}\cup\conditionset$:\\
$
\begin{array}{ll}
\supportwff\cup\set{\pos A_1,\pos (x_1\ge 1), \pos (x_2\ge 1)},\ & //y=\ldots=\myeuf{11}\\
\supportwff\cup\set{\pos A_1,\pos (x_1\ge 1), \neg (x_2\ge 1)},\  & //y=\ldots=\myeuf{12}\\
\supportwff\cup\set{\pos A_1,\neg (x_1\ge 1), \pos (x_2\ge 2)},\  & //y=\ldots=\myeuf{21}\\
\supportwff\cup\set{\pos A_1,\neg (x_1\ge 1), \neg (x_2\ge 2)},\  & //y=\ldots=\myeuf{22}\\
\supportwff\cup\set{\neg A_1,\pos A_2},& //y=\ldots=\myeuf{3}\\
\supportwff\cup\set{\neg A_1,\neg A_2}& //y=\ldots=\myeuf{4}\\
\end{array}
$\\
\noindent
which correspond to the six integrals of Example~\ref{ex:issue1}.
Note that according to~\eqref{eq:newwmi1} the first four
integrals have to be multiplied by 2, because the partial assignment
\set{A_1} covers two total assignments \set{A_1,A_2} and \set{A_1,\neg A_2}.
Notice also that the disjunction of the six partial assignments:
\begin{eqnarray}
  \nonumber
 (A_1\wedge (x_1\ge 1)\wedge (x_2\ge 1)) \vee \ldots 
\vee (\neg A_1\wedge\neg A_2), 
\end{eqnarray}
 matches the definition of \skw{}, which we have computed by progressive enumeration rather
than encoded a priori. \hfill $\diamond$
\end{example}




Based on previous ideas, we proposed \wmisapa,
a novel ``weight-structure aware'' variant of \wmipa{}. The pseudocode of \wmisapa{} is reported in Algorithm~\ref{alg:wmipaeuf}.

\noi%
As with \wmipa{}, we enumerate the assignments in two main steps: in
the first loop
(lines~\ref{line:line2}-\ref{alg:line11}) we generate a set $\MUASTAR{}$ of partial assignments
\muastar{} over the Boolean atoms \allA, s.t.\ \vistarstarmuastar{} is
\larat-satisfiable and
does not anymore contain Boolean atoms. In
Example~\ref{ex:enumeration} 
$\MUASTAR{}\defas\set{\set{A_1},\set{\neg A_1,A_2},\set{\neg A_1,\neg A_2}}$.
In the second loop (lines~\ref{line:startPA}-\ref{line:endPA}),
for each \muastar{} in $\MUASTAR{}$ we enumerate the set $\MULARAT$ of
\larat-satisfiable 
partial assignments satisfying \vistarstarmuastar{} 
(that is, on \larat{} atoms in
$\atoms{\vi\wedge\supportwff}\cup\conditionset$),
we compute the integral \WMINBgen{\mularat}{\wmuastar}{\allx},
multiply it by the $2^{|\allA\setminus\muastar|}$ factor and add
it to the result.
In
Example~\ref{ex:enumeration}, if e.g.\ $\muastar{}=\set{A_1}$,
$\TA{\exists \allx\ally.\vistarstarmuastar}$
computes the four partial assignments $\{\supportwff\cup\set{(x_1\ge  1),(x_2\ge
    1)},\dots,\supportwff\cup\set{\neg(x_1\ge  1),\neg(x_2\ge  2)}\}$.

\begin{algorithm}[t]
\caption{{\sf SA-WMI-PA}$(\vi, \w, \allx, \allA)$%
\label{alg:wmipaeuf}}

\begin{algorithmic}[1]

\STATE $\MUASTAR\gets\emptyset; \vol \gets  0$
\STATE $\tuple{w', \eufwenc{}, \ally} \gets \convert{}(w, \emptyset)$
\STATE $\newvistar \gets \vi \wedge \supportwff \wedge\eufwenc{}$\label{line:line1}
\STATE $\MUA \gets\TA{\exists \allx\ally.\newvistar}$\label{line:line2}
\FOR{$\mua \in \MUA$}\label{line:line3start}
    \STATE ${\sf Simplify}(\vistarstarmua)$\label{alg:simply}
    \IF{\vistarstarmua does not contain Boolean atoms}
        \STATE $\MUASTAR \gets \MUASTAR \cup \set{\mua}$\label{alg:line11}
    \ELSE
    \FOR{$\mua_{residual} \in  \TTA{\exists \allx\ally.\vistarstarmua}$}\label{line:line8}
        \STATE $\MUASTAR \gets \MUASTAR \cup \set{\mua \wedge \mua_{residual}}$\label{line:line9}
    \ENDFOR
    \ENDIF
\ENDFOR\label{line:line3end}
\FOR{$\muastar \in \MUASTAR$}\label{line:startPA}
    \STATE $k \gets |\allA\setminus\muastar|$
    \STATE ${\sf Simplify}(\vistarstarmuastar)$ 
    \IF{${\sf LiteralConjunction}(\vistarstarmuastar)$}\label{line:literalstart}
        \STATE $\vol \gets \vol + 2^{k}\cdot \WMINBgen{\vistarstarmuastar}{\wmuastar}{\allx}$%
        \label{line:literalend}
    \ELSE\label{line:laratstart}
        \STATE $\MULARAT{} \gets \TA{\exists\ally.\vistarstarmuastar}$ 
        \FOR{$\mularat{} \in \MULARAT$} 
            \STATE $\vol \gets \vol + 2^{k} \cdot
            \WMINBgen{\mularat}{\wmuastar}{\allx}$\label{line:laratend}
        \ENDFOR 
    \ENDIF
\ENDFOR\label{line:endPA}
\RETURN $\vol$
\end{algorithmic}
\end{algorithm}

In detail, in line~\ref{line:line1} we extend $\vi\wedge\supportwff$
with $\eufwenc{}$ to provide structural awareness.
(We recall that,
unlike with \wmipa{}, we do not label \larat{} conditions with fresh
Boolean atoms $\allB{}$.)
%
Next, in line~\ref{line:line2} we perform
$\TA{\exists \allx\ally.\vistarstar}$ to obtain a set $\MUA{}$ of partial
assignments restricted to Boolean atoms \allA. Then, for each
assignment $\mua\in\MUA$ we build the (simplified) residual $\vistarstarmua$.
Since $\mua$ is partial,  $\vistarstarmua$ is not guaranteed to be free of Boolean
atoms $\allA$, as shown in Example~\ref{ex:muastar}.
If this is the case, we simply add \mua{} to $\MUASTAR$, 
otherwise, we invoke
$\TTA{\exists\allx\ally.\vistarstarmua}$ to assign the remaining
atoms and conjoin each 
assignment $\muAres$ to $\mua$, ensuring that the residual now
contains only \larat{} atoms (lines~\ref{line:line3start}-~\ref{line:line3end}).  
The second loop (lines~\ref{line:startPA}-\ref{line:endPA})
resembles the main loop in \wmipa{}, with the only relevant
difference that, since $\muastar$ is partial, the integral is
multiplied by a $2^{|\allA\setminus\muastar|}$ factor, as in~\eqref{eq:newwmi1}.

Notice that in general the assignments \muastar{} are partial even if
the steps in lines~\ref{line:line8}-\ref{line:line9} are executed; the set of residual Boolean
atoms in $\vistarstarmua$ are a (possibly much smaller) subset of $\allA\setminus\mua$ because some of them do not occur anymore in
$\vistarstarmua$ after the simplification, as shown in the following example.

\begin{example}%
  \label{ex:muastar}
Let $\vi\defas(A_1 \vee A_2 \vee A_3)\wedge(\neg A_1\vee A_2\vee (x \geq
1))\wedge(\neg A_2\vee (x \geq 2))\wedge(\neg A_3\vee (x \leq 3))$,
$\supportwff\defas(x_1\ge 0)\wedge(x_1\le 4)$ 
and $\wxa \defas 1.0$.\\
Suppose \TA{\exists \allx.\newvistar} finds the partial
assignment \set{(x \geq 0),(x \leq 4),A_2,(x \geq 1),(x \geq 2),(x
  \leq 3)}, whose projected version is $\mua\defas\set{A_2}$ (line~\ref{line:line2}).
Then $\vistarstarmua$ reduces to $(\neg A_3\vee (x \leq 3))$, so that
$\MUASTAR$ is  \set{\set{A_2,A_3},\set{A_2,\neg A_3}}, avoiding
branching on $A_1$. \hfill $\diamond$
\end{example}


We stress the fact that in our actual implementation, like with that
of \wmipa{}, the potentially-large sets $\MUASTAR$ and $\MULARAT$ are not generated
explicitly. Rather, their elements are
generated, integrated and then dropped one-by-one, so that to avoid
space blow-up.

\noi%
We highlight two main differences w.r.t.\ \wmipa{}.
First, unlike with \wmipa{}, the generated assignments $\mua$ on $\allA$ are partial,
each representing $2^{|\allA\setminus\mua|}$ total ones.
Second, the assignments on (non-Boolean) conditions $\conditionset$ inside
the \mularat{}s are also
partial, whereas with \wmipa{} the assignments to the \allB{}s are total.
This may drastically reduce the number of integrals to compute, as
empirically demonstrated in~\sref{sec:expeval}.

\section{Description of the datasets for the DET experiments}%
\label{sec:det-datasets}
Table~\ref{tab:datasets} describes the
datasets used to train the DETs considered in~\ref{sec:exp-det}. For
each dataset, we report the number of Boolean and continuous
variables, the size of the training/validation splits and the size of
the resulting DET in terms of number of internal nodes (conditions).

\begin{table}[t]
  \centering
    \begin{tabular}{|l|c|c|c|c|c|}
\hline

Dataset & $|\allA|$ & $|\allx|$ & \# Train & \# Valid & Size \\
\hline
balance-scale & 3 & 4 & 1875 & 205 & 5 \\
iris & 3 & 4 & 450 & 50 & 3 \\
cars & 33 & 7 & 2115 & 234 & 14 \\
diabetes & 1 & 8 & 4149 & 459 & 28 \\
breast-cancer & 12 & 4 & 1650 & 180 & 13 \\
glass2 & 1 & 9 & 970 & 100 & 7 \\
glass & 7 & 9 & 1280 & 140 & 7 \\
breast & 1 & 10 & 4521 & 495 & 31 \\
solar & 25 & 3 & 2522 & 273 & 13 \\
cleve & 17 & 6 & 2492 & 266 & 18 \\
hepatitis & 14 & 6 & 940 & 100 & 5 \\
heart & 3 & 11 & 2268 & 252 & 16 \\
australian & 34 & 6 & 6210 & 690 & 46 \\
crx & 38 & 6 & 6688 & 736 & 46 \\
german & 41 & 10 & 12600 & 1386 & 93 \\
german-org & 13 & 12 & 15000 & 1650 & 108 \\
auto & 56 & 16 & 2522 & 260 & 18 \\
anneal-U & 74 & 9 & 21021 & 2301 & 144 \\

\hline
\end{tabular}
\caption{UCI datasets considered in our experiments. For each dataset,
  we report the number of Boolean and continuous variables, number of
  training/validation instances and the size of the resulting Density
  Estimation Tree in terms of number of conditions.}%
    \label{tab:datasets}
\end{table}

\section{Details on the program fairness verification task}%
\label{sec:fair-programs}

The population model \pop{} (Algorithm~\ref{alg:pop}) is a probabilistic
program with Gaussian priors, modelling a distribution over
real-valued variables: (i) \ethnicity; \colRank{} (rank
of the college attended by the person, the lower, the better); (iii)
\yExp{} (years of work experience). In this population model,
\colRank{} is influenced by \ethnicity.
The decision program \dec{} (Algorithm~\ref{alg:dec}) takes as input
\colRank{} and \yExp{} and outputs \hire. Notice that,
while \dec{} doesn't have access to the sensitive condition $\minority =
\ethnicity > 10$, the program is still unfair (Figure~\ref{fig:fair}) due
to the high importance attributed to $\colRank$. This can be amended by
modifying the second condition to $(5\cdot \yExp - \colRank > -5)$,
putting more emphasis on the years of work experience. This
modification indeed makes the program pass the fairness criterion with
$\varepsilon = 0.1$ given \pop.

\begin{algorithm}[H]        
  \caption{\pop
    \label{alg:pop}}
  \begin{algorithmic}
    \STATE$\ethnicity \gets \mathcal{N}(0,10)$
    \STATE$\colRank \gets \mathcal{N}(25,10)$
    \STATE$\yExp \gets \mathcal{N}(10,5)$
    \IF{$\ethnicity > 10$}
    \STATE$\colRank \gets \colRank + 5$
    \ENDIF%
  \end{algorithmic}
\end{algorithm}
  
\begin{algorithm}[H]        
  \caption{{\sf dec}
    \label{alg:dec}}
  \begin{algorithmic}[1]
    \IF{$(\colRank \leq 5)$}
    \STATE$\hire \gets \mathit{True}$
    \ELSIF{$(\yExp - \colRank > -5)$}
    \STATE$\hire \gets \mathit{True}$
    \ELSE%
    \STATE$\hire \gets \mathit{False}$
    \ENDIF%
  \end{algorithmic}
\end{algorithm}

\end{document}